\documentclass[
11pt, %
english, %
headsepline, %
openany %
]{MastersDoctoralThesis} %

\usepackage[utf8]{inputenc} %
\usepackage[T1]{fontenc} %

\usepackage[autostyle=true]{csquotes} %
\usepackage{graphicx} %
\usepackage{longtable} %
\usepackage{geometry} %
\usepackage{hyperref}
\usepackage{pdflscape}
\usepackage{adjustbox}
\usepackage{rotating}
\usepackage{lscape}
\usepackage[graphicx]{realboxes}
\usepackage{subcaption}

\thesistitle{Contrastive Learning for Character Detection in Ancient Greek Papyri} 
\supervisor{
     \href{https://www.unifr.ch/inf/en/all/people/16738/0a54b}{Prof. Rolf Ingold} \\ 
     \href{https://www.unifr.ch/inf/en/all/people/2866/825d2}{Prof. Andreas Fischer} \\ 
    \href{https://www.unifr.ch/inf/de/all/people/186328/4d524}{Lars Vögtlin}
}
\examiner{} %
\degree{Master of science} 
\author{\href{https://www.linkedin.com/in/vedasri-goshke-8a7865b3/}{Vedasri Nakka}}

\university{\href{https://www.unifr.ch/home/en/}{University of Fribourg}}
\department{\href{https://www.unifr.ch/inf/diva/en/}{Document, Image and Video Analysis(DIVA)}} 
\group{\href{https://www.unifr.ch/inf/en/}{Department of Informatics}} 
\AtBeginDocument{
\hypersetup{pdftitle=\ttitle} %
\hypersetup{pdfauthor=\authorname} %
\hypersetup{pdfkeywords=\keywordnames} %
}

\begin{document}

\frontmatter %

\pagestyle{plain} %

\begin{titlepage}
\begin{center}

\begin{figure}
    \centering
        \includegraphics[width=.2\textwidth]{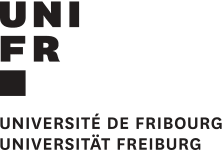}
    \hfill
        \includegraphics[width=.2\textwidth]{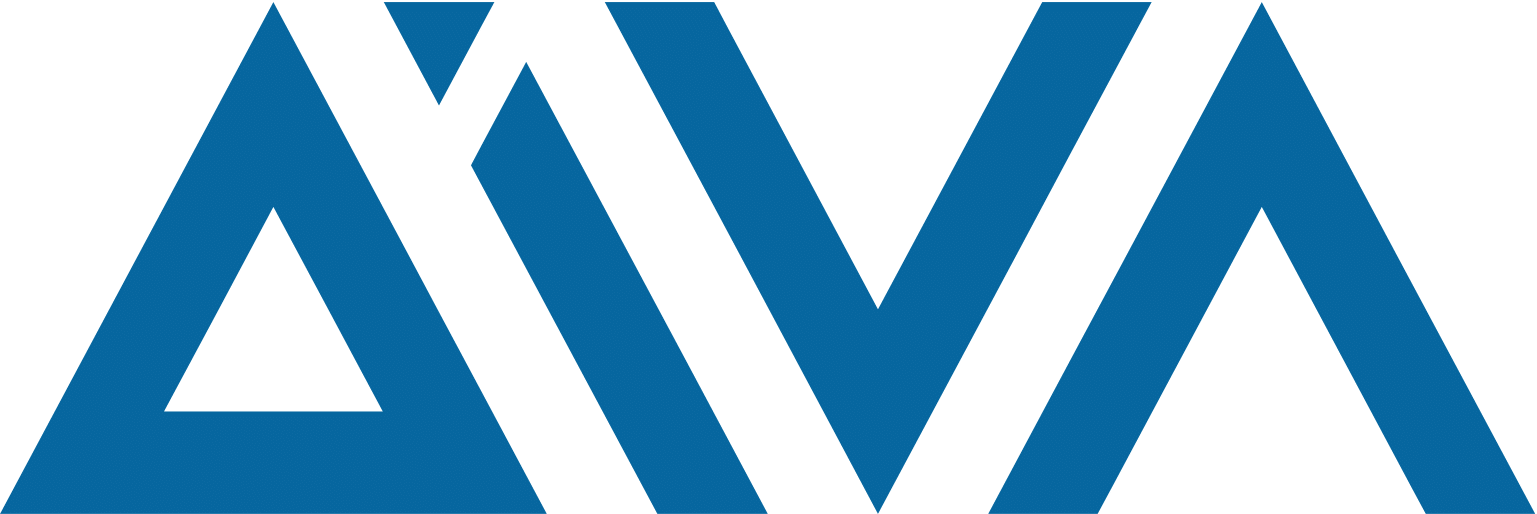}
\end{figure}

\vspace*{.06\textheight}
{\scshape\LARGE \univname\par}\vspace{1.5cm} %
\textsc{\Large Master Thesis}\\[0.5cm] %

\HRule \\[0.4cm] %
{\huge \bfseries \ttitle\par}\vspace{0.4cm} %
\HRule \\[1.5cm] %
 
\begin{minipage}[t]{0.4\textwidth}
\begin{flushleft} \large
\emph{Author:}\\
\href{https://www.linkedin.com/in/vedasri-goshke-8a7865b3/}{Vedasri Nakka} %
\end{flushleft}
\end{minipage}
\begin{minipage}[t]{0.4\textwidth}
\begin{flushright} \large
\emph{Supervisor:} \\
\href{https://www.unifr.ch/inf/en/all/people/16738/0a54b}{\supname} %
\end{flushright}
\end{minipage}\\[2cm]

\textit{in the}\\[0.3cm]
\groupname\\\deptname\\[1cm] %
 
\vfill

{\large \today}\\[2cm] %
 \footnotesize{D\'epartement d'Informatique - Departement f\"{u}r Informatik  $\bullet$
            Universit\'e~de~Fribourg - Universit\"{a}t~Freiburg $\bullet$
            Boulevard de P\'erolles 90 $\bullet$ 1700~Fribourg $\bullet$ Switzerland 
            \vskip \baselineskip \par
            }

\vfill
\end{center}
\end{titlepage}

\renewcommand{\abstractname}{Abstract} %

\begin{abstract}
\addchaptertocentry{\abstractname} %

\noindent This thesis investigates the effectiveness of SimCLR~\cite{simclr}, a contrastive learning technique, specifically in the context of Greek letter recognition, and examines the impact of various augmentation techniques. To achieve this, we use a large Alpub dataset~\cite{alpub} (pretraining dataset) to pretrain the SimCLR backbone, followed by fine-tuning on a smaller ICDAR~\cite{icdar} dataset (finetuning dataset) to evaluate the performance of SimCLR in comparison to traditional baseline models using cross-entropy and triplet loss functions. Furthermore, our work explores the impact of several data augmentation strategies, a critical component of the SimCLR training pipeline. \\

\noindent Methodologically, our study examines three primary approaches: {\bf (1)} a Baseline model with cross-entropy loss, {\bf (2)} a Triplet embedding model, enhanced with a classification layer, and {\bf (3)} a SimCLR pretrained model with a classification layer. Initially, we train the baseline model, triplet model, and SimCLR with 93 different augmentations on ResNet-18 and ResNet-50 networks~\cite{resnet} using the ICDAR dataset. From these, we select the \emph{top-4} augmentations based on the results of a statistical t-test. Finally, we conduct pretraining of SimCLR on the large Alpub dataset, followed by fine-tuning on the smaller ICDAR dataset. The triplet loss model undergoes a similar training process, being pretrained on the top-4 augmentations using the Alpub dataset, and then fine-tuned on the ICDAR dataset.\\

\noindent Our experiments reveal that SimCLR does not outperform the baselines in letter recognition tasks. The baseline model using cross-entropy loss demonstrates superior performance compared to both SimCLR and the triplet loss method. This study provides a detailed evaluation of contrastive learning for letter recognition and highlights the limitations of SimCLR, emphasizing the effectiveness of traditional supervised learning models in this specific application. We believe that the cropping strategies involved in SimCLR lead to a semantic shift of the input image, thereby reducing the effectiveness of training, despite the large amount of pretraining data used. Our code is available at \url{https://github.com/DIVA-DIA/MT_augmentation_and_contrastive_learning/}.

\end{abstract}

\keywordnames

\tableofcontents %

\begin{abbreviations}{ll} %

\textbf{SimCLR} & \textbf{S}imple \textbf{F}ramework for \textbf{C}ontrastive \textbf{L}earning of Visual \textbf{R}epresentations \\
\textbf{ResNet-18} & \textbf{R}esidual Neural \textbf{N}etwork - 18\\
\textbf{ResNet-50} & \textbf{R}esidual Neural \textbf{N}etwork - 50\\
\textbf{ICDAR} & The \textbf{I}nternational \textbf{C}onference on \textbf{D}ocument \textbf{A}nalysis and \textbf{R}ecognition \\
\textbf{InfoNCE} & \textbf{N}oise-\textbf{C}ontrastive \textbf{E}stimation \\
\textbf{CE} & \textbf{C}ross \textbf{E}ntropy loss \\
\textbf{ALPUB} & \textbf{A}ncient \textbf{L}ives \textbf{P}roject for \textbf{U}niversity of \textbf{B}asel \\

\end{abbreviations}

\mainmatter %

\pagestyle{thesis} %

\chapter{Introduction} %
\label{Chapter1}

In recent years, contrastive learning~\cite{CVsupervised, CVcontrastive, CVtian2020makes, CVwang2023cssl, CVzhang2022chaco} has emerged as a powerful unsupervised learning technique~\cite{ULbarlow1989unsupervised, ULghahramani2003unsupervised, ULhastie2009unsupervised, ULdike2018unsupervised, ULdy2004feature, ULweber2000unsupervised} within the field of computer vision, offering promising results across various applications. Among these methods, SimCLR (Simple Framework for Contrastive Learning of Visual Representations), introduced by Chen et al.~\cite{simclr}, has shown remarkable success in a range of image recognition tasks by leveraging the power of contrastive pretraining. As a result, SimCLR has been extended to a variety of computer vision tasks, including semantic segmentation~\cite{SShu2021region, SShenaff2021efficient, SSzhong2021pixel}, object detection~\cite{ODchen2020improved, ODliu2020self, ODxie2021detco}, and object tracking~\cite{OTkim2023ssl, OThuang2023multi}, among others. However, its application in more specialized domains, such as letter recognition, has not been thoroughly explored. This thesis seeks to address this gap by investigating the effectiveness of SimCLR in Greek letter recognition, specifically through pretraining on the large-scale Alpub dataset~\cite{alpub}, followed by fine-tuning on the domain-specific ICDAR dataset~\cite{icdar}. \\

\noindent {\bf Research Questions.} The central question guiding this research is: \emph{How effective is SimCLR contrastive pretraining for letter recognition tasks compared to traditional baseline methods, and which data augmentations enhance its performance?} Specifically, we seek to determine whether SimCLR can outperform baseline models that are traditionally trained using cross-entropy and triplet loss functions in distinguishing between various classes of letters.\\

\noindent {\bf The primary objectives of this study are:}

\begin{itemize}
    \item To assess the performance of SimCLR in letter recognition tasks using both a large dataset (ALPUB) and a smaller ICDAR dataset.
    \item To compare SimCLR's performance with that of baseline models trained with cross-entropy and triplet loss.
    \item To identify effective data augmentations that improve the performance of SimCLR and baseline models.
\end{itemize}

\noindent To systematically investigate these objectives, we structure our analysis across the following chapters:

\begin{itemize}
    \item \texttt{Chapter 2:} Related Work reviews existing literature, covering essential topics such as handwritten character recognition, various neural architectures, contrastive learning methods, triplet loss, and data augmentation techniques. %
    \item \texttt{Chapter 3:} Methodology outlines the experimental design and the approach taken to address our research questions.
    \item \texttt{Chapter 4:} Results presents the outcomes of our experiments, focusing on the comparative performance of the different methods.
    \item \texttt{Chapter 5:} Discussion offers a detailed analysis of the results, exploring the implications of our findings.
    \item \texttt{Chapter 6:} Conclusion summarizes the thesis, discusses the limitations encountered, and suggests directions for future research.
\end{itemize}

\noindent {\bf Our Contributions.} This thesis contributes to the field by exploring the application of SimCLR contrastive pretraining in letter recognition, a task that requires precise classification of letter forms. We focus on two datasets: the ALPUB dataset, comprising 24 classes of letters, and a smaller ICDAR dataset, which includes 25 classes with 153 full-size training images and 34 test images, which are cropped for every letter to obtain a {total of 34,061 cropped images.}  An example of Greek papyri images is shown in Figure \ref{fig:greek_papyri}. The study employs three distinct methods to assess performance:

\begin{figure}
    \centering
    \includegraphics[width=\textwidth]{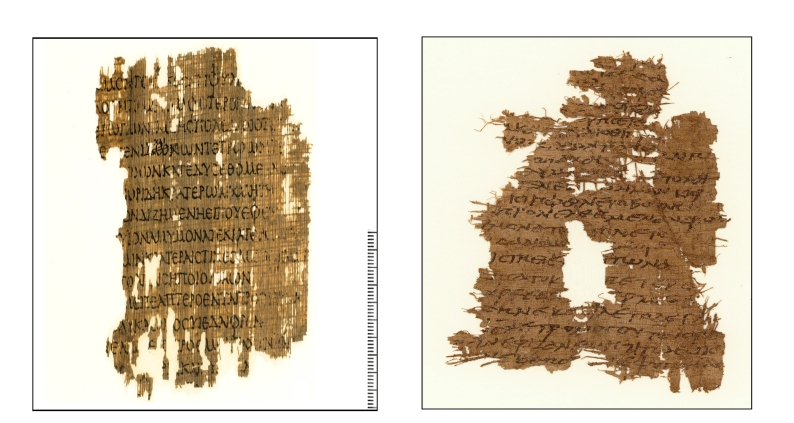}
    \caption{{\bf Reference images.} Sample full images from the ICDAR dataset containing Greek letters. We use ground-truth annotations to crop each letter subimage from the full images. We train models on the cropped letter images, and the finally evaluate the performance on unseen test data.
    }
    \label{fig:greek_papyri}
\end{figure}

\begin{enumerate}
    \item A baseline model trained with cross-entropy loss \vspace*{-0.2cm}
    \item A triplet model trained with triplet loss \vspace*{-0.2cm}
    \item A SimCLR model trained with InfoNCE loss
\end{enumerate}

\noindent By comparing these methods, we aim to understand the relative performance of SimCLR against traditional baselines and to explore whether contrastive learning offers significant advantages in letter recognition tasks. \\

\noindent The experimental process begins with training the baseline models using 93 different augmentations on ResNet-18 and ResNet-50 networks~\cite{resnet}. The top-4 augmentations are selected based on their performance, after which the SimCLR model is trained with these augmentations on the large Alpub dataset~\cite{alpub} and fine-tuned on the smaller ICDAR dataset~\cite{icdar}. The triplet loss model follows a similar procedure, being trained and fine-tuned with the selected augmentations on both datasets, using the ResNet-18 and ResNet-50 architectures.\\

\noindent Our findings indicate that SimCLR does not surpass traditional methods in letter recognition. The baseline model using cross-entropy loss outperforms both SimCLR and the triplet loss model. This study thus provides valuable insights into the limitations of SimCLR for letter recognition tasks and highlights the continued relevance of traditional supervised learning models in this domain. By systematically evaluating these methods, we contribute to the broader research in computer vision and offer practical insights for improving letter recognition systems.

\chapter{Related Work} %

In this chapter, we will review the literature related to our work. We have segregated the works based on their themes and identified four distinct directions.\\

\label{Chapter2} %
\noindent {\bf Letter Recognition.} Handwritten character recognition, particularly for specific alphabets like Greek~\cite{HWmarkou2021convolutional, HWItallapragada2022greek}, has been a research focus for many years. Early approaches relied on traditional machine learning techniques with hand-crafted features. Techniques such as Scale-Invariant Feature Transform (SIFT)~\cite{lowe2004distinctive}, Histogram of Oriented Gradients (HOG)~\cite{dalal2005histograms}, and Local Binary Patterns (LBP)~\cite{ojala2002multiresolution} were commonly used to extract features from images. These methods, however, required significant domain expertise and often struggled to generalize across different handwriting styles and scripts~\cite{HCsanchez2024importance, HCvashist2020comparative}.\\

\noindent The advent of deep learning around 2013 revolutionized character recognition~\cite{krizhevsky2012imagenet, CRyin2013icdar}, leading to substantial improvements in accuracy and robustness. Convolutional Neural Networks (CNNs)~\cite{lecun1998gradient, cnn} have since become the backbone of modern character recognition systems. Advanced architectures such as ResNet~\cite{resnet,he2016deep}, VGG~\cite{simonyan2014very}, and DenseNet~\cite{huang2017densely} have demonstrated exceptional performance in various image recognition tasks, including handwritten character recognition~\cite{HWNNmatan1990handwritten, HWItallapragada2022greek}. These models typically use cross-entropy loss for classification~\cite{crossentropy}, which is widely adopted due to its simplicity and effectiveness. Nevertheless, traditional supervised learning methods have their limitations. They require large amounts of labelled data and can be prone to overfitting, especially with small or imbalanced datasets~\cite{supervisedchallenges}. These challenges have driven researchers to explore alternative approaches that utilize data more efficiently and improve model generalization without heavy reliance on labelled datasets~\cite{simclr, moco, byol, semisupervised, ULbarlow1989unsupervised}.\\

\noindent {\bf Contrastive Learning.} Contrastive learning methods, like SimCLR~\cite{simclr}, have emerged as powerful techniques in self-supervised learning. These methods train models to differentiate between similar and dissimilar data pairs. SimCLR, in particular, uses extensive data augmentations to create positive pairs (views of the same image) and negative pairs (views of different images), enabling the learning of feature representations without labeled data. The effectiveness of SimCLR has been demonstrated across various visual recognition tasks~\cite{visualrecodecaf, visual}, often outperforming traditional supervised learning methods. However, its application in handwritten character recognition, particularly in non-Latin scripts like Greek, has not been thoroughly explored. Most research has focused on commonly used datasets such as CIFAR-10~\cite{cifar-10} and ImageNet~\cite{krizhevsky2012imagenet}, leaving a gap in understanding how contrastive learning methods perform in more specialized contexts~\cite{CVcontrastive, supervisedcontrastivelearning}.\\

\noindent {\bf Triplet Loss.} Triplet loss, introduced by Hoffer and Ailon (2015)~\cite{triplet}, is another method used for learning discriminative embeddings by comparing an anchor image with a positive (similar) and a negative (dissimilar) image. This approach has been widely applied in tasks like face recognition~\cite{schroff2015facenet, tripletfacerecognition}, 
person reidentification~\cite{personreidentification, defensetripletpersonreide}, Image retrieval~\cite{imageRetrival_1, localizedtripletimageretrival}, fine-grained image recognition~\cite{localizedtripletloss} and has shown promise in character recognition as well. Triplet loss aims to bring embeddings of similar examples closer in the embedding space while pushing dissimilar ones apart. Despite its simplicity and effectiveness, triplet loss methods face challenges such as careful triplet selection and high computational cost during training~\cite{hermans2017defense}. These challenges have limited its widespread adoption in character recognition tasks, especially when dealing with large and complex datasets~\cite{tripletlimitation}.\\

\noindent {\bf Data Augmentation.} Data augmentation is essential for enhancing the generalization capabilities of machine learning models, particularly in visual tasks. Augmentations introduce variability into the training data, enabling models to learn more robust and diverse features. The Albumentations library~\cite{albumentations} provides a comprehensive suite of augmentation techniques, including both spatial and pixel-level transformations, which have been effectively applied to tasks such as object detection~\cite{dataaugobjectdetection} and segmentation~\cite{dataaugsegmentation, dataaugsegmentation22}. Furthermore, advanced techniques like Mixup~\cite{mixupdataaug}, StyleMix~\cite{stylemix}, and CutMix~\cite{cutmix} have demonstrated superior performance, pushing the boundaries of data augmentation strategies. Previous studies have highlighted that the choice of augmentation strategies can significantly influence model performance, especially in contrastive learning frameworks where augmentations are crucial for generating positive and negative pairs~\cite{cubuk2020randaugment}. Despite these advancements, the specific impact of various augmentations on the performance of models like SimCLR in handwritten character recognition remains underexplored, presenting an open question in the field~\cite{dataaug_handwritten}.\\

\noindent {\bf Our Work.} This thesis builds on prior work by applying SimCLR to the task of Greek letter recognition and comparing its performance with traditional models trained using cross-entropy and triplet loss functions. By systematically evaluating the impact of different augmentation strategies~\cite{albumentations} on SimCLR's performance, this research provides new insights into the strengths and limitations of contrastive learning in the context of handwritten character recognition~\cite{icdar}. The findings contribute to a deeper understanding of how contrastive learning models, particularly SimCLR, perform with specialized datasets~\cite{icdar,alpub} and tasks that involve subtle distinctions between characters. In the following chapter, we will elaborate on different data augmentations and training methods, including SimCLR, for the task of letter recognition.

\chapter{Methodology} 
\label{Chapter3} 

In this chapter, we shall first explore and provide a detailed discussion on the various data augmentations that were investigated in the context of Greek-letter recognition in Section~\ref{sec:primary_augmentations}. Following this, in Section~\ref{sec:training_methods}, we will elaborate comprehensively on the three distinct approaches that were examined during the course of this study. Ultimately, this chapter details the training pipeline of the core methods that form the foundation for the experiments in the following chapters. Let us now first study the diverse data augmentations in detail. \\

\section{Primary Augmentations}
\label{sec:primary_augmentations}
\noindent Data augmentation~\cite{dataaugsurvey, dataaug_handwritten, dataaugobjectdetection} is one of the key components in modern machine learning algorithms~\cite{simclr, triplet}. Indeed, several simple augmentation strategies~\cite{autoaugmentstrategies, dataauglearning} have led to significant improvements in results, particularly for tasks such as ImageNet classification~\cite{krizhevsky2012imagenet}. Moreover, in the specific context of contrastive learning~\cite{CVsupervised, CVcontrastive, CVtian2020makes, CVwang2023cssl, CVzhang2022chaco}, the algorithm relies on learning from two cropped views of the same image. Therefore, our initial goal is to conduct a comprehensive study of different data augmentation strategies that will be employed in conjunction with various training methods.\\

\noindent In this section, we will discuss the primary augmentations that were explored for training the Greek-letter recognition models. These primary augmentations will be combined to form complex, higher-order data augmentation strategies. To achieve this, we utilize the Albumentations~\cite{albumentations} library and conduct experiments on ten primary augmentations, comprising six spatial and four pixel-level augmentations. We will now discuss each augmentation in detail to provide a deeper understanding and to highlight the associated hyperparameters. Pixel-level augmentations modify the image at each pixel independently. In contrast, spatial-level transforms modify the input image at a global level, simultaneously affecting the entire image and any associated targets.\\

\noindent {\bf Spatial Augmentations.} We consider six spatial augmentations for our analysis. Spatial augmentations are a key step in the preprocessing pipeline of the training methods.

\begin{itemize}
\item { \texttt{\bf Resize256.}} The resize transformation resizes the input of any given image to a fixed dimension of $256 \times 256$ pixels (shown in image \ref{fig:resize_image_spacial}). This ensures uniformity in image dimensions across the dataset, where images typically exhibit diverse resolutions. Performing this step is crucial for training neural networks~\cite{cnn, krizhevsky2012imagenet, HWNNmatan1990handwritten}, which often require a consistent input size. By applying this transformation, we standardize the input data, making it compatible with model requirements~\cite{simclr, triplet} and also improving training efficiency. Each input image in the batch undergoes this resizing process to maintain standardized input dimensions. It is important to note that we perform this data augmentation step on all images during both the training and evaluation phases, and across all training methods. 

\item {\texttt{\bf Randomcrop224.}} The random crop transformation extracts a $224 \times 224$ pixel region from the $256 \times 256$ image at a random location. This introduces variability in the training images by focusing on different sections of each image, which helps the model to better generalize and handle variations in object positioning. This technique is particularly useful for simulating different views of the letter images during training, ensuring that the model can recognize letters irrespective of their location. As illustrated in Figure~\ref{fig:randomcrop224_image}, a 16-image batch is randomly cropped to a size of $224 \times 224$ pixels, effectively cropping 76\% of the original image area. It is important to note that this augmentation step is highly sensitive, as cropping too small a region can shift the semantic meaning of the letter to a different label, while cropping too large a region makes the augmentation trivial.

\begin{figure}
    \centering
    \begin{subfigure}[b]{0.8\textwidth}
        \centering
        \includegraphics[width=\textwidth]{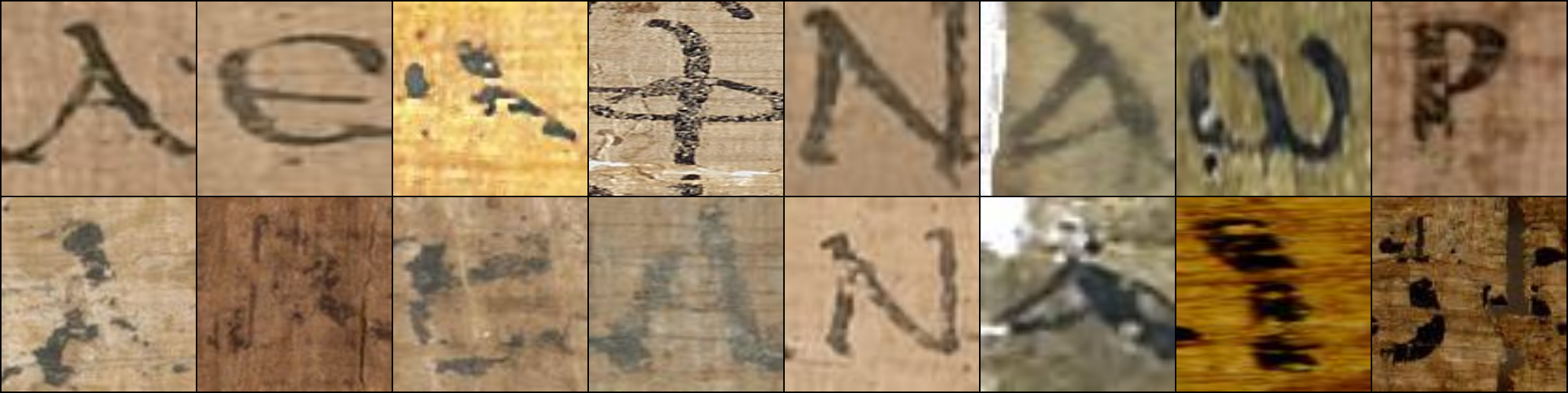}
        \caption{{\bf Resize256} augmentation. The original images of varying resolutions are resized to a consistent resolution of $256 \times 256$ pixels.}
        \label{fig:resize_image_spacial}
    \end{subfigure}
    \vspace{0.3cm}
    \begin{subfigure}[b]{0.8\textwidth}
        \centering
        \includegraphics[width=\textwidth]{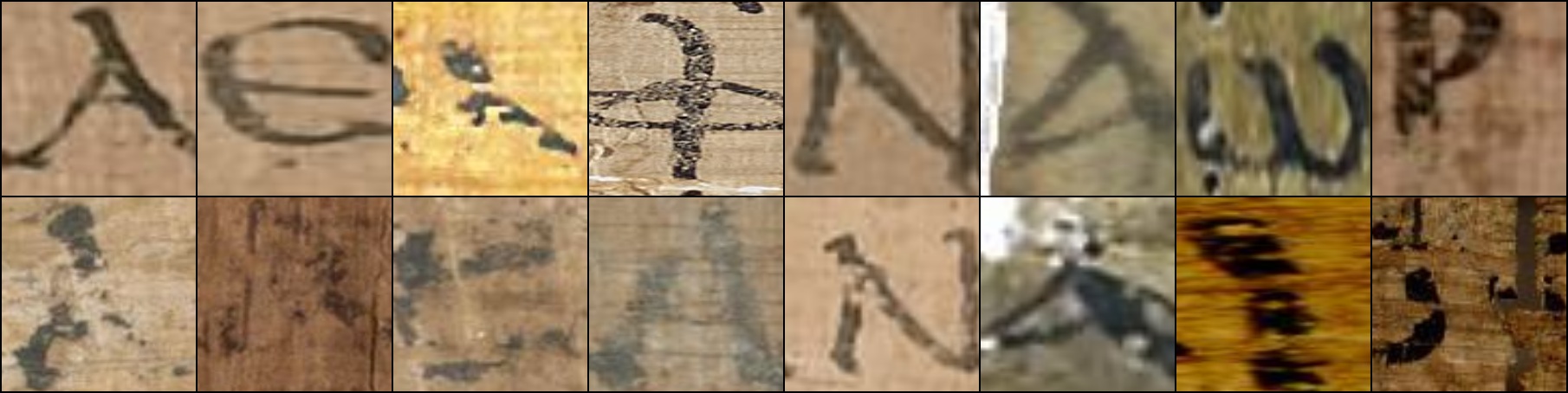}
        \caption{{\bf Randomcrop224} augmentation. The images are cropped to a size of $224 \times 224$ pixels, retaining 76\% of the original image area.}
        \label{fig:randomcrop224_image}
    \end{subfigure}
    \vspace{0.3cm}
    \begin{subfigure}[b]{0.8\textwidth}
        \centering
        \includegraphics[width=\textwidth]{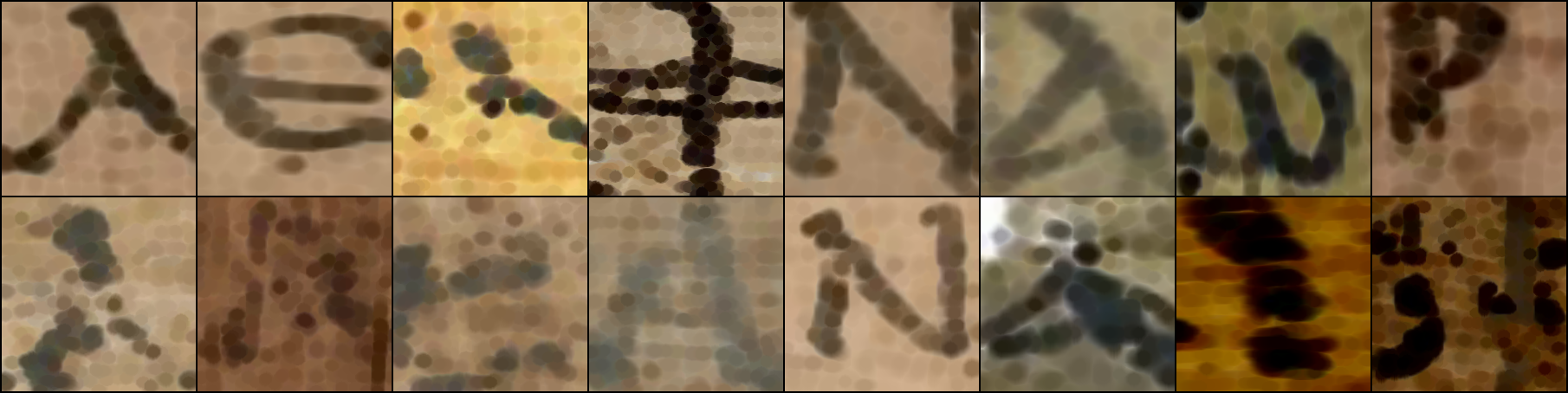}
    \caption{{\bf Morpho Erosion} removes pixels from the boundaries of objects using a $13\times13$ kernel.}
    \label{fig:morpho_erosion_image}
    \end{subfigure}
    \vspace{0.3cm}
    \begin{subfigure}[b]{0.8\textwidth}
        \centering
         \includegraphics[width=\textwidth]{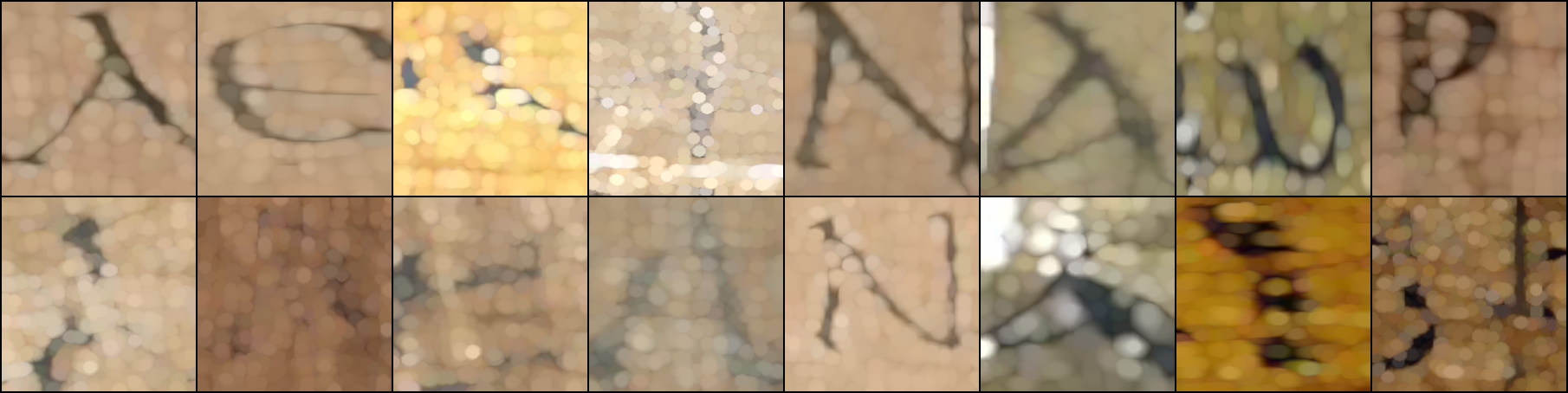}
        \caption{{\bf Morpho Dilation} expands the boundaries of objects using a $13 \times 13$ kernel.}
        \label{fig:morpho_dilation_image}
     \end{subfigure}
    \vspace{0.3cm}
    \begin{subfigure}[b]{0.8\textwidth}
        \centering
        \includegraphics[width=\textwidth]{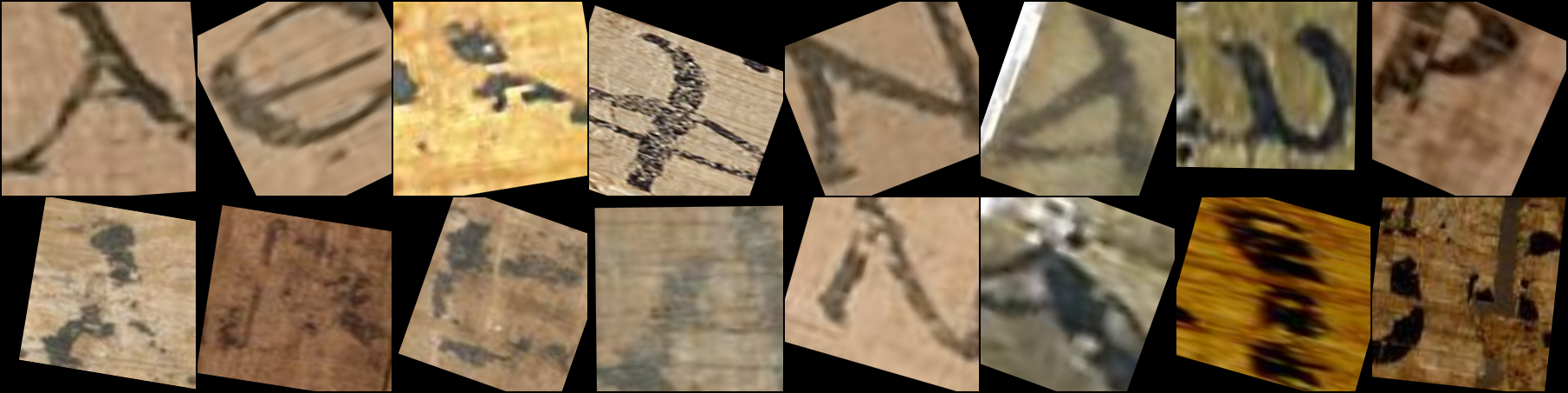}
        \caption{{\bf Affine} transformation combines shifting (0.05), scaling(0.1), and rotating(30) the image.}
        \label{fig:affine_image}
        \end{subfigure}
    \vspace{0.3cm}
    \begin{subfigure}[b]{0.8\textwidth}
        \centering
        \includegraphics[width=\textwidth]{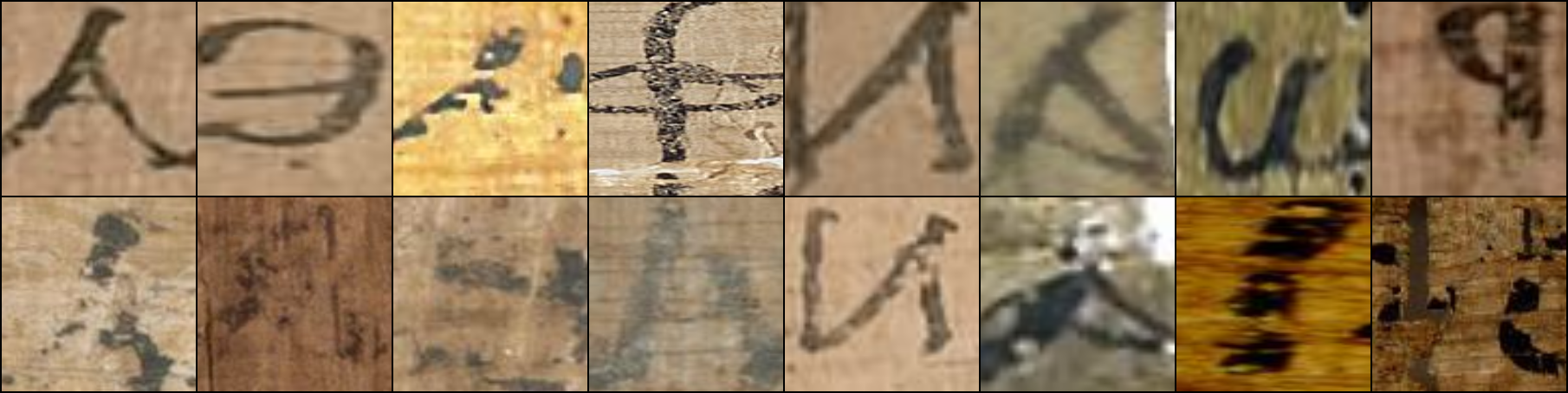}
        \caption{{\bf Hflip} horizontally flips the images.}
        \label{fig:hflip_image}
        \end{subfigure}
    \caption{Visualizations of the {\bf spatial augmentations} applied in our experiments. The first block shows the original resized image at a $256 \times 256$ resolution, and the following blocks show the resulting visualizations from different augmentation strategies.}
\end{figure}

\item {\bf Erosion.} This is a widely used augmentation in the context of letter images. Morphological erosion expands the size of objects in the image by adding pixels to their boundaries using a $7\times7$ kernel. This operation enhances the visibility of features by making objects more pronounced, which can improve object detection and recognition. It is particularly useful for making features more distinct and easier for the model to detect, especially in cases where objects have thin or irregular boundaries. We can observe the effect of erosion on a 16-image batch in Figure~\ref{fig:morpho_erosion_image}.

\item {\bf Dilation.} Morphological dilation reduces the size of objects in the image by removing pixels from their boundaries using a $7\times7$ kernel. This operation is helpful for eliminating small noise and artifacts by shrinking object boundaries. It improves image quality by reducing the impact of minor irregularities and focusing on the main features of the objects. The effect of dilation on a 16-image batch can be seen in Figure~\ref{fig:morpho_dilation_image}.

\item{\bf Affine):} The affine (shown in image~\ref{fig:affine_image}) transformation combines shifting, scaling, and rotating the image with a shift limit of 5\%, scale limit of 10\%, and rotation limit of 30 degrees. This augmentation simulates various perspectives and distortions, helping the model to handle changes in object positioning, size, and orientation. By applying these transformations, the model becomes more versatile in recognizing objects under different spatial conditions. 

\item {\bf Hflip.} The horizontal flip, as illustrated in Figure~\ref{fig:hflip_image}, mirrors the image along the vertical axis. This augmentation introduces left-right symmetry into the dataset, helping the model become invariant to horizontal orientations. By applying this transformation with a 50\% probability, the model can better generalize and recognize objects regardless of their horizontal position in the image.

\end{itemize}

\noindent {\bf Pixel-level Augmentations.} We consider four pixel augmentations for our analysis. Note that the pixel-level augmentations are applied to each pixel in the image independently with the probability $p$ during the training. \\

\begin{figure}
    \centering
    \begin{subfigure}[b]{0.8\textwidth}
        \centering
        \includegraphics[width=\textwidth]{Figures/batch_data_9_resize256_train_new.png}
        \caption{{\bf Resize256} augmentation.}
        \label{fig:resize_image}
    \end{subfigure}
    \vspace{0.5cm}
    
    \begin{subfigure}[b]{0.8\textwidth}
        \centering
        \includegraphics[width=\textwidth]{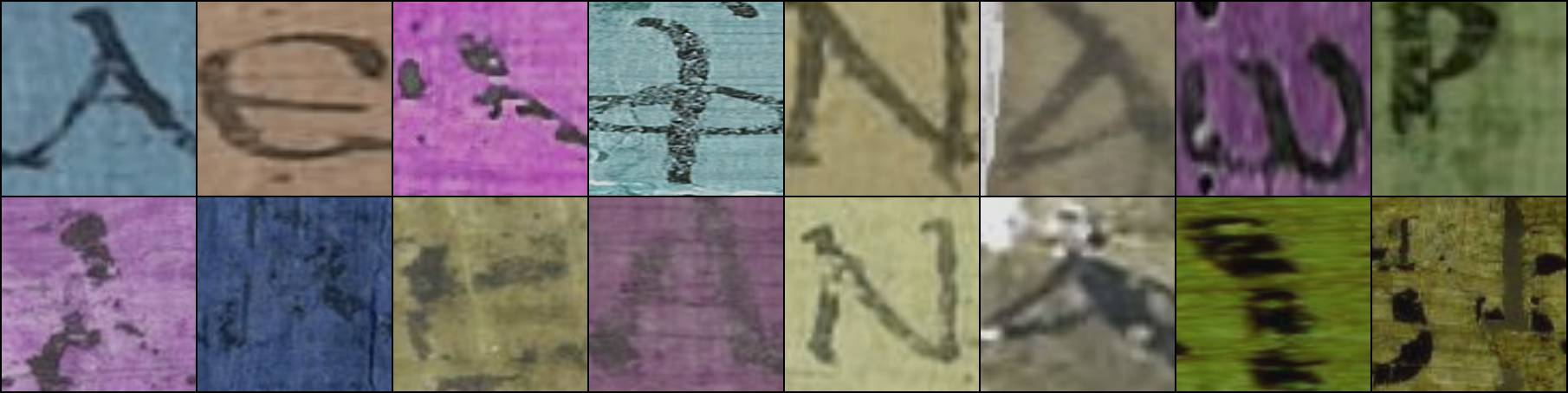}
        \caption{{\bf Colorjitter} augmentation applied to the images.}
        \label{fig:colorjitter_image}
    \end{subfigure}
    \vspace{0.5cm}
    \begin{subfigure}[b]{0.8\textwidth}
        \centering
        \includegraphics[width=\textwidth]{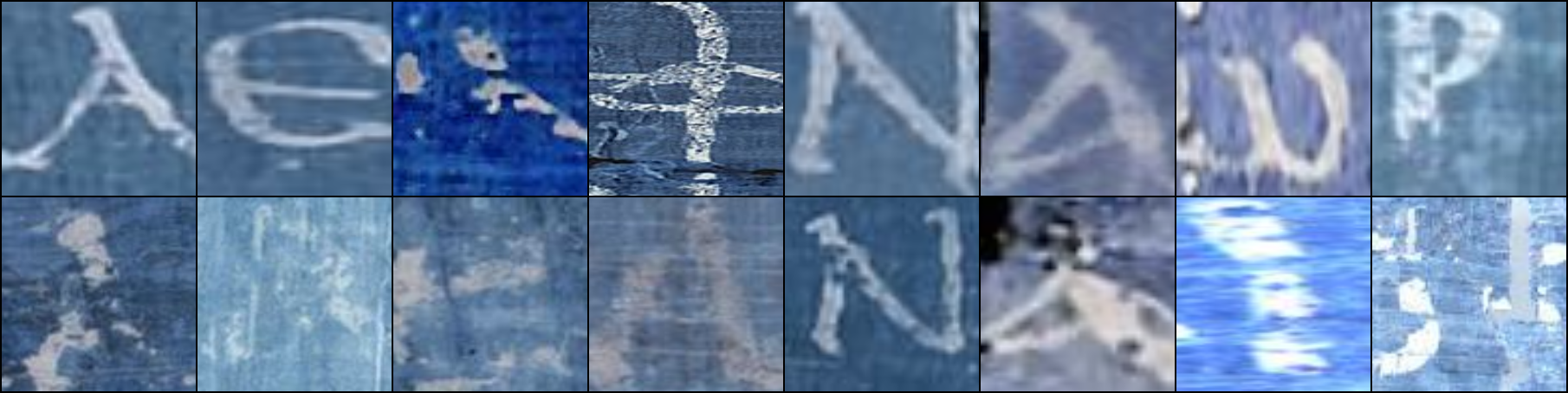}
        \caption{{\bf Invert} augmentation applied to the images.}
        \label{fig:invert_image}
    \end{subfigure}
    \vspace{0.5cm}
    \begin{subfigure}[b]{0.8\textwidth}
        \centering
        \includegraphics[width=\textwidth]{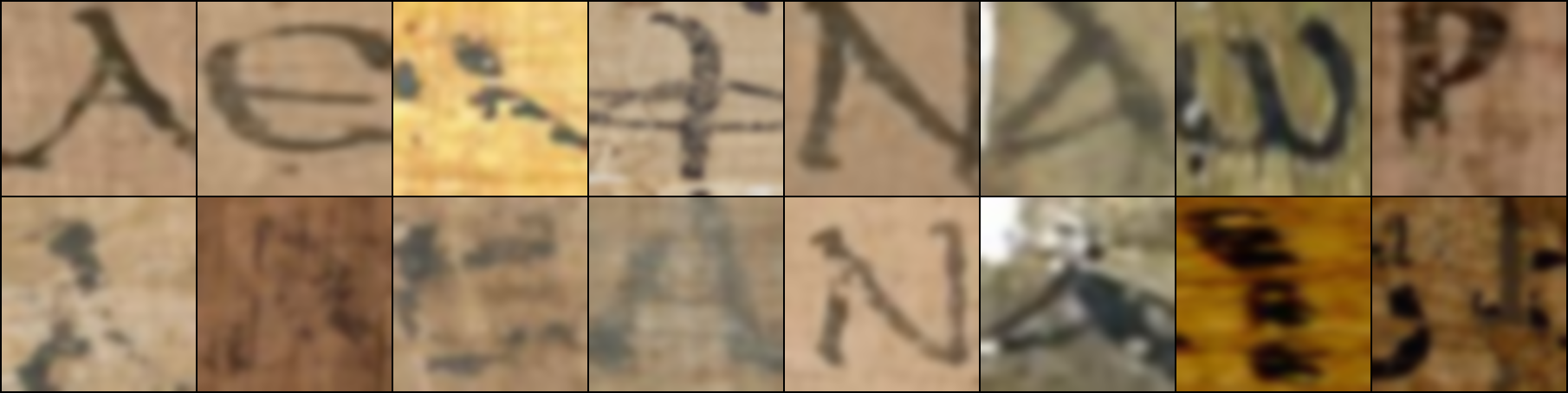}
        \caption{{\bf Gaussianblur} augmentation applied to the images, removing pixels from the boundaries using a $29\times29$ kernel.}
        \label{fig:gaussianblur_image}
    \end{subfigure}
    \vspace{0.5cm}
    \begin{subfigure}[b]{0.8\textwidth}
        \centering
         \includegraphics[width=\textwidth]{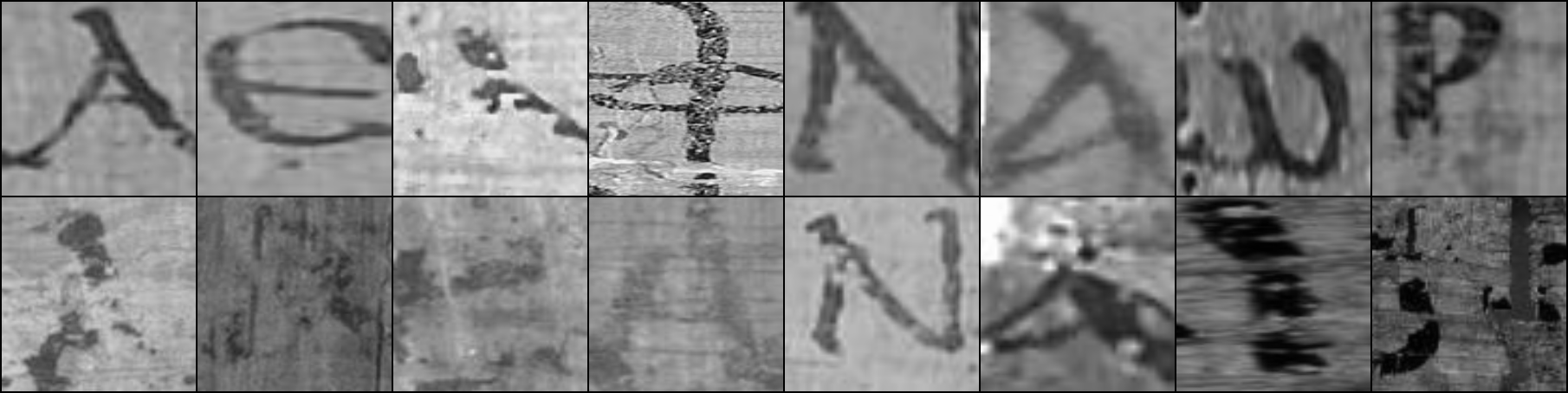}
        \caption{{\bf Grayscale} augmentation applied to convert the images to shades of gray.}
        \label{fig:gray_image}
     \end{subfigure}
    \caption{Visualizations of the {\bf pixel-level} augmentations applied in our experiments.The first block shows the original resized image at a $256 \times 256$ resolution, and the following blocks show the resulting visualizations from different pixel-level augmentation strategies.}
\end{figure}

\begin{itemize}
\item{\bf Colorjitter.} Color jitter (shown in image~\ref{fig:colorjitter_image}) adjusts the brightness, contrast, saturation, and hue of the image, with brightness and contrast varying within the range [0.8, 1], saturation within [0.8, 1], and hue within [-0.5, 0.5]. This augmentation increases the variability of color and lighting conditions in the dataset, helping the model to generalize across different environmental conditions and improving robustness against changes in color and illumination.

\item{\bf Gaussianblur.} Gaussian blur(~\ref{fig:gaussianblur_image}) applies a blurring effect to the image with a kernel size ranging from 3x3 to 7x7 pixels. This transformation smooths the image by averaging pixel values, which reduces sharpness and noise. By applying Gaussian blur, we help the model focus on more prominent features and improve its robustness to variations in image sharpness.

\item {\bf Invert.} The invert(we can see in image~\ref{fig:invert_image}) transformation flips the colors of the image, producing a negative of the original image. This enhancement emphasizes contrasts and highlights features that might otherwise be less visible. By inverting the colors, the model can learn to recognize features regardless of their color schemes, which can improve feature extraction and object detection performance.

\item {\bf Gray.} The grayscale, illustrated in image~\ref{fig:gray_image} transformation converts the image to shades of gray, removing color information and focusing solely on luminance. This reduction in color complexity helps the model to concentrate on structural and textural information. By applying this transformation, the model becomes better at analyzing and recognizing objects based on their shapes and textures rather than their colors.
\end{itemize}

{\noindent \bf Remark.} An important point to emphasize at the end of this augmentation section is that several augmentations, such as Gaussian blur, colorjitter, etc., have various associated internal hyperparameters. We have fixed these internal parameters after a manual inspection of the images, and we cannot guarantee that these are optimal settings. A thorough analysis of the internal hyperparameters for each algorithm is deferred to future work. In Table~\ref{table:albumentation_type}, we summarize our list of primary augmentation types and their internal hyperparameters. We provide the exact parameters used for each augmentation in the Table~\ref{tab:hyperparameters} in the Experiments.

\begin{table}[h!]
\centering
\begin{tabular}{llll}
\toprule
\textbf{Index} & \textbf{Augmentation} & \textbf{Type} & \textbf{Hyperparameters} \\
\midrule
\texttt{1} & \texttt{gray} & \texttt{Pixel-level} & \texttt{-}  \\
\texttt{2} & \texttt{invert} & \texttt{Pixel-level} & \texttt{-} \\
\texttt{3} & \texttt{gaussianblur} & \texttt{Pixel-level}  & \texttt{blur\_limit, sigma\_limit} \\
\texttt{4} & \texttt{colorjitter} & \texttt{Pixel-level} & \texttt{brightness, contrast,}\\
& & & \texttt{saturation, hue}  \\
 \hline
\texttt{5} & \texttt{resize256} & \texttt{Spatial-level} & \texttt{-} \\
\texttt{6} & \texttt{randomcrop224} & \texttt{Spatial-level}  & \texttt{-} \\
\texttt{7} & \texttt{hflip} & \texttt{Spatial-level} & \texttt{-} \\
\texttt{8} & \texttt{morpho\_dilation} & \texttt{Spatial-level} & \texttt{kernel}  \\
\texttt{9} & \texttt{morpho\_erosion} & \texttt{Spatial-level} & \texttt{kernel}  \\
\texttt{10} & \texttt{affine} & \texttt{Spatial-level}  & \texttt{shift\_limit, scale\_limit,} \\
&  &  &\texttt{rotate\_limit}\\

\bottomrule
\end{tabular}
\caption{{\bf Primary Augmentations.}}
\label{table:albumentation_type}
\end{table}

\clearpage
\section{Training Methods}
\label{sec:training_methods}
In this section, we will explain the three training methods: one baseline model and two embedding-based methods, namely Triplet and SimCLR. In short,  the baseline model serves as a reference point, providing a standard against which the performance of the embedding-based methods can be compared. The Triplet method focuses on learning by comparing anchor, positive, and negative examples to enhance the model's ability to distinguish between similar and dissimilar inputs. On the other hand, SimCLR leverages contrastive learning to maximize agreement between differently augmented views of the same data, further improving the model's robustness and generalization capabilities at the evaluation time.

\subsection{Baseline Model}
\label{sec:baseline_model}

We will discuss the core backbone architecture of ResNet~\cite{resnet} to understand the building blocks and how it encodes the input data into a final probability vector over the classes. By examining the ResNet architecture, we aim to provide a clear understanding of its structural components and the way it processes data to achieve accurate predictions. Let us now study the ResNet backbone from an architectural perspective.\\

\noindent \textbf{ResNet18 backbone:} We run all our experiments on the ResNet backbone~\cite{resnet}. ResNet18 is a CNN~\cite{cnn} designed for image recognition tasks and is part of the ResNet (Residual Network) family~\cite{resnet}, which introduces residual connections between the layer inputs and outputs. These residual blocks allow the network to learn residuals, or differences, between input and output, facilitating the training of deeper architectures, and solves the problem of vanishing graidents~\cite{deepresidualblock} in Neural networks. The network begins with an initial convolutional layer with a $7 \times 7$ kernel, producing an output feature map of size $256\times 256$ pixels, followed by a max-pooling layer that reduces the size to $128\times 128$ pixels. It then includes four residual stages:

\begin{itemize}
    \item {\bf Stage 1}: Contains 2 residual blocks, each with two $3\times3$ convolutional layers (64 filters), producing an output shape of $128\times128$ pixels.
    \item {\bf Stage 2}: Contains 2 residual blocks, each with two $3\times3$ convolutional layers (128 filters), with the output shape reduced to $64\times64$ pixels.
    \item {\bf Stage 3}: Contains 2 residual blocks, each with two $3\times3$ convolutional layers (256 filters), with the output shape reduced to $32\times32$ pixels.
    \item {\bf Stage 4}: Contains 2 residual blocks, each with two $3\times3$ convolutional layers (512 filters), with the output shape reduced to $16\times16$ pixels.
    \item {\bf Pooling}: The network concludes with a global average pooling layer that produces a feature vector of length 512. The layers up to this point are considered the backbone.
    \item {\bf FC layer}: The final feature vector is passed through a fully connected layer for classification.
\end{itemize}

\subsubsection{Data Augmentation}
In the baseline model, during training, each image is first resized to $256 \times 256$ pixels using the \texttt{Resize256} operation to standardize the input dimensions across all dataset images. Following the resizing operation, a series of augmentations, specified in a string format and split into a list of transform types, are applied. Apart from the 10 primary augmentations, a few additional transformations are applied by default to the training image. These include \texttt{CenterCrop}, which further crops the image to $224 \times 224$ pixels, followed by the \texttt{Normalization} operation, which adjusts the image pixel values to have a mean of \texttt{(0.485, 0.456, 0.406)} and a standard deviation of \texttt{(0.229, 0.224, 0.225)}.  For validation and test images, only resizing to $256 \times 256$ pixels, center cropping to $224 \times 224$ pixels, and normalization are applied. This ensures consistency in image dimensions and normalization while avoiding the introduction of additional variability that could affect model performance evaluation.\\

\subsubsection{Cross-Entropy Loss} 
We train ResNet-18 using cross-entropy loss~\cite{crossentropy} to guide the optimization process during model training. Formally, the CE loss is computed as follows:
\begin{equation}
    L = -\sum_{i=1}^{N} y_i \log(\hat{y}_i)
    \label{eq:ce_loss}
\end{equation}

where:
\begin{itemize}
    \item \( L \) is the loss value.
    \item \( N \) is the number of classes in the training dataset.
    \item \( y_i \) is the true label (1 if the class is the correct classification, 0 otherwise).
    \item \( \hat{y}_i \) is the predicted probability of class \( i \).
\end{itemize}

\noindent Cross-entropy loss evaluates the divergence between the true labels and the predicted probabilities. By penalizing higher deviations between predicted probabilities and actual labels, it ensures that the model learns to predict probabilities that are as close to the true labels as possible.

\begin{figure}
    \centering
    \includegraphics[width=\textwidth]{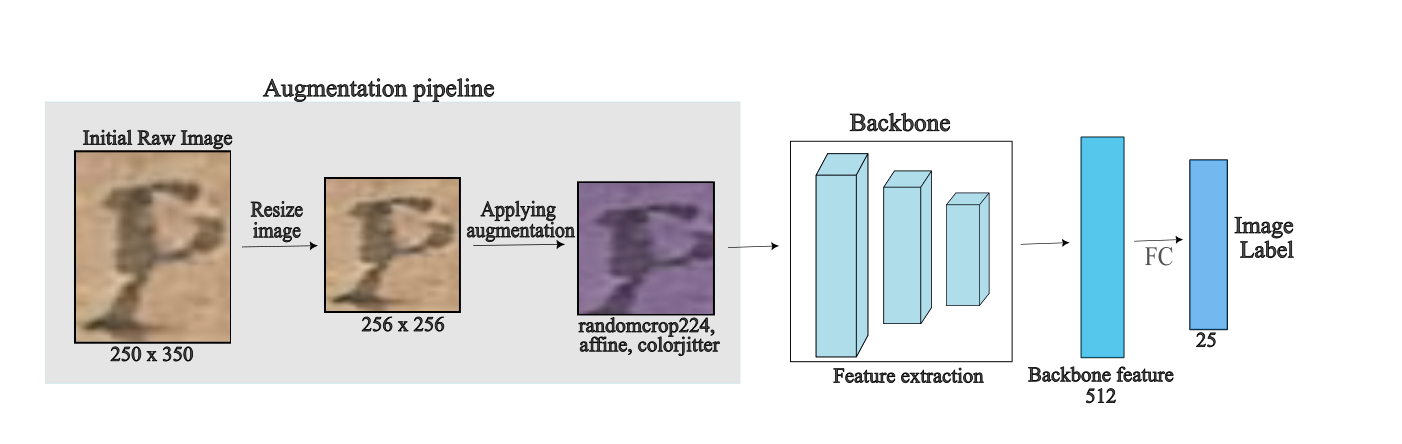}
    \caption{{\bf Baseline model pipeline}. Spatial (eg., \texttt{randomcrop224}) and pixel-level augmentations (eg., \texttt{colorjitter}) are first performed on the original raw image to obtain the final pre-processed image. The augmented image is then passed through the model, which extracts a feature vector at the end of the backbone. The final feature vector is sent to the classification layer to produce output probabilities.}
    \label{fig:baseline_pipeline}
\end{figure}

\subsubsection{Overall Training Pipeline}

As shown in Figure~\ref{fig:baseline_pipeline}, the baseline model process follows a straightforward sequence from input image to output class label:
\begin{enumerate}
    \item Each training image is resized to a standard dimension, typically $256 \times 256$ pixels. The resized image is then subjected to a series of augmentations, both spatial and pixel-level. We apply spatial augmentations with a probability of 1 but this only applies to \texttt{resize256} and \texttt{randomcrop224}, meaning all images in the batch undergo \texttt{randomcrop224}; the remaining primary spatial-level augmentations have a probability of $0.5$. However, pixel-level augmentations such as \texttt{color-jittering}, \texttt{blurring}, and \texttt{grayscale} are applied with a probability of $0.5$. This introduces variability and diversity into the training data, helping the model generalize better. 

    \item The augmented images are fed into ResNet-18, which extracts hierarchical features from the images through multiple layers of convolutions and residual blocks. At the end of the backbone, a globally averaged pooled feature vector is obtained. This feature vector is passed through a fully connected classification layer, which outputs class probabilities for each image. The class with the highest probability is chosen as the predicted label for that image. 

    \item To drive the optimization, we use the cross-entropy loss function, which evaluates the discrepancy between the predicted probabilities and the actual class labels. For each image in the batch, the cross-entropy loss is computed as shown in Equation~\ref{eq:ce_loss}. The cross-entropy loss for the entire batch is averaged from the losses of individual images. Using the computed loss, the optimizer(~\cite{adam}) performs backpropagation to update the model’s weights. 

    \item The model undergoes iterative training, processing multiple batches of images. During each iteration, the model learns from the computed losses and adjusts its weights accordingly. This iterative process continues until the model’s performance stabilizes. 
    
    \item After training, the model is evaluated on validation and test datasets. Performance metrics, such as accuracy, are computed to assess how well the model generalizes to new, unseen data.
\end{enumerate}

\subsection{Triplet Model}
\label{sec:triplet_model}

The Triplet Model~\cite{triplet} was integrated into our experiments to complement the baseline model by learning more discriminative embeddings and to provide a comparison with the baseline and SimCLR self-supervised models. It is particularly effective for tasks requiring fine-grained distinctions between classes, such as face recognition~\cite{schroff2015facenet, tripletfacerecognition} and person reidentification~\cite{personreidentification, defensetripletpersonreide}. The model is trained using the triplet loss, which helps in creating compact clusters of similar examples while pushing apart dissimilar ones.

\subsubsection{Architecture} 
We use ResNet as the backbone in the experiment. The final feature vector, which is the output of the layer just before the classification layer, is extracted. This feature vector is then fed into a shallow single layer to project the feature from its original dimension (512 dimension) to 64 dimensions, improving the efficiency of the training process. It is important to note that during the triplet pretraining stage, no classification layer is involved. However, during the fine-tuning stage, we add a classification layer on top of the projection layer to predict the label.

\begin{figure}[h]
    \centering
    \includegraphics[width=\textwidth]{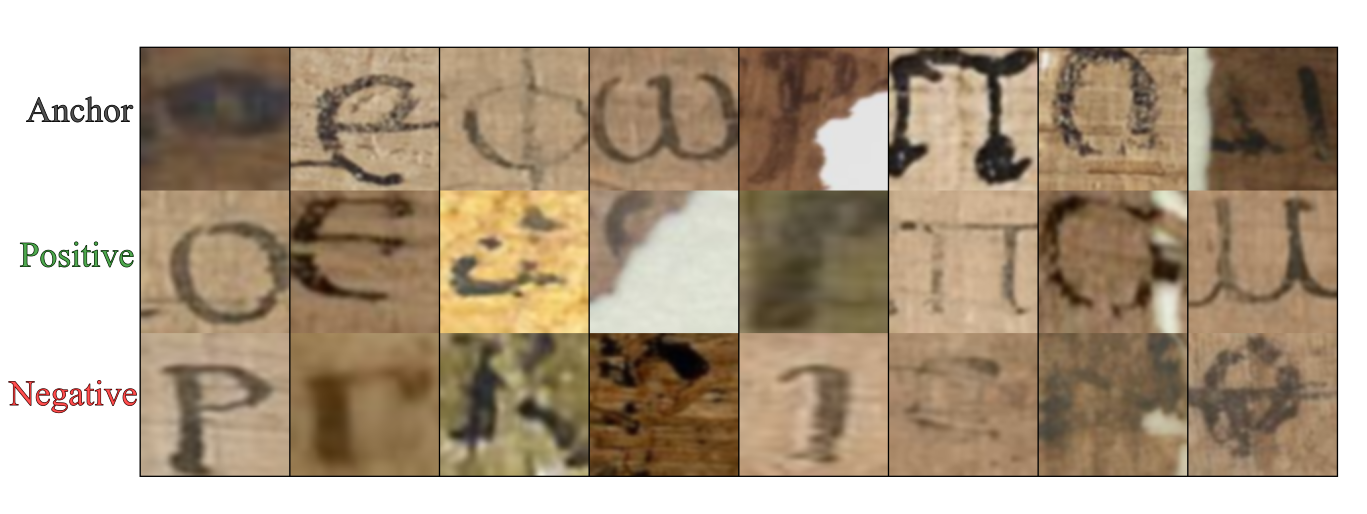}
    \caption{{\bf Batch of images used in Triplet model training.} The first row represents the {\bf anchors}, the second row represents the {\bf positives}, which contain images with the same label as the anchors, and the third row shows the {\bf negatives}, which are images with different labels from the anchors.}
    \label{fig:triplet_order2}
\end{figure}

\subsubsection{Triplet Loss}
The Triplet Loss model was integrated into our experiments to compare with the baseline model. The Triplet Embedding model is designed to train a neural network to generate embeddings where the distance between an anchor and a positive example (from the same class) is smaller than the distance between the anchor and a negative example (from a different class) by a defined margin, which is a hyperparameter. This loss function encourages the network to create tightly packed clusters of similar items while pushing dissimilar items apart. The goal is to minimize the distance between the anchor and positive images while maximizing the distance between the anchor and negative images. This approach is particularly beneficial for tasks such as face recognition, where it is essential to differentiate between similar and dissimilar identities. Formally, we calculate the triplet loss as follows:

\begin{equation}
L(\mathbf{x}_a, \mathbf{x}_p, \mathbf{x}_n) = \max \left(0, D(\mathbf{x}_a, \mathbf{x}_p) - D(\mathbf{x}_a, \mathbf{x}_n) + \alpha \right)
\label{eq:triplet_loss}
\end{equation}

where:
\begin{itemize}
    \item \( \mathbf{x}_a \) is the representation of the anchor instance.
    \item \( \mathbf{x}_p \) is the representation of the positive instance (same class as anchor).
    \item \( \mathbf{x}_n \) is the representation of the negative instance (different class from anchor).
    \item \( \alpha \) is the margin, a positive constant that ensures a minimum separation between positive and negative pairs.
    \item \( D(\mathbf{x}_a, \mathbf{x}_p) \) is the distance metric (often Euclidean distance) between the embeddings of \( \mathbf{x}_a \) and \( \mathbf{x}_p \), which should be minimized.
    \item \( D(\mathbf{x}_a, \mathbf{x}_n) \) is the distance between the anchor and negative examples, which should be maximized.
\end{itemize}

\noindent The term \( \max \left(0, D(\mathbf{x}_a, \mathbf{x}_p) - D(\mathbf{x}_a, \mathbf{x}_n) + \alpha \right) \) ensures that the loss is computed only when the distance between the anchor and the positive instance is not at least the margin \( \alpha \) smaller than the distance between the anchor and the negative instance. In other words, the loss is zero if the positive pair is at least \( \alpha \) closer to the anchor compared to the negative pair.

\begin{figure}
    \centering
    \includegraphics[width=\textwidth]{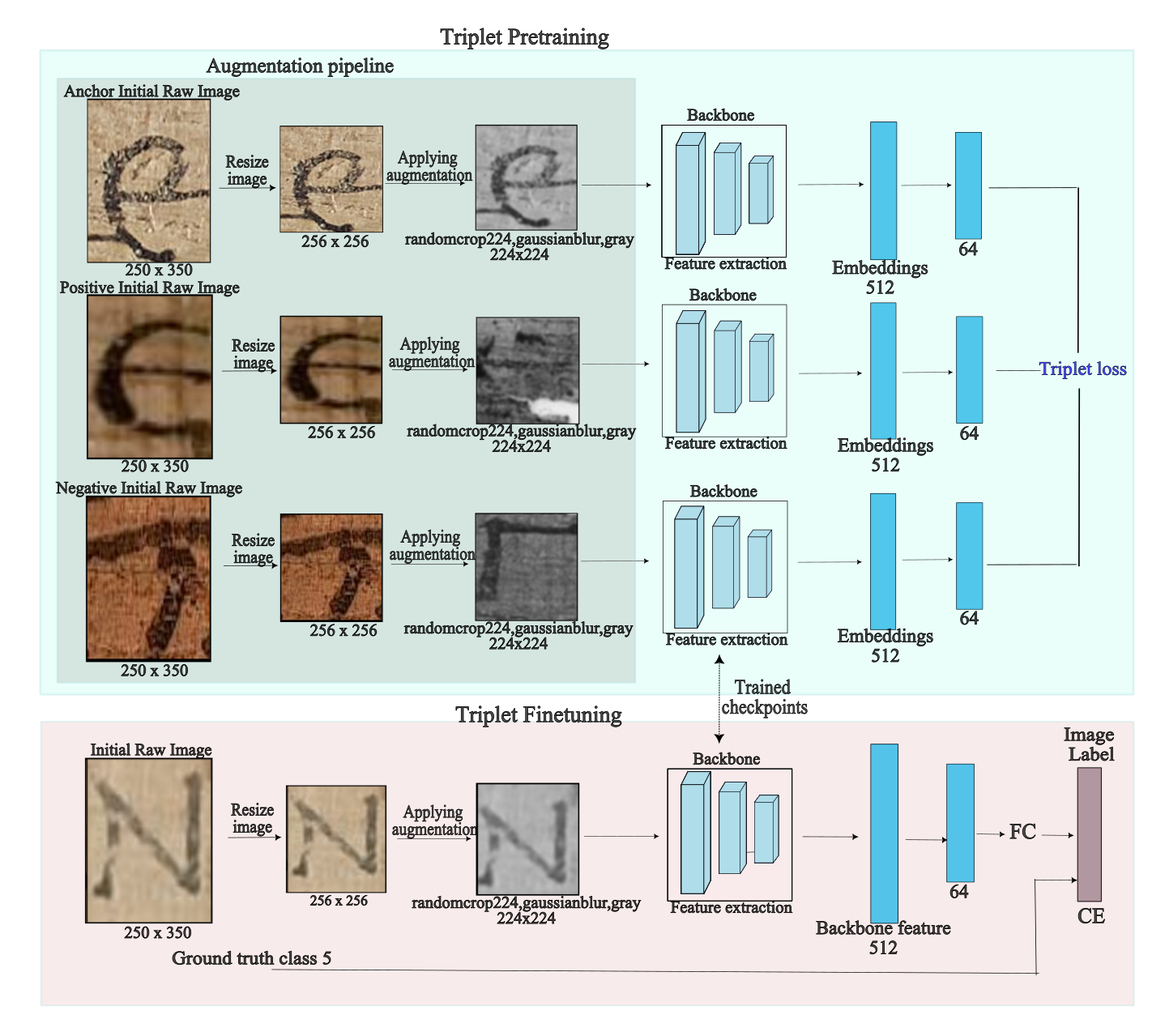}
    \caption{{\bf Triplet Model Pipeline.} The top block, highlighted with a green background, represents the pretraining stage, where the model is trained on the pretraining dataset using triplet loss to learn embeddings from triplet pairs. In the next stage, a classification layer is added on top of the embedding layer, and the model is trained end-to-end with cross-entropy (CE) loss.}
    \label{fig:triplet_pipeline}
\end{figure}

\subsubsection{Data Augmentation}

This follows a similar theme as in the baseline model. Here, we first resize images to $256 \times 256$ pixels and apply spatial augmentation by randomly cropping a $224 \times 224$ region and series of augmentations, followed by pixel-level augmentations such as \texttt{CenterCrop}, \texttt{Normalization}, etc. For validation and test images, we apply spatial-level augmentation like resizing to $256 \times 256$ pixels, then apply \texttt{CenterCrop} to $224 \times 224$ pixels, followed by \texttt{Normalization}. We do not apply pixel augmentations during the testing phase.\\

\noindent In triplet model training, the dataloader extracts a batch of \(N\) original Greek letter images, randomly selected from the training set. These images are considered as anchors. The dataloader also extracts another \(N\) images from the same class as the anchor images, which are considered positives, and another set of \(N\) images that have different labels from the original images, which are considered negatives. Overall, the resulting batch contains \(N \times 3\) images (see Figure \ref{fig:triplet_order2} for an example with a batch of 8 images). In short, each triplet consists of an anchor image, a positive image (another crop of the same original image as the anchor), and a negative image (a crop from a different original image). \\

\subsubsection{Overall Training Pipeline}

\begin{enumerate}

\item {\bf Pretraining Stage}
{
\begin{enumerate}
    \item As shown in the Figure~\ref{fig:triplet_pipeline} top block denoted within a green background, the triplet model operates based on three types of samples: anchor, positive, and negative. The model starts with a batch of images. Each batch contains multiple triplets, where each triplet is composed of three images: an anchor image, a positive image, and a negative image. For instance, if the batch size is 16, it will consist of several triplets where each triplet contains three different images. Before feeding the images into the model, they undergo a series of augmentations, such as cropping, color jittering, and blurring. These augmentations help in creating diverse views of the same image and improve the robustness of the learned embeddings.
    
 \item Each image in the batch (anchor, positive, and negative) is passed through a neural network, such as ResNet-18 or ResNet-50. This forward pass generates embeddings, which are vector representations of the images in a high-dimensional space. After passing through the network, each image is represented by an embedding vector. The embeddings for the anchor, positive, and negative images are obtained from the output of the network.

 \item The triplet loss function (Equation~\ref{eq:triplet_loss}) calculates the distances between the anchor-positive and anchor-negative embeddings. Typically, the Euclidean distance or cosine similarity is used to measure these distances. The loss function enforces that the distance between the anchor and positive embeddings should be smaller than the distance between the anchor and negative embeddings by at least \( \alpha \). If the distance condition is met, the loss is zero; otherwise, it penalizes the model based on how much the condition is violated.

  \item The computed loss is used to calculate gradients with respect to the model parameters. This involves backpropagation, where the gradients are propagated backward through the network to update the weights. The optimizer (e.g., Adam~\cite{adam} or SGD~\cite{sgd}) updates the network weights based on the gradients. This process adjusts the embeddings so that similar images (anchor and positive) are closer together, and dissimilar images (anchor and negative) are pushed further apart in the embedding space.
\end{enumerate}
}

\item {\bf Fine-tuning Stage}
{
\begin{enumerate}
    \item Once the embedding model is pre-trained using the triplet loss, it can be fine-tuned on a smaller, more specific dataset. During fine-tuning, the pre-trained embeddings are used as a starting point, and the model is further trained to adapt to the new dataset's class distribution. To this end, a classification layer is added on top of the embeddings to predict class labels. This layer uses the embeddings from the backbone to classify images into one of the predefined classes based on their proximity in the embedding space. We train this model end-to-end using the CE loss. Optionally, we can freeze the backbone network to minimize computational costs.

    \item Note that during this stage, we use the same augmentation that was applied during the pretraining stage. 
\end{enumerate}
}
\end{enumerate}

\subsection{SimCLR}
\label{sec:simclr_model}

\begin{figure}
    \centering
    \includegraphics[width=\textwidth]{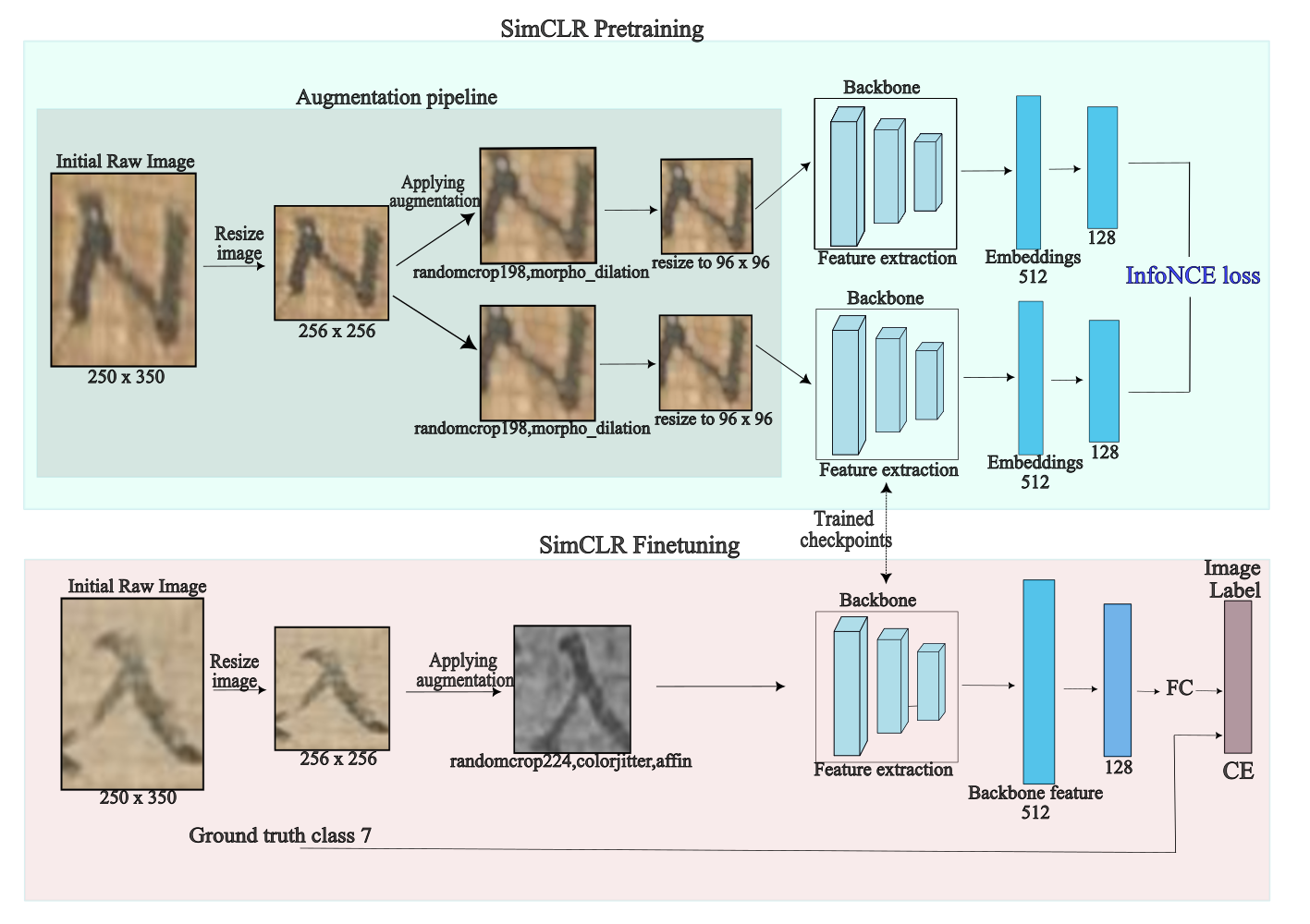}
    \caption{{\bf SimCLR Pipeline.} The top block, highlighted with a green background, represents the pretraining stage, where the model is trained on the pretraining dataset using InfoNCE loss to learn embeddings from SimCLR augmented views of the same image. In the next stage, a classification layer is added on top of the embedding layer, and the model is trained end-to-end with cross-entropy (CE) loss.}
    \label{fig:simclr_pipeline}
\end{figure}

We now discuss the third training method, SimCLR~\cite{simclr}, the main focus of our thesis. SimCLR~\cite{simclr}  is a self-supervised learning framework and one of the popular contrastive learning techniques in machine learning~\cite{CVcontrastive, CVsupervised, CVtian2020makes, CVwang2023cssl}. Unlike the baseline method in Section~\ref{sec:baseline_model} and triplet training method in Section~\ref{sec:triplet_model} which requires ground truth labels, SimCLR pretraining does not require image labels and hence can be pretrained on a massive corpus of data without expensive annotations. The goal of SimCLR is to learn meaningful embeddings from the raw images. To this end, the SimCLR model takes augmented views, i.e., cropped subimages, of the same image and pulls the embeddings of these augmented images closer in the latent space, while the embeddings coming from two different images are pushed apart. We illustrate the pipeline of SimCLR in Figure~\ref{fig:simclr_pipeline}, which contains the details about the pretraining on large corpus dataset  in the top block and the details about fine-tuning in the bottom block. We will now discuss the architecture of SimCLR below: \\

\noindent {\bf Architecture.}  The SimCLR architecture follows the same theme as the Triplet model architecture, where we focus on training embedding than the labels. Here is the overview:

\begin{itemize}
    \item \textbf{Backbone:} Typically based on ResNet, producing a feature vector of dimension $D$ which is 512 for ResNet18~\cite{resnet}.
    \item \textbf{Projection Head:} A multi-layer perceptron (MLP) that maps the backbone's feature vector (e.g., 512 dimensions in ResNet18~\cite{resnet}) to a lower-dimensional space (e.g., 128 dimensions).
    \item \textbf{Contrastive Loss Function:} The model is trained using a  InfoNCE loss function that pulls the embeddings of crops from the same image closer together while pushing apart embeddings of different images. 
\end{itemize}

\noindent \textbf{InfoNCE Loss:} In SimCLR, we use InfoNCE loss~\cite{ULbarlow1989unsupervised, ULghahramani2003unsupervised, simclr} (Information Noise Contrastive Estimation) to train the model. InfoNCE loss is designed to maximize the mutual information between different views or augmentations of the same data instance while minimizing the similarity between different data instances. This loss function leverages the concept of contrasting positive samples (augmentations of the same instance) against a set of negative samples (other instances). By doing so, it encourages the model to learn representations that bring similar instances closer together in the embedding space while pushing dissimilar instances apart.  In short, each image is assigned a label index corresponding to the position where the sister image, originating from the same parent image, is present in the batch. The feature is represented based on the normalized distance between the given image and all the other images in the batch. \\

Formally, the InfoNCE loss is calculated as follows:

\begin{equation}
L = -\sum_{i=1}^{N} \log \left(\frac{\exp\left(\frac{\text{sim}(\mathbf{x}_i, \mathbf{x}_i^+)}{\tau}\right)}
{ {\sum_{j=1}^{2N} } \mathbf{1}_{[j \neq i]} \exp\left(\frac{\text{sim}(\mathbf{x}_i, \mathbf{x}_j)}{\tau}\right)}\right)
\end{equation}
\label{eq:infonce_loss}

where:
\begin{enumerate}
    \item $L$ is the loss.
    \item $N$ is the number of positive pairs.
    \item $ \mathbf{x}_{i}$ and $ \mathbf{x}_i^{+}$ are the representations of the anchor and positive examples, respectively.
    \item $\text{sim}(\mathbf{x}_i, \mathbf{x}_j)$ is the similarity function (often the dot product or cosine similarity).
    \item \( \tau \) is a temperature parameter.
    \item $\mathbf{1}_{[j \neq i ]}$ is an indicator function that is 1 if \( j \neq i \), and 0 otherwise.
\end{enumerate}

\begin{figure}
    \centering
    \begin{subfigure}[b]{0.8\textwidth}
        \centering
        \includegraphics[width=\textwidth]{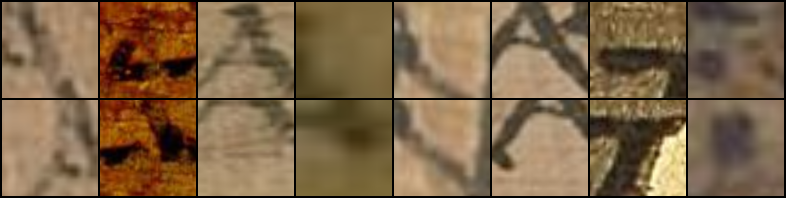}
        \caption{SimCLR cropping with {\bf 28\%} area of the original image area}
        \label{fig:simclr_28}
    \end{subfigure}
    \vspace{0.5cm}
    \begin{subfigure}[b]{0.8\textwidth}
        \centering
        \includegraphics[width=\textwidth]{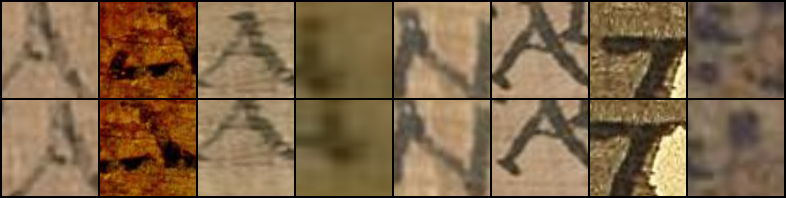}
        \caption{SimCLR cropping with {\bf 50\%} area of the original image area}
        \label{fig:simclr_50}
    \end{subfigure}
    \vspace{0.5cm}
    \begin{subfigure}[b]{0.8\textwidth}
        \centering
        \includegraphics[width=\textwidth]{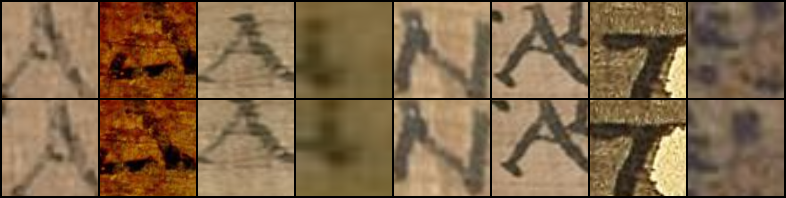}
        \caption{SimCLR cropping with {\bf 60\%} area of the original image area}
        \label{fig:simclr_60}
    \end{subfigure}
    \vfill
    \begin{subfigure}[b]{0.8\textwidth}
        \centering
        \includegraphics[width=\textwidth]{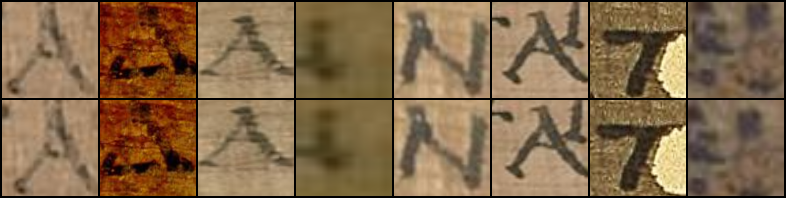}
        \caption{SimCLR cropping with {\bf 80\%} area of the original image area}
        \label{fig:simclr_80}
    \end{subfigure}
\caption{{\bf Visualizations of cropping image batch with varying areas.} We observe that cropping areas limited to 28\% altered the image substantially, whereas the other three areas, such as 50\%, 60\%, and 80\%, appear more reasonable. After consensus with the project team, we chose the 60\% area for our main experiments.}
\end{figure}

\subsubsection{Data Augmentations}

In the SimCLR model, images are initially resized to $256 \times 256$ pixels to standardize input dimensions, similar to earlier methods. Following this, a sequence of augmentations, including spatial and pixel-level augmentations specified as a list of transform types, is applied. After applying these augmentations, we resize the image to $96 \times 96$ pixels. This step is necessary because SimCLR requires larger batch sizes for convergence. However, due to memory constraints, we perform resizing to $96 \times 96$ pixels to manage memory usage effectively while still maintaining performance.\\

\noindent An important aspect when performing augmentations is the cropping size. After resizing the original image to $256\times 256$ pixels, we apply random cropping to ensure that at least 60\% of the original image is visible during augmentation. In Figure \ref{fig:simclr_60}, we observe a batch of 8 images with 60\% visibility, where the model evaluates images where the letter occupies at least 60\% of the original image area, corresponding to a size of $198 \times 198$ pixels. Additional examples of cropping with 50\%, 28\%, and 80\% visibility can be seen in Figures \ref{fig:simclr_50}, \ref{fig:simclr_28}, and \ref{fig:simclr_80}, respectively.  \\

\noindent {\bf Remark.} While it is challenging to determine the optimal cropping size, due to limited resources, we agreed to use 60\% of the original image after multiple rounds of discussion with the professor. \\

\subsubsection{Overall Training Pipeline}

Typically, SimCLR is trained in two stages. In the first stage, embeddings are learned on a large corpus of unlabeled data. In the second stage, supervised training is conducted on top of the embeddings learned from the first stage using an additional classification layer. It is important to note that one can also choose other methods, such as k-Nearest Neighbors (kNN), in the second stage. Let us now discuss the two stages in detail:

\begin{enumerate}
\item {\bf Pretraining Stage}{
\begin{enumerate}
\item In this stage, we train the network in a self-supervised manner on a large-scale dataset without the need for ground truth labels. The algorithm first resizes the original images to a uniform size of $256\times 256$ pixels. Following this, each image is cropped to $198 \times 198$ pixels and then subjected to various augmentations, including further cropping, color jittering, and blurring, to create {\bf two} diverse views of each image. 

\item The two cropped views form the positive pair and are then passed through the model to obtain the positive embeddings. The other images in the batch form the negative images, which are then passed through the model to get the negative embeddings.

\item We optimize the backbone network by minimizing the InfoNCE loss, ultimately learning a rich embedding space.
\end{enumerate}
}

\item {\bf Fine-tuning Stage}{

\begin{enumerate}
\item In the fine-tuning stage, we use the pre-trained backbone from the earlier pretraining stage and then fine-tune it on a small labeled dataset such as ICDAR~\cite{icdar} in our setting.

\item To perform supervised learning, we add a classification layer on top of the SimCLR backbone to get the probability distribution over the classes. This is a simple fully connected layer.

\item We train the full network, including the backbone, using cross-entropy loss. We will also show in the experiments that the full training of the backbone is essential since the SimCLR embeddigns are not quite discriminative even after training on the large corpus.
\end{enumerate}
}
\end{enumerate}

\noindent Overall, the SimCLR training leverages large publicly available datasets to learn rich embedding representations, which are then transferred to the target fine-tuning domain with small-scale supervised training.

\section{Summary}

In this chapter, we first studied the data augmentations that were explored for our experiments on the task of Greek-letter recognition. Specifically, we selected ten augmentations, comprising six spatial and four pixel-level augmentations. These augmentations were tailored to each training method.  \\

\noindent Next, we discussed three training methods: the baseline model using cross-entropy loss, the triplet model with triplet loss, and the SimCLR model with InfoNCE loss. While the baseline model learns directly from the labels, the other two models learn a rich embedding space based on the images in the batch. Moreover, the triplet model requires labels to prepare the training batch, whereas SimCLR pretraining does not depend on labels. In the second stage, both the triplet and SimCLR models undergo supervised fine-tuning by adding an additional classification layer. We will present the key results, highlighting the effectiveness of the augmentations and training methods, in the following chapter.

\chapter{Experiments} 
\label{Chapter4} 

In this chapter, we present the experimental results of Greek papyrus letter recognition using different methods. We begin by introducing the datasets and detailing the key experimental settings. Following this, we provide quantitative results for each method, supported by t-SNE ~\cite{tsne} visualizations of the embeddings.

\section{Experimental Setting}

We provide a detailed discussion of the key datasets, implementation details, training-time hyperparameters, and the various data augmentation strategies employed in our experiments. To begin, let's first examine the datasets used in this study.

\subsection{Datasets}
\label{datasets}

We conduct our experiments using two datasets: ALPUB~\cite{alpub} and ICDAR~\cite{icdar}. \\

\begin{enumerate}
    \item 
\noindent{\bf Pretraining Dataset.} We use the ALPUB~\cite{alpub} dataset for the pretraining stage. This dataset comprises a comprehensive collection of ancient Greek papyrus fragments, featuring a wide variety of handwritten letters. It contains 24 distinct letter classes, making it an invaluable resource for pretraining unsupervised models for Greek character recognition. The dataset includes a total of 205,797 cropped Greek letter images. We utilize this dataset in the pretraining phase for the Baseline~\cite{resnet, crossentropy}, Triplet~\cite{triplet}, and SimCLR~\cite{triplet} embedding models. %

\item \noindent{\bf Finetuning Dataset.} The ICDAR dataset is used for finetuning. This dataset, which was part of the ICDAR 2023 competition focused on the detection and recognition of Greek letters on papyri~\cite{icdar}, consists of 34,061 cropped images. We split the dataset into training, validation, and testing sets in a 70\%, 15\%, and 15\% ratio, resulting in 23,842 images for training, 5,109 for validation, and 5,110 for testing. It is important to note that the ICDAR dataset~\cite{icdar} includes an additional class compared to the ALPUB dataset~\cite{alpub}, bringing the total to 25 classes. Furthermore, the original dataset provided full images, which we cropped ourselves using ground-truth annotations provided by the authors. \\

\end{enumerate}

\subsection{Implementation Details}

Our experiments were conducted on a setup consisting of four NVIDIA GPUs, each with 9GB of RAM. For hyperparameter tuning, we utilized RayTune~\cite{raytune}, a robust library that efficiently explores the hyperparameter space. The experiments were implemented using the PyTorch library~\cite{pytorch}, which provides a flexible framework for deep learning tasks. Additionally, we leveraged the Albumentations library~\cite{albumentations} to generate a diverse set of augmentations, ensuring that our models were exposed to a wide range of data variations. Below, we discuss the key experimental parameters used at different stages for each method. \\

\noindent{\bf Pretraining Stage.} During the pretraining phase, we used the Adam optimizer~\cite{adam} with a learning rate of 0.001 for both the Triplet embedding model and the Baseline model. A learning rate scheduler with a gamma of 0.1 was employed, and the models were trained for 20 epochs {and temperature values used for triplet 1.0}. For training the SimCLR embedding model, we also used the Adam optimizer~\cite{adam}, but with a slightly lower learning rate of 0.0003, and temperature value is 0.07. The SimCLR model was trained for 100 epochs, utilizing a Cosine Annealing scheduler~\cite{cosinesgdr} to gradually reduce the learning rate throughout the training process. This careful adjustment of the learning rate was intended to help the model converge more effectively. \\

\noindent{\bf Finetuning Stage.} In the finetuning stage, we transitioned to training our models on the ICDAR dataset~\cite{icdar}. For this phase, we added a classification layer to the embedding models (Triplet and SimCLR) and finetuned them by either adjusting only the last layer or the entire model. All models during this stage were trained using the Adam optimizer~\cite{adam} with a learning rate of 0.001, and a learning rate scheduler was employed over 20 epochs. Due to the constraints of our computational resources, we did not conduct extensive tuning of the optimizer hyperparameters, opting instead to focus on ensuring that the models were sufficiently trained under the given conditions. \\

\subsection{Methods}

In this study, we evaluate the performance of Greek letter recognition using three different training strategies during the pretraining stage:

\begin{enumerate}
    \item {\bf Baseline Model:} The baseline model is trained using the Cross-entropy loss~\cite{crossentropy} function, which relies on ground-truth labels to guide the learning process. This method serves as a standard supervised learning approach. Detailed training procedures for this method are provided in Section~\ref{sec:baseline_model}.

    \item {\bf Triplet Embedding Model:} This model utilizes the Triplet loss function, where triplets of images (anchor, positive, and negative) are selected based on ground-truth labels to learn discriminative embeddings. As a result, the pretraining phase for this model requires access to labeled data from the pretraining dataset. Complete training details for this method are available in Section~\ref{sec:triplet_model}.

    \item {\bf SimCLR Model:} The SimCLR model is trained using the InfoNCE loss in a completely unsupervised manner, meaning it does not require access to image labels. This method leverages contrastive learning to learn robust feature representations. We outline the full training details for this method in Section~\ref{sec:simclr_model}.
\end{enumerate}

\noindent In the finetuning stage, we add an additional classification layer to all the models and retrain them using Cross-entropy loss. This final step allows us to benchmark the performance of each model on the ICDAR dataset, providing a fair comparison across different training strategies.

\subsection{Data Augmentations}
\label{93_augmentation}

Our experiments utilize a pool of 93 augmentations, consisting of various combinations of primary augmentation techniques, as discussed in Section~\ref{sec:primary_augmentations}. We create augmentation pipelines by combining two, three, or four primary augmentations for each image. For experiments conducted without pretraining on the large-scale ALPUB dataset, we evaluated all 93 augmentations. However, when pretraining on ALPUB was involved, we focused on the \texttt{top-4} augmentations identified from our initial experiments without pretraining. During training, both spatial augmentations and {pixel}-level augmentations are applied to enhance model generalization. For testing, we report the results without applying any additional augmentations, ensuring that the evaluation reflects the model's performance on unaltered data. We provide the exact values for hyperparameters in Table~\ref{tab:hyperparameters}. Note that, we selected these values after careful manual inspection so that the augmented image retains same semantic label meaning.\\

\begin{table}[h!]
\centering
\begin{tabular}{llll}
\toprule
\textbf{Index} & \textbf{Augmentation} & \textbf{Type} & \textbf{Hyperparameters} \\
\midrule
\texttt{1} & \texttt{invert} & \texttt{Pixel-level} & \texttt{-} \\
\texttt{2} & \texttt{gray} & \texttt{Pixel-level} & \texttt{-}  \\
\texttt{3} & \texttt{gaussianblur} & \texttt{Pixel-level} & \texttt{blur limit = (3, 7),}  \\
& & & \texttt{sigma limit = 0}\\
\texttt{4} & \texttt{colorjitter} & \texttt{Pixel-level} & \texttt{brightness = (0.8, 1),} \\
 & & & \texttt{saturation = (0.8, 1),}  \\
 & & & \texttt{contrast = (0.8, 1),}\\
 & & & \texttt{hue = (-0.5, 0.5)}\\
 \hline
\texttt{5} & \texttt{resize256} & \texttt{Spatial-level} & \texttt{-} \\
\texttt{6} & \texttt{randomcrop224} & \texttt{Spatial-level} & \texttt{-}  \\
\texttt{7} & \texttt{hflip} & \texttt{Spatial-level} & \texttt{-} \\
\texttt{8} & \texttt{morpho\_dilation} & \texttt{Spatial-level} & \texttt{kernel(w, h) = (7, 7)}  \\
\texttt{9} & \texttt{morpho\_erosion} & \texttt{Spatial-level} & \texttt{kernel(w, h) = (7, 7)}  \\
\texttt{10} & \texttt{affine} & \texttt{Spatial-level} & \texttt{shift\_limit = 0.05,} \\
& & & \texttt{scale\_limit = 0.1.}\\
& & & \texttt{rotate\_limit = 30}  \\

\bottomrule
\end{tabular}
\caption{{\bf Hyperparameters of Primary Augmentations}}
\label{tab:hyperparameters}
\end{table}

\begin{itemize}
    \item \noindent {\bf First Order Combinations.} Each combination includes only one augmentation applied to the base "randomcrop224". This augmentation test the effect of individual augmentations on the dataset. It serves as a baseline with no additional augmentations applied. We show the augmentation in  Table \ref{table:first_order} for the sake of the clarity.\\

    \item  \noindent{\bf Second Order Combinations.} Table \ref{table:second_order} lists all possible combinations of two augmentations applied to randomcrop224. This is useful for understanding how pairs of augmentations interact with each other.\\

    \item \noindent{\bf Third Order Combinations} Table \ref{table:third_order} covers all combinations of three augmentations. This helps in examining how combinations of three augmentations affect the dataset and interact with each other.\\

\item \noindent{\bf Fourth Order Combinations} This table \ref{table:fourth_order} includes all combinations of four augmentations applied together. It helps to understand the combined effect of multiple augmentations and their interactions.

\end{itemize}

\begin{table}[h!]
\centering
\begin{tabular}{ll}
\toprule
\textbf{Index} & \textbf{Augmentation} \\
\midrule
1 & randomcrop224 \\
\bottomrule
\end{tabular}
\caption{{\bf First-Order Augmentation.} This baseline augmentation randomly crops a $224\times 224$ region from a resized $256\times 256$ image.}
\label{table:first_order}
\end{table}

\begin{table}[h!]
\centering
\begin{tabular}{ll}
\toprule
\textbf{Index} & \textbf{Augmentation} \\
\midrule

1 & randomcrop224,morpho\_erosion \\
2 & randomcrop224,morpho\_dilation \\
3 & randomcrop224,affine \\
4 & randomcrop224,colorjitter \\
5 & randomcrop224,hflip \\
6 & randomcrop224,invert \\
7 & randomcrop224,gaussianblur \\
8 & randomcrop224,gray \\
\bottomrule
\end{tabular}
\caption{{\bf Second-Order Combinations.} This augmentation first performs the baseline step of randomly cropping a $224\times 224$ region from a resized $256\times 256$ image, followed by the application of additional augmentations.}

\label{table:second_order}
\end{table}

\begin{longtable}{ll}
\hline
\textbf{Index} & \textbf{Augmentation} \\
\hline
\endfirsthead
\hline
\textbf{Index} & \textbf{Augmentation} \\
\hline
\endhead
1 & randomcrop224,morpho\_erosion,morpho\_dilation \\
2 & randomcrop224,morpho\_erosion,affine \\
3 & randomcrop224,morpho\_erosion,colorjitter \\
4 & randomcrop224,morpho\_erosion,hflip \\
5 & randomcrop224,morpho\_erosion,invert \\
6 & randomcrop224,morpho\_erosion,gaussianblur \\
7 & randomcrop224,morpho\_erosion,gray \\
8 & randomcrop224,morpho\_dilation,affine \\
9 & randomcrop224,morpho\_dilation,colorjitter \\
10 & randomcrop224,morpho\_dilation,hflip \\
11 & randomcrop224,morpho\_dilation,invert \\
12 & randomcrop224,morpho\_dilation,gaussianblur \\
13 & randomcrop224,morpho\_dilation,gray \\
14 & randomcrop224,affine,colorjitter \\
15 & randomcrop224,affine,hflip \\
16 & randomcrop224,affine,invert \\
17 & randomcrop224,affine,gaussianblur \\
18 & randomcrop224,affine,gray \\
19 & randomcrop224,colorjitter,hflip \\
20 & randomcrop224,colorjitter,invert \\
21 & randomcrop224,colorjitter,gaussianblur \\
22 & randomcrop224,colorjitter,gray \\
23 & randomcrop224,hflip,invert \\
24 & randomcrop224,hflip,gaussianblur \\
25 & randomcrop224,hflip,gray \\
26 & randomcrop224,invert,gaussianblur \\
27 & randomcrop224,invert,gray \\
28 & randomcrop224,gaussianblur,gray \\
\bottomrule
\caption{{\bf Third-Order Combinations.} This augmentation begins with the baseline step of randomly cropping a $224\times 224$ region from a resized $256\times 256$ image, followed by the application of two additional augmentations.}
\label{table:third_order}
\end{longtable}

\begin{longtable}{ll}
\hline
\textbf{Index} & \textbf{Augmentation} \\
\hline
\endfirsthead
\hline
\textbf{Index} & \textbf{Augmentation} \\
\hline
\endhead

1 & randomcrop224,morpho\_erosion,morpho\_dilation,affine \\
2 & randomcrop224,morpho\_erosion,morpho\_dilation,colorjitter \\
3 & randomcrop224,morpho\_erosion,morpho\_dilation,hflip \\
4 & randomcrop224,morpho\_erosion,morpho\_dilation,invert \\
5 & randomcrop224,morpho\_erosion,morpho\_dilation,gaussianblur \\
6 & randomcrop224,morpho\_erosion,morpho\_dilation,gray \\
7 & randomcrop224,morpho\_erosion,affine,colorjitter \\
8 & randomcrop224,morpho\_erosion,affine,hflip \\
9 & randomcrop224,morpho\_erosion,affine,invert \\
10 & randomcrop224,morpho\_erosion,affine,gaussianblur \\
11 & randomcrop224,morpho\_erosion,affine,gray \\
12 & randomcrop224,morpho\_erosion,colorjitter,hflip \\
13 & randomcrop224,morpho\_erosion,colorjitter,invert \\
14 & randomcrop224,morpho\_erosion,colorjitter,gaussianblur \\
15 & randomcrop224,morpho\_erosion,colorjitter,gray \\
16 & randomcrop224,morpho\_erosion,hflip,invert \\
17 & randomcrop224,morpho\_erosion,hflip,gaussianblur \\
18 & randomcrop224,morpho\_erosion,hflip,gray \\
19 & randomcrop224,morpho\_erosion,invert,gaussianblur \\
20 & randomcrop224,morpho\_erosion,invert,gray \\
21 & randomcrop224,morpho\_erosion,gaussianblur,gray \\
22 & randomcrop224,morpho\_dilation,affine,colorjitter \\
23 & randomcrop224,morpho\_dilation,affine,hflip \\
24 & randomcrop224,morpho\_dilation,affine,invert \\
25 & randomcrop224,morpho\_dilation,affine,gaussianblur \\
26 & randomcrop224,morpho\_dilation,affine,gray \\
27 & randomcrop224,morpho\_dilation,colorjitter,hflip \\
28 & randomcrop224,morpho\_dilation,colorjitter,invert \\
29 & randomcrop224,morpho\_dilation,colorjitter,gaussianblur \\
30 & randomcrop224,morpho\_dilation,colorjitter,gray \\
31 & randomcrop224,morpho\_dilation,hflip,invert \\
32 & randomcrop224,morpho\_dilation,hflip,gaussianblur \\
33 & randomcrop224,morpho\_dilation,hflip,gray \\
34 & randomcrop224,morpho\_dilation,invert,gaussianblur \\
35 & randomcrop224,morpho\_dilation,invert,gray \\
36 & randomcrop224,morpho\_dilation,gaussianblur,gray \\
37 & randomcrop224,affine,colorjitter,hflip \\
38 & randomcrop224,affine,colorjitter,invert \\
39 & randomcrop224,affine,colorjitter,gaussianblur \\
40 & randomcrop224,affine,colorjitter,gray \\
41 & randomcrop224,affine,hflip,invert \\
42 & randomcrop224,affine,hflip,gaussianblur \\
43 & randomcrop224,affine,hflip,gray \\
44 & randomcrop224,affine,invert,gaussianblur \\
45 & randomcrop224,affine,invert,gray \\
46 & randomcrop224,affine,gaussianblur,gray \\
47 & randomcrop224,colorjitter,hflip,invert \\
48 & randomcrop224,colorjitter,hflip,gaussianblur \\
49 & randomcrop224,colorjitter,hflip,gray \\
50 & randomcrop224,colorjitter,invert,gaussianblur \\
51 & randomcrop224,colorjitter,invert,gray \\
52 & randomcrop224,colorjitter,gaussianblur,gray \\
53 & randomcrop224,hflip,invert,gaussianblur \\
54 & randomcrop224,hflip,invert,gray \\
55 & randomcrop224,hflip,gaussianblur,gray \\
56 & randomcrop224,invert,gaussianblur,gray \\
\hline
\caption{{\bf Fourth-Order Combinations.} This augmentation starts with the baseline step of randomly cropping a $224\times 224$ region from a resized $256\times 256$ image, followed by the application of three additional augmentations.}
\label{table:fourth_order}
\end{longtable}

\section{Results}
\label{results}

In this section, we present and discuss the experimental results obtained from the three different methods explored in this study. Our analysis is divided into two main parts to provide a clear comparison of the impact of pretraining on the large-scale ALPUB dataset.\\

\noindent We first delve into the results obtained without any pretraining on the ALPUB dataset, which are discussed in detail in Section~\ref{sec:result_wo_pretraining}. This analysis allows us to understand how each method performs when trained directly on the ICDAR dataset without leveraging additional pretraining data. The performance of these models in this setting provides a baseline for comparison, highlighting the strengths and limitations of each approach when operating in a more constrained environment with potentially less data diversity.\\

\noindent Subsequently, in Section~\ref{sec:result_w_pretraining}, we shift our focus to the results achieved after pretraining on the ALPUB dataset. This phase involves first pretraining the models on the extensive ALPUB dataset to learn robust feature representations, followed by finetuning on the more specific ICDAR dataset. By comparing these results with those from the previous section, we aim to elucidate the benefits and potential improvements that pretraining on a large and diverse dataset like ALPUB can bring to the task of Greek letter recognition.\\

\noindent Overall, this two-part analysis will provide a thorough understanding of how pretraining influences model performance and help identify the most effective strategies for enhancing Greek letter recognition accuracy.

\subsection{Results without Pretraining on Alpub}\label{sec:result_wo_pretraining}

In this section, we present the results obtained from direct training on the ICDAR dataset~\cite{icdar} using the three methods under consideration: the Baseline model, the Triplet Embedding model, and the SimCLR model. The results for ResNet-18~\cite{resnet} and ResNet-50~\cite{resnet} architectures are summarized in Table~\ref{sec:baseline_model},~\ref{sec:triplet_model} and Table~\ref{sec:simclr_model}, respectively.\\

\noindent We conducted experiments using 93 different data augmentations (as detailed in Tables~\ref{table:first_order}, \ref{table:second_order}, \ref{table:third_order}, \ref{table:fourth_order}) for each method. We distill the key results and report only the best-performing augmentation for each method in this section, and defer the complete results to the Tables(\ref{tab:baseline_ce_augmentation_results_18}, ~\ref{tab:ce_18_alpub}, ~\ref{tab:triplet_18_with_backbone_pretrain_icdar_finetune_icdar}, ~\ref{tab:triplet_18_with_backbone_pretrain_alpub_finetune_icdar}, ~\ref{tab:simclr_18_with_backbone_pretrain_alpub_finetune_icdar}, ~\ref{tab:simclr_18_with_backbone_pretrain_icdar_finetune_icdar}) and (\ref{tab:baseline_ce_augmentation_results_50}, ~\ref{tab:ce_50_alpub}, ~\ref{tab:triplet_50_with_backbone_pretrain_icdar_finetune_icdar}, ~\ref{tab:triplet_50_with_backbone_pretrain_alpub_finetune_icdar} ~\ref{tab:simclr_50_pretrain_alpub_finetune_icdar_with_backbone}, ~\ref{tab:simclr_50_with_backbone_pretrain_icdar_finetune_icdar}) in the Appendix. \\

\noindent {\bf Key finding.} Our findings reveal that the {\bf Baseline model} trained with cross-entropy loss consistently achieved the highest test accuracy, reaching {\bf 80.67\%} on ResNet-18~\cite{resnet}, and slightly lower at {\bf 80.47\%} on ResNet-50~\cite{resnet}. These results indicate that the cross-entropy-based Baseline model outperformed both the embedding-based methods across both architectures.\\

\noindent In comparison, the Triplet and SimCLR models achieved lower accuracies, with the Triplet model reaching 78.22\% and the SimCLR model achieving 79.24\% on ResNet-50. This trend highlights the robustness of the Baseline model in this direct training scenario. However, it's important to note that the best-performing augmentation strategy varied between methods, with the more complex fourth-order augmentations often leading to better performance. Specifically, the augmentation pipeline involving \texttt{randomcrop224}, \texttt{morpho\_erosion}, \texttt{morpho\_dilation}, \texttt{gaussianblur} stood out, achieving the highest accuracy in two out of the six experiments (ie., six rows) conducted across the two architectures.\\

\noindent These results underscore the significance of selecting the appropriate combination of augmentations tailored to each method. The success of more advanced augmentation strategies, such as third and fourth-order combinations that integrate multiple morphological transformations and pixel-level operations, suggests that careful tuning is essential for maximizing model performance. The diversity and complexity of these augmentations appear to be particularly beneficial, likely by enhancing the model's ability to generalize across varied data conditions. Overall, these findings highlight the need for a nuanced approach to data augmentation, where the choice of strategy is method-specific and can significantly influence the outcome of the training process.

\begin{table}[h!]

\small
\renewcommand{\arraystretch}{3} %
\begin{tabular}{|p{3cm}|p{1.5cm}|p{4cm}|p{1.4cm}|p{1.4cm}|}

\hline
\textbf{Experiment} & \textbf{Dataset} & \textbf{Best augmentation} & \textbf{Valid Acc.} & \textbf{Test Acc.} \\
\hline
Baseline model & \texttt{ICDAR} & \parbox{4cm}{\texttt{randomcrop224, morpho\_erosion, morpho\_dilation, gaussianblur}} & 81.19\% & \bf 80.67\% \\ \hline 

Triplet model & \texttt{ICDAR} & \parbox{4cm}{\texttt{randomcrop224, morpho\_dilation, affine, colorjitter}} & 80.11\% & 79.16\% \\ \hline

SimCLR model & \texttt{ICDAR} & \parbox{4cm}{\texttt{randomcrop224, affine, colorjitter, gray}} & 80.33\% & 80.00\% \\

\hline
\end{tabular}
\caption{{\bf Results on ResNet-18 without pretraining on Alpub dataset.}  We report the best found augmentation and their corresponding validation and test set accuracies by directly finetuning on ICDAR. We observe the baseline model achieves the best results than other two methods.}
\label{tab:resnet18_icdar_results}

\label{tab:resnet18_icdar_results}
\end{table}

\begin{table}[h!]
\centering
\renewcommand{\arraystretch}{3} %
\begin{tabular}{|p{3cm}|p{1.5cm}|p{4cm}|p{1.4cm}|p{1.4cm}|}

\hline
\textbf{Experiment} & \textbf{Dataset} & \textbf{Best augmentation} & \textbf{Valid Acc.} & \textbf{Test Acc.} \\
\hline
Baseline model & \texttt{ICDAR} & \parbox{4cm}{\texttt{randomcrop224, morpho\_erosion, gaussianblur}} & 80.70\% & \bf 80.47\% \\ \hline 

Triplet model & \texttt{ICDAR} & \parbox{4cm}{\texttt{randomcrop224, morpho\_erosion, morpho\_dilation, gaussianblur}} & 79.29\% & 78.22\% \\ \hline

SimCLR model & \texttt{ICDAR} & \parbox{4cm}{\texttt{randomcrop224, colorjitter, gaussianblur}} & 80.05\% & 79.24\% \\

\hline
\end{tabular}
\caption{{\bf Results on ResNet-50 without pretraining on Alpub dataset.}  We report the best found augmentation and their corresponding validation and test set accuracies. We observe the baseline model achieves the best results than other two methods. }
\label{tab:resnet50_icdar_results}
\end{table}

\subsection{Results with Pretraining}\label{sec:result_w_pretraining}

\begin{table}[h!]
\centering
\renewcommand{\arraystretch}{3} %
\begin{tabular}{|p{3cm}|p{1.8cm}|p{4cm}|p{1.4cm}|p{1.4cm}|}

\hline
\textbf{Experiment} & \textbf{Dataset} & \textbf{Best augmentation} & \textbf{Valid Acc.} & \textbf{Test Acc.} \\
\hline
Baseline model & \parbox{1.8cm} {\texttt{Alpub +  ICDAR}} & \parbox{4cm}{\texttt{randomcrop224, hflip, gray}} & 80.49\% & \bf 79.94\% \\ \hline 

Triplet model & \parbox{1.8cm} {\texttt{Alpub +  ICDAR}} & \parbox{4cm}{\texttt{randomcrop224, morpho\_dilation, hflip}} & 78.19\% & 77.51\% \\ \hline

SimCLR model & \parbox{1.8cm} {\texttt{Alpub +  ICDAR}} & \parbox{4cm}{\texttt{randomcrop224, colorjitter, hflip, invert}} & 77.55\% & 76.14\% \\

\hline
\end{tabular}
\caption{{\bf Results on ResNet-18 with pretraining on Alpub dataset (with top-4 selected using strategy 1).}  We report the best found augmentation and their corresponding validation and test set accuracies. We observe the baseline model achieves the best results than other two methods. }
\label{tab:resnet18_Alpub_results}
\end{table}

\begin{table}[h!]
\centering
\renewcommand{\arraystretch}{3} %
\begin{tabular}{|p{3cm}|p{1.8cm}|p{4cm}|p{1.4cm}|p{1.4cm}|}

\hline
\textbf{Experiment} & \textbf{Dataset} & \textbf{Best augmentation} & \textbf{Valid Acc.} & \textbf{Test Acc.} \\
\hline
Baseline model & \parbox{1.8cm} {\texttt{Alpub +  ICDAR}} & \parbox{4cm}{\texttt{randomcrop224, morpho\_dilation, hflip}} & 80.21\% & \bf 79.75\% \\ \hline 

Triplet model & \parbox{1.8cm} {\texttt{Alpub +  ICDAR}} & \parbox{4cm}{\texttt{randomcrop224, invert, gaussianblur, gray}} & 77.90\% & 77.03\% \\ \hline

SimCLR model & \parbox{1.8cm} {\texttt{Alpub +  ICDAR}} & \parbox{4cm}{\texttt{randomcrop224, invert, gaussianblur, gray}} & 76.90\% & 76.59\% \\

\hline
\end{tabular}
\caption{{\bf Results on ResNet-50 with pretraining on Alpub dataset (with top-4 selected using strategy 1)}  We report the best found augmentation and their corresponding validation and test set accuracies. We observe the baseline model achieves the best results than other two methods.}
\label{tab:resnet50_Alpub_results}
\end{table}

This section discusses the core results that are central to this thesis: specifically, how pretraining on a large-scale dataset impacts the performance of contrastive learning techniques. We will address this question with empirical evidence, analyzing the effects of pretraining on model performance in the following paragraphs. \\

\noindent{\bf Selecting Top-4 Augmentations for Pretraining on the ALPUB Dataset.} Before presenting the results of pretraining on the ALPUB dataset using different augmentations, we encountered significant computational bottlenecks. Given these constraints of limited computational resources, we selected only a few top-performing augmentations from our earlier experiment, which involved direct finetuning on the ICDAR dataset using SimCLR.  We perform top-4 selection with two strategies.

\begin{enumerate}
    \item {\bf Strategy 1: T-test based selection. } 
To ensure the effectiveness of our selection, we conducted a statistical analysis to identify the top four augmentations. This process involved running the SimCLR method across 93 different augmentations, using three random seeds for each. Through a paired t-test, we identified the top four augmentations, which yielded p-values of 0.0080, 0.01091, 0.00872, and 0.00874, respectively. Top-4 augmentations are shown below.%

\begin{enumerate}
    \item \texttt{randomcrop198,morpho\_dilation,hflip} \vspace*{-0.2cm}
    \item \texttt{randomcrop198,colorjitter,hflip,invert} \vspace*{-0.2cm}
    \item \texttt{randomcrop198,hflip,gray} \vspace*{-0.2cm}
    \item \texttt{randomcrop198,invert,gaussianblur,gray} \vspace*{-0.2cm}
\end{enumerate}

\item \noindent {\bf Strategy 2: Best average validation accuracy.} {In this strategy, we selected the top four augmentations based on the average performance across three runs. We sorted the augmentations and selected the top four from this list to compare results. The top-four augmentations from the sorted list are:}

\begin{enumerate}
    \item \texttt{randomcrop224,morpho\_erosion,morpho\_dilation,affine} \vspace*{-0.2cm}
    \item \texttt{randomcrop224,morpho\_dilation,affine,colorjitter} \vspace*{-0.2cm}
    \item \texttt{randomcrop224,morpho\_erosion,affine,colorjitter} \vspace*{-0.2cm}
    \item \texttt{randomcrop224,affine,colorjitter,gaussianblur} \vspace*{-0.2cm}
\end{enumerate}

\end{enumerate}

\noindent The selected top four augmentations were then used to pretrain the models on the ALPUB dataset. Following this pretraining, all models were finetuned on the ICDAR dataset by adding a classification layer to the embedding models and applying the base augmentation.\\

\noindent {\bf Our Findings.} We present the results of ICDAR letter recognition, leveraging pretraining on the ALPUB dataset and finetuning on ICDAR, in Tables~\ref{tab:resnet18_Alpub_results} and~\ref{tab:resnet50_Alpub_results} for ResNet-18 and ResNet-50, respectively. Consistent with our findings in Section~\ref{AppendixA}, the cross-entropy baseline achieved performances of 79.94\% on ResNet-18 and 79.75\% on ResNet-50, continuing to outperform the embedding-based methods. Specifically, the Triplet model attained performances of 77.51\% on ResNet-18 and 77.03\% on ResNet-50, while the SimCLR model reached 76.14\% on ResNet-18 and 76.59\% on ResNet-50. These results underscore the robustness of the cross-entropy approach, which consistently yields higher accuracy compared to the other methods tested. \\

\noindent Moreover, we observed that each method responded differently to the augmentations applied, with no single augmentation strategy emerging as universally optimal across all models. This variability highlights the importance of carefully selecting augmentation strategies tailored to each specific method and architecture. The impact of data augmentation on the training process is therefore significant, as it directly influences the performance outcomes for each model.\\

\noindent However, it is noteworthy that the final performance after finetuning on the ICDAR dataset, despite pretraining on the ALPUB dataset, reached only 79.75\%, which did not surpass the 80.47\% achieved in earlier experiments conducted without pretraining on the ALPUB dataset. This result is unexpected, as pretraining on a large-scale dataset like ALPUB was anticipated to improve the model's performance on the downstream ICDAR task.\\

\noindent The above results are with top-4 selected with strategy 1. Furthermore, in the following, we present the results with stragey 2. Table \ref{tab:resnet18_Alpub_results_sort} displays the results for the three models. The observed pattern is consistent with the pretraining results. The baseline model achieved a test accuracy of 81.14\%, while the embedding-based models, such as Triplet, reached 78.88\% and SimCLR achieved 79.18\%. For result 50 also follows same pattern as before, In table~\ref{tab:resnet50_Alpub_results_sort} we can observe that baseline model is performing best than other 2 embedding models. Test accuracies and baseline model, triple model and SimCLR models are 81.17\%, 78.24\%, 78.85\% \\

\begin{table}[h!]
\centering
\renewcommand{\arraystretch}{3} %
\begin{tabular}{|p{3cm}|p{1.8cm}|p{4cm}|p{1.4cm}|p{1.4cm}|}

\hline
\textbf{Experiment} & \textbf{Dataset} & \textbf{Best augmentation} & \textbf{Valid Acc.} & \textbf{Test Acc.} \\
\hline
Baseline model & \parbox{1.8cm} {\texttt{Alpub +  ICDAR}} & \parbox{4cm}{\texttt{randomcrop224, morpho\_erosion, affine, colorjitter}} & 80.68\% & {\bf81.14}\% \\ \hline 

Triplet model & \parbox{1.8cm} {\texttt{Alpub +  ICDAR}} & \parbox{4cm}{\texttt{randomcrop224, morpho\_dilation, affine, colorjitter}} & 79.57\% & 78.88\% \\ \hline

SimCLR model & \parbox{1.8cm} {\texttt{Alpub +  ICDAR}} & \parbox{4cm}{\texttt{randomcrop224, morpho\_erosion, affine, colorjitter}} & 79.74\% & 79.18\% \\
\hline
\end{tabular}
\caption{{\bf Results on ResNet-18 with pretraining on Alpub dataset {(with top-4 selected using strategy 2)}.}  We report the best found augmentation and their corresponding validation and test set accuracies. We observe the baseline model achieves the best results than other two methods. }
\label{tab:resnet18_Alpub_results_sort}
\end{table}

\begin{table}[h!]
\centering
\renewcommand{\arraystretch}{3} %
\begin{tabular}{|p{3cm}|p{1.8cm}|p{4cm}|p{1.4cm}|p{1.4cm}|}

\hline
\textbf{Experiment} & \textbf{Dataset} & \textbf{Best augmentation} & \textbf{Valid Acc.} & \textbf{Test Acc.} \\
\hline
Baseline model & \parbox{1.8cm} {\texttt{Alpub +  ICDAR}} & \parbox{4cm}{\texttt{randomcrop224, affine, colorjitter, gaussianblur}} & 81.35\% & \bf 81.17\% \\ \hline 

Triplet model & \parbox{1.8cm} {\texttt{Alpub +  ICDAR}} & \parbox{4cm}{\texttt{randomcrop224, morpho\_dilation, affine, colorjitter}} & 79.17\% & 78.24\% \\ \hline

SimCLR model & \parbox{1.8cm} {\texttt{Alpub +  ICDAR}} & \parbox{4cm}{\texttt{randomcrop224, affine, colorjitter, gaussianblur}} & 78.68\% & 78.85\% \\
\hline
\end{tabular}
\caption{{\bf Results on ResNet-50 with pretraining on Alpub dataset {(with top-4 selected using strategy 2)}.}  We report the best found augmentation and their corresponding validation and test set accuracies. We observe the baseline model achieves the best results than other two methods. }
\label{tab:resnet50_Alpub_results_sort}
\end{table}

\noindent This unexpected result—where pretraining on the large-scale ALPUB dataset did not enhance performance on the finetuned ICDAR dataset—warrants further investigation. To better understand this outcome, we provide embedding visualizations generated from different methods to support and explain these quantitative results. These visualizations will offer deeper insights into how the representations learned during pretraining might have affected the final model performance and why the expected improvements did not materialize.\\

\section{t-SNE Analysis}
\label{tsne_analysisi}

To gain a deeper understanding of the different methods, we visualize t-SNE plots of the embeddings 1. at the end of pretraining with ALPUB and 2. after finetuning with ICDAR. Specifically, we select approximately 1,000 samples from the ICDAR test set for this visualization. These embeddings are analyzed at these two stages of the training pipeline to provide insights into how the models' representations evolve at the end of pretraining and after final finetuning. \\

\begin{figure}[h!]
    \centering
    \begin{subfigure}[b]{0.49\textwidth}
        \centering
        \includegraphics[width=\textwidth]{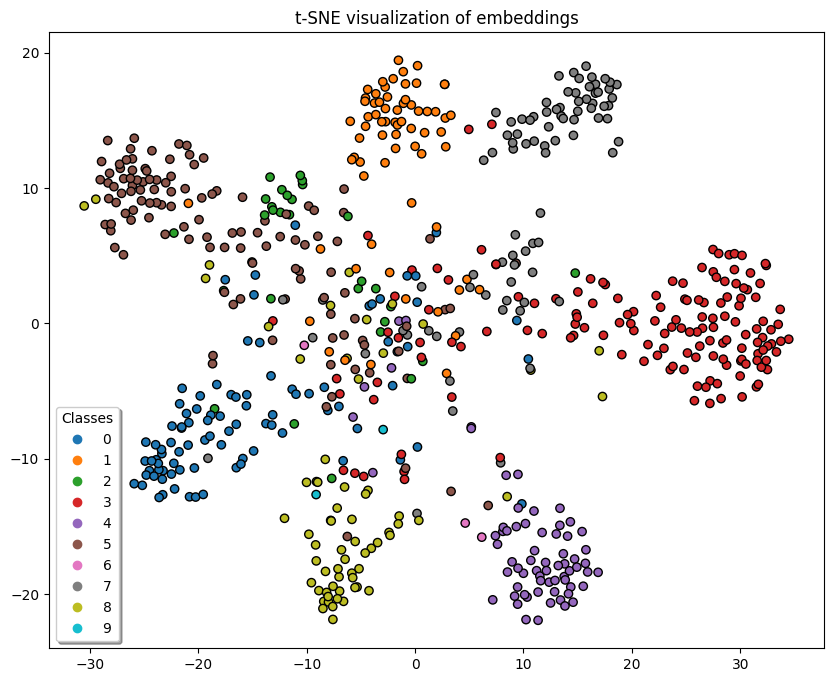}
        \caption{{\bf Embeddings at the End of Pretraining on ALPUB~\cite{alpub}}}
        \label{fig:ce_alpub_18}
    \end{subfigure}
    \hfill
    \begin{subfigure}[b]{0.49\textwidth}
        \centering
        \includegraphics[width=\textwidth]{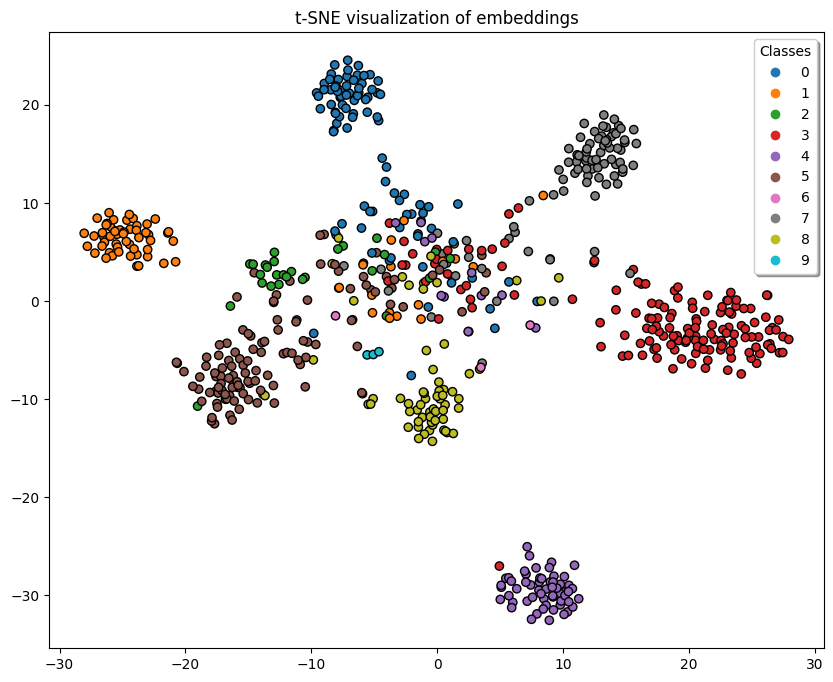}
        \caption{{\bf Embeddings at the End of Further Finetuning on ICDAR~\cite{icdar}}}
        \label{fig:ce_alpub_18_finetuning}
    \end{subfigure}
    \caption{{\bf Comparison of t-SNE Visualizations of the Baseline Model} at the end of pretraining with ALPUB (left) and after further finetuning with ICDAR (right). We visualize the embeddings of 1,000 data points from the ICDAR test set using the ResNet-18 backbone. The embeddings are derived from the feature representation just before the classification layer.}
    \label{fig:ce_comparison}
\end{figure}

\begin{figure}[h!]
    \centering
    \begin{subfigure}[b]{0.49\textwidth}
        \centering
        \includegraphics[width=\textwidth]{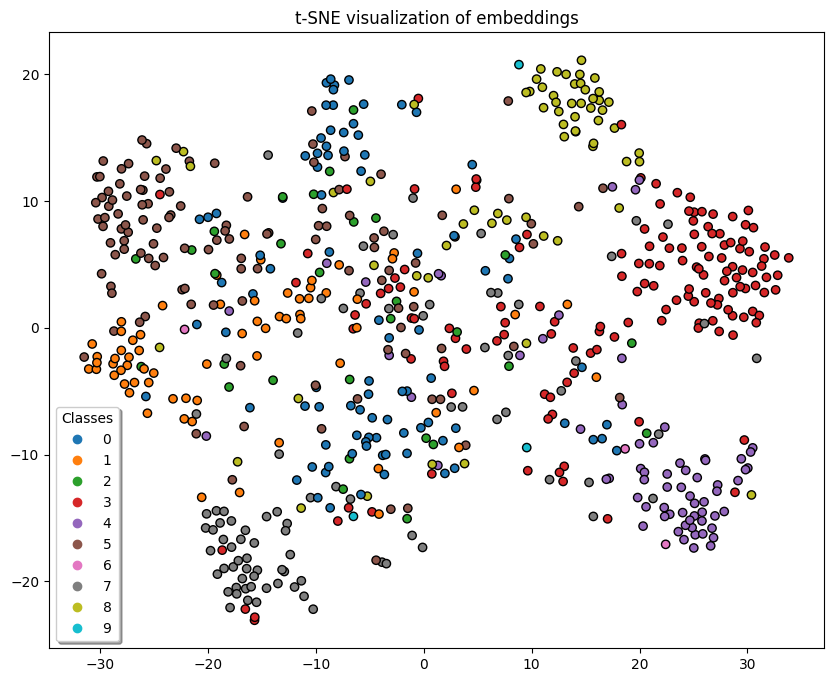}
        \caption{{\bf Embeddings at the End of Pretraining on ALPUB~\cite{alpub}}}
        \label{fig:triplet_tsne_alpub_18}
    \end{subfigure}
    \hfill
    \begin{subfigure}[b]{0.49\textwidth}
        \centering
        \includegraphics[width=\textwidth]{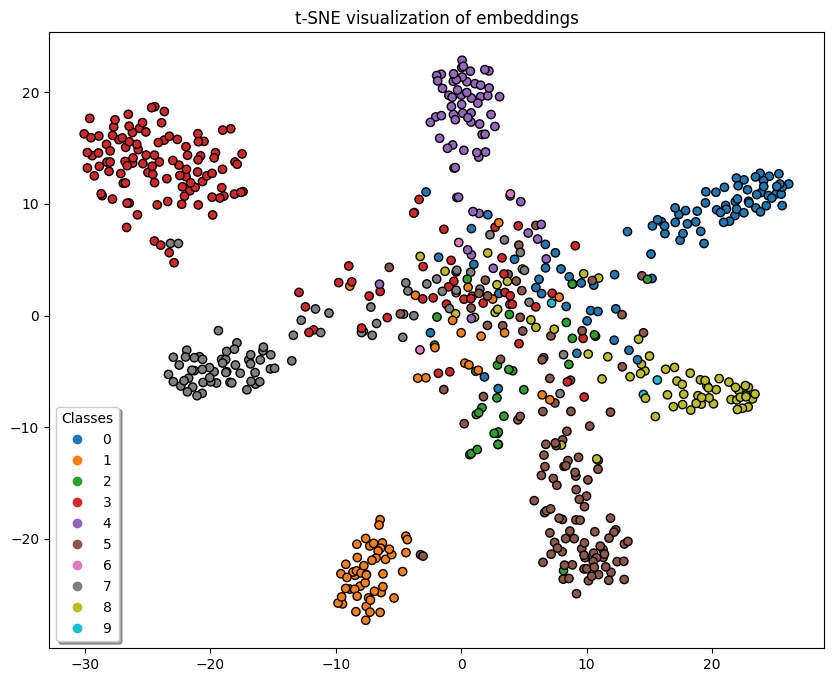}
        \caption{{\bf Embeddings at the End of Further Finetuning on ICDAR~\cite{icdar}}}
        \label{fig:triplet_tsne_alpub_18_finetune}
    \end{subfigure}
    \caption{{\bf Comparison of t-SNE Visualizations of the Triplet Model} at the end of pretraining with ALPUB (left) and after further finetuning with ICDAR (right). The visualizations depict embeddings of 1,000 data points from the ICDAR test set using the ResNet-18 backbone. The pretraining and finetuning was conducted with the augmentations: \texttt{randomcrop224, invert, gaussianblur, gray}.}
    \label{fig:triplet_comparison}
\end{figure}

\begin{figure}[h!]
    \centering
    \begin{subfigure}[b]{0.49\textwidth}
        \centering
        \includegraphics[width=\textwidth,height=6.7cm]{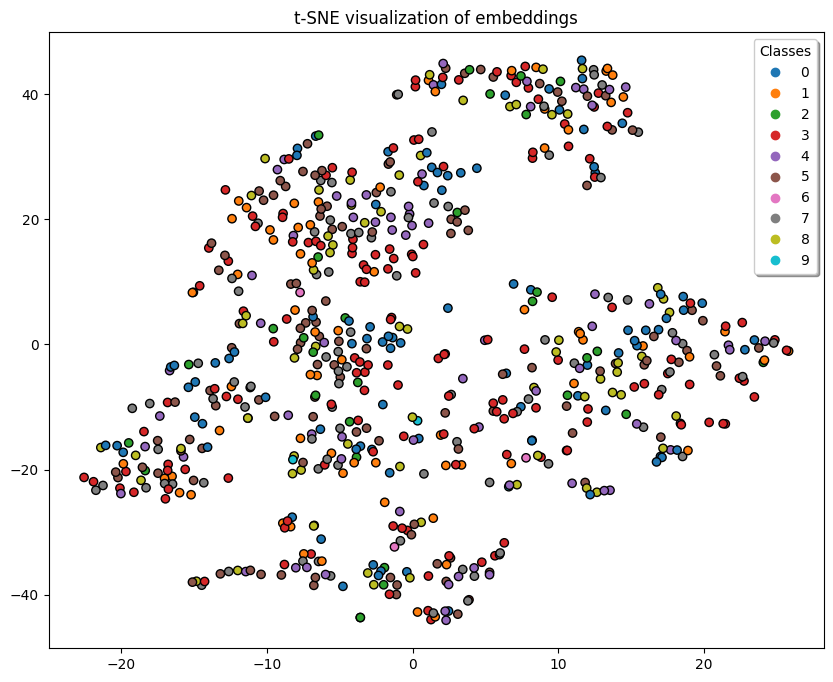}
        \caption{{\bf Embeddings at the End of Pretraining on ALPUB~\cite{alpub}}}
        \label{fig:simclr_tsne_alpub_18}
    \end{subfigure}
    \hfill
    \begin{subfigure}[b]{0.49\textwidth}
        \centering
        \includegraphics[width=\textwidth]{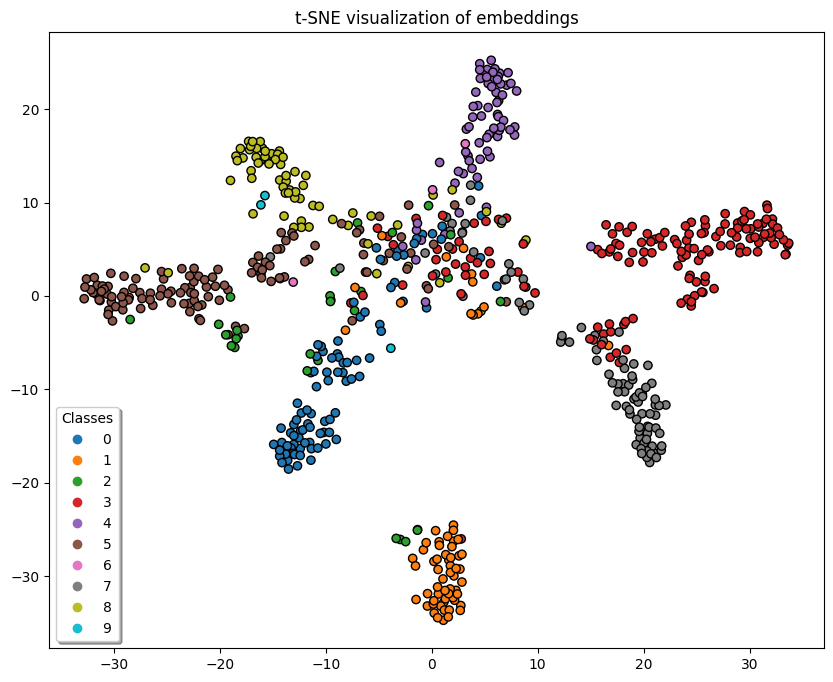}
        \caption{{\bf Embeddings at the End of Further Finetuning on ICDAR~\cite{icdar}}}
        \label{fig:simclr_alpub_finetune_18}
    \end{subfigure}
    \caption{{\bf Comparison of t-SNE Visualizations of the SimCLR Model} at the end of pretraining with ALPUB (left) and after further finetuning with ICDAR (right). The visualizations depict embeddings of 1,000 data points from the ICDAR test set using the ResNet-18 backbone. The pretraining was conducted with the augmentations: \texttt{randomcrop224, hflip, gray}.}
    \label{fig:simclr_comparison}
\end{figure}

\begin{enumerate}
    \item \noindent{\bf t-SNE with Baseline Model.} Figure \ref{fig:ce_alpub_18} shows the t-SNE visualization of 1,000 ICDAR test samples from the model obtained at the end of pretraining with ALPUB. Additionally, Figure \ref{fig:ce_alpub_18_finetuning} presents the embeddings after the model has been further finetuned with the ICDAR dataset. The t-SNE plots for the baseline model trained with cross-entropy loss exhibit well-defined clusters for each class both at the end of pretraining and after finetuning. This clear cluster separation indicates that the baseline model is highly effective at distinguishing between different letter classes in the embedding space, which likely contributes to its superior performance relative to the other methods.
    
\item \noindent{\bf t-SNE with Triplet Model.}
Figure \ref{fig:triplet_tsne_alpub_18} presents the t-SNE visualization of ICDAR test samples using the Triplet embedding model, which was pretrained on the ALPUB dataset. We observe that while the classes form distinct clusters, these clusters are not as clearly separated as those produced by the cross-entropy baseline method. This suggests that the Triplet model, though effective, may not be as precise in distinguishing between certain letter classes in the embedding space. Additionally, Figure \ref{fig:triplet_tsne_alpub_18_finetune} displays the t-SNE embeddings after the model has been finetuned with the ICDAR dataset, indicating that the cluster separations remain less distinct compared to the cross-entropy method.

\item 
\noindent{\bf t-SNE with SimCLR.} Figure \ref{fig:simclr_tsne_alpub_18} shows the t-SNE visualization of ICDAR test samples using the SimCLR embedding model pretrained on the ALPUB dataset. We observe that the classes are not well-clustered, raising concerns about the convergence of the SimCLR method on the ALPUB dataset. This lack of clear clustering suggests that the SimCLR model may struggle to learn distinct class separations during the pretraining phase. In contrast, Figure \ref{fig:simclr_alpub_finetune_18} visualizes the t-SNE embeddings after finetuning with the ICDAR dataset with an additional classification layer. Here, we see that the embeddings are more clearly separated, indicating some improvement in class distinction following finetuning. However, the additional benefit of pretraining with SimCLR on the ALPUB dataset appears limited, as the downstream performance on ICDAR does not show significant enhancement. These results are consistent with the quantitative observations discussed in Section~\ref{sec:result_wo_pretraining}, where the SimCLR method underperformed relative to the baseline CE method. Overall, the lack of clear clustering patterns in the t-SNE plots after pretraining suggests that SimCLR struggles to achieve strong class separation, even with the advantage of a larger pretraining dataset.
    
\end{enumerate}

\section{Summary}

In this chapter, we conducted an extensive evaluation of various models and training strategies for Greek letter recognition using the ICDAR dataset. The experiments were designed to explore the impact of different pretraining and finetuning techniques, particularly focusing on the effectiveness of pretraining on the large-scale ALPUB dataset. \\

\noindent In Section~\ref{results}, we began by examining the results of direct training on the ICDAR dataset without pretraining on the Alpub, comparing the performance of three distinct methods: the Baseline model, Triplet model~\cite{triplet}, and SimCLR model~\cite{simclr}. As shown in Tables~\ref{tab:resnet18_icdar_results}, \ref{tab:resnet50_icdar_results}, \ref{tab:resnet18_Alpub_results}, \ref{tab:resnet50_Alpub_results} and \ref{tab:resnet18_Alpub_results_sort}, the Baseline model trained with cross-entropy loss consistently outperformed the embedding-based methods across different architectures, highlighting the robustness of this approach in the absence of pretraining. \\

\noindent Subsequently in Section~\ref{datasets}, we introduced pretraining on the ALPUB dataset~\cite{alpub}, followed by finetuning on the ICDAR dataset~\cite{icdar}. Despite the expectation that pretraining on a larger and more diverse dataset would enhance performance, the results revealed no substantial improvements, with the cross-entropy Baseline model still achieving the best overall accuracy. This finding was further supported by t-SNE visualizations, which showed clearer class separations in the Baseline model compared to the Triplet and SimCLR models. \\

\noindent Furthermore, in Section~\ref{tsne_analysisi}, the t-SNE analysis provided deeper insights into how each model's embeddings evolved during the training process. Notably, the SimCLR model exhibited less distinct clustering of classes even after pretraining on a large scale Alpub dataset~\cite{alpub}, suggesting that the benefits of unsupervised contrastive learning might be limited in this specific application. \\

\noindent Overall, this chapter underscores the importance of careful method selection and the potential limitations of certain training strategies, such as contrastive learning, in specialized tasks like Greek letter recognition. The results suggest that traditional supervised learning approaches, particularly those utilizing cross-entropy loss, may still offer the most reliable performance in such contexts. In the following chapter, we delve into potential reasons for the unexpected performance observed in our experiments and highlight the key limitations within our experimental setup.

\chapter{Discussion and Limitations} %
\label{Chapter5}

In this chapter, we will try to understand better into the results presented in the previous chapters, offering few potential reasons of the outcomes observed across the different models and training strategies. By closely examining the performance metrics and visualizations, we aim to uncover the underlying factors that contributed to the models' successes and challenges. This analysis will help clarify the broader implications of our findings and provide insights into the effectiveness of various approaches in the context of Greek letter recognition. Additionally, we will explore the key limitations inherent in our experimental setup, which may have influenced the results. Understanding these constraints is crucial for assessing the validity and generalizability of our conclusions.

\begin{figure}[htbp]
    \centering
    \includegraphics[width=0.7\textwidth]{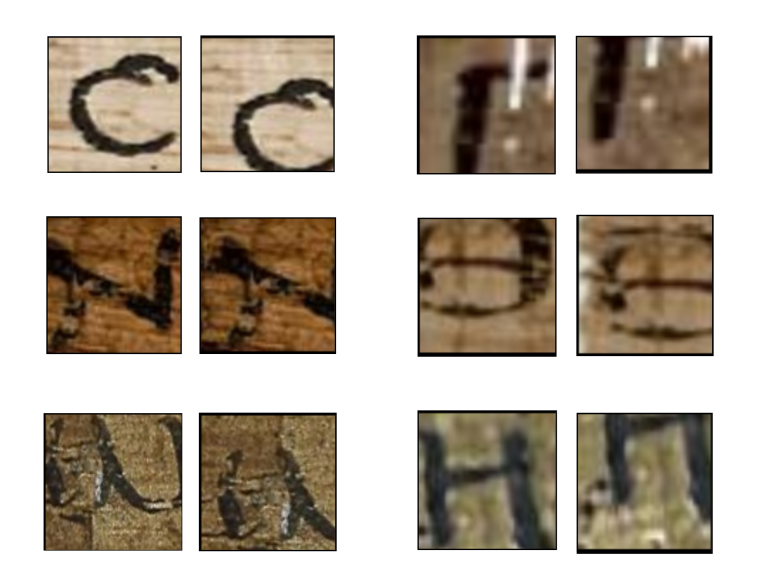}
   \caption{{\bf SimCLR cropping scheme leads to semantic shift in the labels.} For example, we observe the two views of the image cropped from the original image with 60\% area. It can be seen that this cropping scheme leads to a change in the labels.}
    \label{fig:simclr_letter}
\end{figure}

\section{Discussion}

\noindent {\bf Q1. What is the impact of the cropping scheme in SimCLR?} In Figure~\ref{fig:simclr_letter}, we observe that a spatial cropping scheme with a coverage of 60\% sometimes significantly alters the semantic content of the image, often resulting in a shift from one label to another. This raises a critical question: can the standard cropping techniques commonly used in natural image recognition tasks be directly applied to the domain of letter recognition, particularly for Greek letters? Our experimental results strongly suggest that SimCLR underperforms relative to baseline methods, a shortcoming that can largely be attributed to the semantic shift caused by the cropping process. Conversely, when the cropped region is too large, the resulting positive pairs in SimCLR become nearly indistinguishable, causing the network to struggle in learning meaningful features. These findings indicate that current cropping techniques may not be adequate for maintaining semantic integrity in letter recognition tasks. Therefore, we argue that further research is needed to develop a cropping scheme that preserves semantic consistency, enabling the effective application of SimCLR to Greek letter recognition.\\

\begin{figure}
    \centering
    \includegraphics[width=\linewidth]{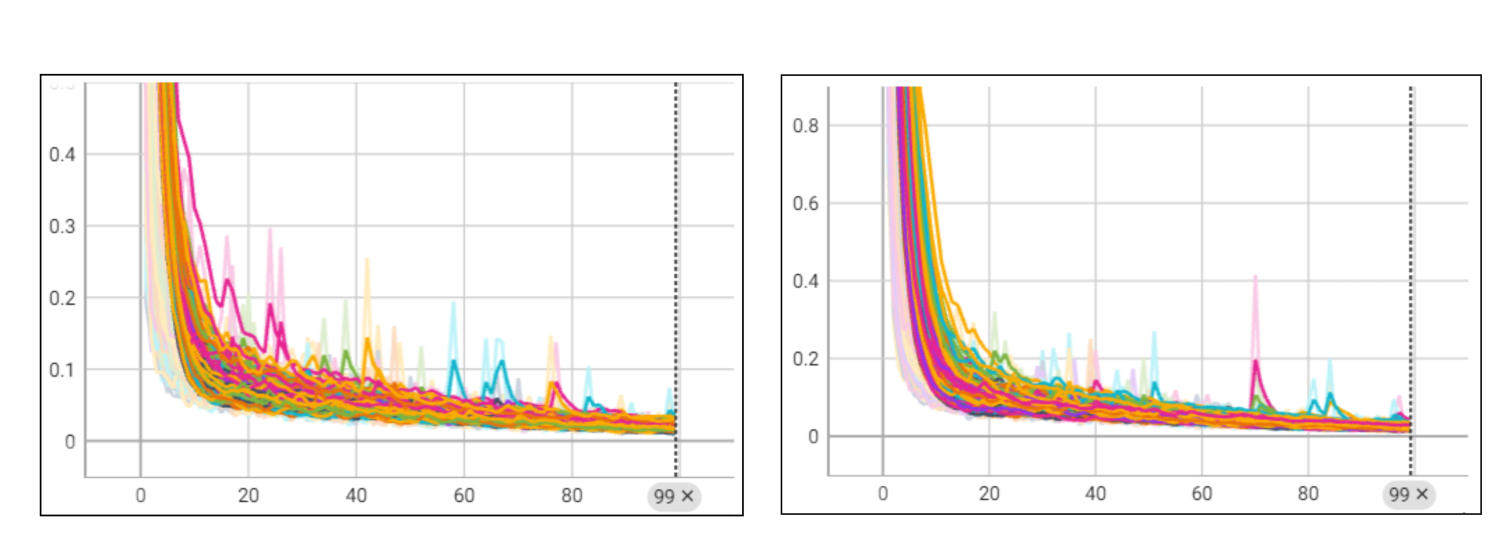}
    \caption{{\bf SimCLR Validation loss.} Comparison between ResNet-18 (left) and ResNet-50 (right) over 20 epochs.}
    \label{fig:simclr_loss}
\end{figure}

\noindent {\bf Q2. How to verify that SimCLR training is converging?} To determine whether SimCLR training is converging effectively, we analyze the validation loss across various augmentations over multiple epochs, as shown in Figure~\ref{fig:simclr_loss}. The decreasing trend in SimCLR loss throughout the training process suggests that the model is indeed improving. However, it is noteworthy that the pretraining of SimCLR on the extensive ALPUB dataset does not result in enhanced performance on the downstream tasks. This discrepancy may be attributed to errors introduced during the cropping phase in the pretraining stage, which could have negatively impacted the model's ability to generalize effectively to new datasets.

\section{Limitations}

This thesis, while presenting evidence on the performance of three widely used methods, acknowledges several limitations inherent in the conducted research. These limitations are detailed below:

\begin{enumerate}
    \item {\bf Hyperparameter Tuning.} Although we made efforts to optimize the data-augmentation strategies, a number of model-specific hyperparameters, such as dropout rates and optimizer-specific parameters including learning rate, momentum, beta, and decay rate, were not exhaustively tuned due to constraints in hardware resources. {Additionally, the temperature value also requires tuning.} This limitation may have impacted the overall performance and generalizability of the models.

    \item {\bf Data Augmentation.} Our study explored 10 data augmentation strategies in Section~\ref{sec:primary_augmentations}; however, the Albumentations library~\cite{albumentations} supports up to 40 augmentations, many of which were not examined in this research. Additionally, several of the augmentations implemented required the manual setting of hyperparameters, which were fixed at predefined values. A comprehensive sweep of these hyperparameters might have provided more extensive insights into the model's performance under varied conditions.

    \item {\bf Cropping Size for SimCLR.} The decision to crop 60\% of the original image area in SimCLR was heuristic and lacked a theoretical foundation. We recognize that this arbitrary choice in spatial cropping may have unintentionally altered the semantic content of the images, possibly resulting in label inconsistencies and affecting the model's performance.

    \item {\bf Batch Size in SimCLR.} SimCLR, a popular approach in metric learning, is known for its sensitivity to batch size. Typically, during pretraining, SimCLR is applied with a substantial batch size—often around 2048. This large batch size is critical as it allows the model to effectively leverage hard negatives within the batch, which are pivotal for the model’s ability to learn meaningful representations in a self-supervised manner. However, due to computational limitations in the present experiments, the batch size was significantly reduced to 115. This reduction implies that the model may not have encountered a sufficient number of hard negatives during training, potentially impacting its ability to learn robust representations.
    \item {{\bf Dataset construction} We randomly split the available dataset in three partitions:  70\%-15\%-15\% for training, validation, and testing sets, and use the same split for all experiments. 
    However, this potentially leads to bias and it would be worth looking into splitting multiple times and average the results over multiple runs.  
    }
\end{enumerate}

\chapter{Conclusion \& Future Work} %
\label{Chapter6} 

\section{Conclusion}
This thesis evaluates the performance of SimCLR~\cite{simclr}, a contrastive learning technique, for the task of Greek letter recognition~\cite{HWItallapragada2022greek} and compares it against traditional models that utilize cross-entropy and triplet loss functions~\cite{triplet}. A significant portion of this work (in Section~\ref{results}) involves ablating the data augmentation strategy across a pool of 93 different augmentations, comprising both spatial and pixel-level techniques~\ref{93_augmentation}. Through a comprehensive analysis involving both a large pretraining dataset (ALPUB~\cite{alpub}) and a smaller fine-tuning competition dataset~\cite{icdar}, it has become evident that SimCLR, despite its growing popularity in various image recognition tasks, does not outperform traditional supervised learning approaches such as cross-entropy baseline and triplet learning within the specific context of Greek letter recognition.\\

\noindent The unexpected underperformance of SimCLR raises intriguing questions. While it is challenging to pinpoint the exact reasons behind this result, our analysis in Chapter~\ref{Chapter4} and ~\ref{Chapter5} suggests that the method of cropping sub-images to generate positive pairs in SimCLR may introduce significant semantic shifts, which can be particularly detrimental in the context of letter images. This semantic drift likely contributes to the unsatisfactory performance observed. In conclusion, our results indicate that the baseline model trained with cross-entropy loss consistently outperforms both SimCLR and the triplet loss model.

\section{Future Work}

In this section, we present the few avenues for future research.

\begin{itemize}
    \item \textbf{Exploration of Additional Contrastive Learning Strategies:} Future work could investigate other contrastive learning frameworks and strategies, such as MoCo~\cite{moco} or BYOL~\cite{byol}, to determine if they offer better performance in specialized tasks like letter recognition.
    
    \item \textbf{Hyperparameter Tuning:} Experiments with different hyperparameter values, including learning rates, batch sizes, and the temperature parameter in SimCLR's contrastive loss function, could yield further insights into optimizing contrastive learning models for letter recognition.
    
    \item \textbf{Data Augmentation Strategies:} Additional research could focus on the impact of different data augmentation techniques, including varying cropping sizes and types of augmentations, on the performance of SimCLR and other contrastive learning models.
    
    \item \textbf{Integration with Supervised Learning:} Investigating hybrid approaches that combine contrastive learning with traditional supervised methods might provide a more robust solution, especially in cases where labeled data is limited.
    
    \item \textbf{Domain-Specific Contrastive Learning:} Future research could aim to develop contrastive learning techniques specifically tailored to the unique characteristics of Greek letters or other non-Latin scripts, potentially improving recognition accuracy in these domains.
\end{itemize}

\noindent We hope that our comprehensive empirical analysis of contrastive learning in Greek letter recognition will inspire future research aimed at developing domain-specific contrastive learning techniques tailored to such specialized tasks.\\

\noindent {\bf Note:} For writing this thesis, I have taken assistance of GPT for refining the text at few places.

\bibliographystyle{plain}  %
\bibliography{References}

\appendix %

\chapter{Full Results of Experiments} %
\label{AppendixA} %

In this chapter, we present all the computed results from our experiments, organized by model and training strategy.

\section{Baseline Model Results}
\subsection{ResNet-18 Architecture}
Tables~\ref{tab:baseline_ce_augmentation_results_18} provides the baseline model results using the ResNet-18 architecture on the fine-tuning dataset. The best-performing augmentations were \texttt{randomcrop224}, \texttt{morpho\_erosion}, \texttt{morpho\_dilation}, and \texttt{gaussianblur}, achieving a validation accuracy of 81.19\% and a test accuracy of 80.67\%.

\subsection{ResNet-50 Architecture}
Tables~\ref{tab:baseline_ce_augmentation_results_50} provides the baseline model results using the ResNet-50 architecture on the fine-tuning dataset. The top augmentations were \texttt{randomcrop224}, \texttt{morpho\_erosion}, and \texttt{gaussianblur}, resulting in a validation accuracy of 80.70\% and a test accuracy of 80.47\%.

\subsection{ALPUB Dataset Evaluation}
We evaluated the top four performing baseline models on the ALPUB dataset using both ResNet-18 and ResNet-50. The results are presented in Tables~\ref{tab:ce_18_alpub}, \ref{tab:ce_18_alpub_finetune}, \ref{tab:ce_50_alpub}, and~\ref{tab:ce_50_alpub_finetune}. For ResNet-18, the optimal augmentation combination was \texttt{randomcrop224}, \texttt{invert}, \texttt{gaussianblur}, and \texttt{gray}, yielding validation and test accuracies of 86.34\% and 86.72\%, respectively. Similarly, for ResNet-50, the best augmentation set was the same, resulting in validation and test accuracies of 86.47\% and 86.89\%, respectively. When fine-tuning on ICDAR, ResNet-18 achieved a maximum validation accuracy of 80.49\% and a test accuracy of 79.94\%. ResNet-50 reached a maximum validation accuracy of 80.15\% and a test accuracy of 79.98\%.

\section{Triplet Model Results}
\subsection{ResNet-18 Architecture}
The pretraining results for the Triplet model using the ResNet-18 architecture are shown in Table~\ref{tab:triplet_18_without_classification}. Fine-tuning results on the smaller dataset, both with and without the backbone, are available in Tables~\ref{tab:triplet_18_with_backbone_pretrain_icdar_finetune_icdar} and ~\ref{tab:triplet_18_without_backbone_pretrain_icdar_finetune_icdar}.

\subsection{ResNet-50 Architecture}
For the ResNet-50 model, pretraining on the small dataset results are shown in Table~\ref{tab:triplet_50_without_classification}. Fine-tuning results on the small dataset, with and without the backbone, are presented in Tables~\ref{tab:triplet_50_with_backbone_pretrain_icdar_finetune_icdar} and ~\ref{tab:triplet_50_without_backbone_pretrain_icdar_finetune_icdar}.

\subsection{ALPUB Dataset Evaluation}
The results of the Triplet model on the ALPUB dataset are provided in Tables~\ref{tab:triplet_18_pretrain_on_alpub}, ~\ref{tab:triplet_18_with_backbone_pretrain_alpub_finetune_icdar}, ~\ref{tab:triplet_18_without_backbone_pretrain_alpub_finetune_icdar}, ~\ref{tab:triplet_50_pretrain_on_alpub}, ~\ref{tab:triplet_50_with_backbone_pretrain_alpub_finetune_icdar}, and ~\ref{tab:triplet_50_without_backbone_pretrain_alpub_finetune_icdar}.

\section{SimCLR Model Results}
\subsection{ResNet-18 Architecture}
The pretraining results for the SimCLR model using the ResNet-18 architecture are shown in Table~\ref{tab:simclr_18_pretrain_icdar}. Fine-tuning results on the smaller dataset, both with and without the backbone, are available in Tables~\ref{tab:simclr_18_with_backbone_pretrain_icdar_finetune_icdar} and ~\ref{tab:simclr_18_without_backbone_pretrain_icdar_finetune_icdar}.

\subsection{ResNet-50 Architecture}
For the ResNet-50 model, pretraining on the ALPUB dataset results are shown in Table~\ref{tab:simclr_50_without_classification}. Fine-tuning results on the small dataset, with and without the backbone, are presented in Tables~\ref{tab:simclr_50_with_backbone_pretrain_icdar_finetune_icdar} and ~\ref{tab:simclr_50_without_backbone_pretrain_icdar_finetune_icdar}.

\subsection{ALPUB Dataset Evaluation}
SimCLR pretraining on the ALPUB dataset results for both ResNet-18 and ResNet-50 are available in Tables~\ref{tab:simclr_18_pretrain_on_alpub} and ~\ref{tab:simclr_50_pretrain_on_alpub}. Fine-tuning on the small dataset results for both ResNet-18 and ResNet-50, with and without backbone, are available in Tables~\ref{tab:simclr_18_with_backbone_pretrain_alpub_finetune_icdar}, ~\ref{tab:simclr_18__without_backbone_pretrain_alpub_finetune_icdar}, ~\ref{tab:simclr_50_pretrain_alpub_finetune_icdar_with_backbone}, and ~\ref{tab:simclr_50_pretrain_alpub_finetune_icdar_without_backbone}.

\section{Baseline Results with Cross-Entropy Loss Using ResNet-18 and ResNet-50 Networks}

\subsection{Baseline using ResNet-18 architecture on ICDAR Dataset}

\begin{longtable}[c]{|p{0.4cm}|p{9.3cm}|p{1.2cm}|p{1.5cm}|p{1.2cm}|}
\hline
\textbf{No.} & \textbf{Transform Type} & \textbf{Train Acc} & \textbf{Validation Acc} & \textbf{Test Acc} \\
\hline
\endfirsthead
\hline
\textbf{No.} & \textbf{Augmentations} & \textbf{Train Acc} & \textbf{Validation Acc} & \textbf{Test Acc} \\
\hline
\endhead
\small

1 & randomcrop224,morpho\_erosion,invert & 81.81\% & 79.54\% & 79.63\% \\
2 & randomcrop224,morpho\_erosion,colorjitter,gray & 83.70\% & 79.74\% & 79.00\% \\
3 & randomcrop224,morpho\_erosion,gaussianblur & 83.62\% & 80.39\% & 80.27\% \\
4 & randomcrop224,morpho\_erosion,affine,gaussianblur & 81.15\% & 80.66\% & 79.71\% \\
5 & randomcrop224,affine,colorjitter,gray & 79.35\% & 79.94\% & 79.02\% \\
6 & randomcrop224,affine,hflip,gray & 79.46\% & 79.29\% & 78.81\% \\
7 & randomcrop224,affine,gray & 78.89\% & 79.92\% & 79.59\% \\
8 & randomcrop224,invert,gaussianblur,gray & 79.95\% & 78.96\% & 79.30\% \\
\end{longtable}

\subsection{Baseline using ResNet-50 architecture on ICDAR Dataset}

\begin{longtable}[c]{|p{0.4cm}|p{9.3cm}|p{1.2cm}|p{1.5cm}|p{1.2cm}|}
\hline
\textbf{No.} & \textbf{Augmentations } & \textbf{Train Acc} & \textbf{Validation Acc} & \textbf{Test Acc} \\
\hline
\endfirsthead
\hline
\textbf{No.} & \textbf{Transform Type} & \textbf{Train Acc} & \textbf{Validation Acc} & \textbf{Test Acc} \\
\hline
\endhead

1 & randomcrop224,affine,hflip & 78.15\% & 79.12\% & 78.34\% \\
2 & randomcrop224,colorjitter,invert & 85.25\% & 79.49\% & 79.14\% \\
3 & randomcrop224,morpho\_erosion,morpho\_dilation, & 81.12\% & 80.29\% & 80.39\% \\
& colorjitter &&&\\
4 & randomcrop224,morpho\_dilation,affine,colorjitter & 80.87\% & 80.29\% & 80.18\% \\
5 & randomcrop224,affine,gaussianblur,gray & 80.24\% & 80.15\% & 79.61\% \\
6 & randomcrop224,colorjitter,invert,gray & 82.48\% & 79.51\% & 78.90\% \\
7 & randomcrop224,colorjitter,gray & 82.66\% & 80.64\% & 80.35\% \\
\end{longtable}

\subsection{Baseline using ResNet-18 architecture pertaining on Alpub Dataset}

\begin{longtable}[c]{|p{0.4cm}|p{9.3cm}|p{1.2cm}|p{1.5cm}|p{1.2cm}|}
\hline
\textbf{No.} & \textbf{Augmentations} & \textbf{Train Acc} & \textbf{Validation Acc} & \textbf{Test Acc} \\
\hline
\endfirsthead
\hline
\textbf{No.} & \textbf{Transform Type} & \textbf{Train Acc} & \textbf{Validation Acc} & \textbf{Test Acc} \\
\hline
\endhead
1 & randomcrop224,hflip,gray & 86.90\% & 85.82\% & 86.02\% \\
2 & randomcrop224,colorjitter,hflip,invert & 87.83\% & 85.70\% & 86.15\% \\
3 & randomcrop224,morpho\_dilation,hflip & 88.43\% & 86.02\% & 86.34\% \\
\textbf{4} & \textbf{randomcrop224,invert,gaussianblur,gray} & \textbf{90.24\%} & \textbf{86.34\%} & \textbf{86.72\%} \\
&&&&\\
5 & randomcrop224,morpho\_dilation,affine,colorjitter & 88.58\% & 86.96\% & 86.99\% \\
6 & randomcrop224,affine,colorjitter,gaussianblur & 87.57\% & 86.59\% & 86.73\% \\
7 & randomcrop224,morpho\_erosion,morpho\_dilation,affine & 88.78\% & 86.76\% & 87.05\% \\
\textbf{8} & \textbf{randomcrop224,morpho\_erosion,affine,colorjitter} & \textbf{89.38\%} & \textbf{87.03\%} & \textbf{87.02\%} \\

\hline
\caption{Performance of Baseline ResNet-18 Model with Cross-Entropy Loss Across Various Data Augmentation Combinations pertaining on Alpub Dataset}
\label{tab:ce_18_alpub}
\end{longtable}

\subsection{Baseline using ResNet-18 architecture pertaining on Alpub Dataset \& fine-tuning on ICDAR dataset}

\begin{longtable}[c]{|p{0.4cm}|p{9.3cm}|p{1.2cm}|p{1.5cm}|p{1.2cm}|}
\hline
\textbf{No.} & \textbf{Augmentations} & \textbf{Train Acc} & \textbf{Validation Acc} & \textbf{Test Acc} \\
\hline
\endfirsthead
\hline
\textbf{No.} & \textbf{Transform Type} & \textbf{Train Acc} & \textbf{Validation Acc} & \textbf{Test Acc} \\
\hline
\endhead
1 & randomcrop224,colorjitter,hflip,invert & 86.44\% & 79.62\% & 78.83\% \\
\textbf{2} & \textbf{randomcrop224,hflip,gray} & \textbf{84.16\%} & \textbf{80.49\%} & \textbf{79.94\%} \\
3 & randomcrop224,invert,gaussianblur,gray & 82.48\% & 79.62\% & 80.39\% \\
4 & randomcrop224,morpho\_dilation,hflip & 83.14\% & 80.29\% & 80.47\% \\
&&&&\\
\textbf{5} & \textbf{randomcrop224,morpho\_erosion,affine,colorjitter} & \textbf{86.36\%} & \textbf{80.68\%} & \textbf{81.14\%} \\
6 & randomcrop224,morpho\_dilation,affine,colorjitter & 80.07\% & 80.41\% & 79.67\% \\
7 & randomcrop224,affine,colorjitter,gaussianblur & 80.10\% & 80.56\% & 79.67\% \\
8 & randomcrop224,morpho\_erosion,morpho\_dilation,affine & 83.89\% & 80.45\% & 80.70\% \\

\hline
\caption{Performance of Baseline ResNet-18 Model with Cross-Entropy Loss Across Various Data Augmentation Combinations pertaining on Alpub Dataset \& fine-tuning on ICDAR dataset With backbone}
\label{tab:ce_18_alpub_finetune}
\end{longtable}

\subsection{Baseline using ResNet-50 architecture on Alpub Dataset}

\begin{longtable}[c]{|p{0.4cm}|p{9.3cm}|p{1.2cm}|p{1.5cm}|p{1.2cm}|}
\hline
\textbf{No.} & \textbf{Augmentations } & \textbf{Train Acc} & \textbf{Validation Acc} & \textbf{Test Acc} \\
\hline
\endfirsthead
\hline
\textbf{No.} & \textbf{Transform Type} & \textbf{Train Acc} & \textbf{Validation Acc} & \textbf{Test Acc} \\
\hline
\endhead
1 & randomcrop224,hflip,gray & 88.24\% & 85.73\% & 86.16\% \\
\textbf{2} & \textbf{randomcrop224,invert,gaussianblur,gray} & \textbf{88.34\%} & \textbf{86.47\%} & \textbf{86.89\%} \\
3 & randomcrop224,morpho\_dilation,hflip & 88.05\% & 85.84\% & 85.90\% \\
4 & randomcrop224,colorjitter,hflip,invert & 88.17\% & 85.67\% & 85.93\% \\
&&&&\\
\textbf{5} & \textbf{randomcrop224,morpho\_erosion,} & \textbf{88.12\%} & \textbf{87.23\%} & \textbf{87.35\%} \\
& \textbf{morpho\_dilation,affine} &&&\\
6 & randomcrop224,morpho\_dilation,affine,colorjitter & 89.38\% & 86.98\% & 86.77\% \\
7 & randomcrop224,affine,colorjitter,gaussianblur & 88.16\% & 86.92\% & 87.10\% \\
8 & randomcrop224,morpho\_erosion,affine,colorjitter & 89.60\% & 86.93\% & 87.06\% \\

\hline
\caption{Performance of Baseline ResNet-50 Model with Cross-Entropy Loss Across various Data Augmentation combinations pre-training Alpub Dataset}
\label{tab:ce_50_alpub}
\end{longtable}

\subsection{Baseline using ResNet-50 architecture on Alpub Dataset \& fine-tuning on ICDAR dataset}

\begin{longtable}[c]{|p{0.4cm}|p{9.3cm}|p{1.2cm}|p{1.5cm}|p{1.2cm}|}
\hline
\textbf{No.} & \textbf{Augmentations } & \textbf{Train Acc} & \textbf{Validation Acc} & \textbf{Test Acc} \\
\hline
\endfirsthead
\hline
\textbf{No.} & \textbf{Transform Type} & \textbf{Train Acc} & \textbf{Validation Acc} & \textbf{Test Acc} \\
\hline
\endhead
\textbf{1} & \textbf{randomcrop224,morpho\_dilation,hflip} & \textbf{86.00\%} & \textbf{80.21\%} & \textbf{79.75\%} \\
2 & randomcrop224,invert,gaussianblur,gray & 83.99\% & 80.15\% & 79.98\% \\
3 & randomcrop224,hflip,gray & 80.27\% & 80.04\% & 79.65\% \\
4 & randomcrop224,colorjitter,hflip,invert & 81.40\% & 79.78\% & 79.82\% \\
&&&&\\
5 & randomcrop224,morpho\_erosion,morpho\_dilation,affine & 80.01\% & 80.66\% & 80.72\% \\
\textbf{6} & \textbf{randomcrop224,affine,colorjitter,gaussianblur} & \textbf{83.25\%} & \textbf{81.35\%} & \textbf{81.17\%} \\
7 & randomcrop224,morpho\_dilation,affine,colorjitter & 83.20\% & 81.11\% & 80.92\% \\
8 & randomcrop224,morpho\_erosion,affine,colorjitter & 80.02\% & 80.94\% & 80.65\% \\

 \\
\hline
\caption{Performance of Baseline ResNet-50 Model with Cross-Entropy Loss Across various Data Augmentation combinations pre-training Alpub Dataset \& fine-tuning on ICDAR dataset With backbone}
\label{tab:ce_50_alpub_finetune}
\end{longtable}

\section{Baseline with Triplet embedding model Using ResNet-18 and ResNet-50 Networks}

\subsection{Triplet Model using ResNet-18 Pre-training on ICDAR Dataset}

\begin{longtable}[c]{|p{0.4cm}|p{9.3cm}|p{1.2cm}|p{1.5cm}|p{1.2cm}|}
\hline
\textbf{No.} & \textbf{Augmentations} & \textbf{Train Acc} & \textbf{Validation Acc} & \textbf{Test Acc} \\
\hline
\endfirsthead
\hline
\textbf{No.} & \textbf{Transform Type} & \textbf{Train Acc} & \textbf{Validation Acc} & \textbf{Test Acc} \\
\hline
\endhead
1 & randomcrop224,morpho\_dilation,invert,gaussianblur & 51.97\% & 52.26\% & 51.44\% \\
2 & randomcrop224,morpho\_dilation,invert & 97.84\% & 82.13\% & 79.41\% \\
3 & randomcrop224,affine,colorjitter,gray & 92.41\% & 70.97\% & 68.50\% \\
4 & randomcrop224,affine,hflip,gaussianblur & 51.59\% & 55.78\% & 50.89\% \\
5 & randomcrop224,colorjitter,hflip,invert & 48.09\% & 51.81\% & 47.97\% \\
6 & randomcrop224,hflip,gaussianblur & 74.19\% & 82.68\% & 81.20\% \\
7 & randomcrop224,invert & 98.01\% & 79.86\% & 77.13\% \\
8 & randomcrop224,affine,colorjitter & 52.43\% & 53.06\% & 52.63\% \\
\end{longtable}

\subsection{Triplet Model Using ResNet-18 fine-tuning on ICDAR Dataset Without Backbone}

\begin{longtable}[c]{|p{0.4cm}|p{9.3cm}|p{1.2cm}|p{1.5cm}|p{1.2cm}|}
\hline
\textbf{No.} & \textbf{Augmentations } & \textbf{Train Acc} & \textbf{Validation Acc} & \textbf{Test Acc} \\
\hline
\endfirsthead
\hline
\textbf{No.} & \textbf{Transform Type} & \textbf{Train Acc} & \textbf{Validation Acc} & \textbf{Test Acc} \\
\hline
\endhead

1 & randomcrop224,morpho\_dilation,affine & 10.92\% & 12.12\% & 11.35\% \\
2 & randomcrop224,morpho\_dilation,hflip & 11.49\% & 12.53\% & 11.82\% \\
3 & randomcrop224,morpho\_dilation,affine,gray & 46.75\% & 55.24\% & 53.70\% \\
4 & randomcrop224,morpho\_erosion,morpho\_dilation, & 11.28\% & 12.17\% & 11.33\% \\
& colorjitter &&&\\
5 & randomcrop224,affine,colorjitter,gray & 27.51\% & 31.85\% & 31.51\% \\
6 & randomcrop224,morpho\_dilation,affine,invert & 11.35\% & 12.31\% & 10.90\% \\
7 & randomcrop224,affine,hflip,invert & 10.77\% & 12.16\% & 11.49\% \\
\end{longtable}

\subsection{Triplet Model Using ResNet-18 fine-tuning on ICDAR Dataset With Backbone}

\begin{longtable}[c]{|p{0.4cm}|p{9.3cm}|p{1.2cm}|p{1.5cm}|p{1.2cm}|}
\hline
\textbf{No.} & \textbf{Augmentations} & \textbf{Train Acc} & \textbf{Validation Acc} & \textbf{Test Acc} \\
\hline
\endfirsthead
\hline
\textbf{No.} & \textbf{Transform Type} & \textbf{Train Acc} & \textbf{Validation Acc} & \textbf{Test Acc} \\
\hline
\endhead

1 & randomcrop224,morpho\_erosion & 81.90\% & 79.53\% & 78.69\% \\ 
2 & randomcrop224,affine,hflip,gaussianblur & 78.97\% & 78.49\% & 78.38\% \\ 
3 & randomcrop224,morpho\_erosion,morpho\_dilation, & 79.55\% & 78.98\% & 78.34\% \\ 
& colorjitter &&&\\
4 & randomcrop224,morpho\_erosion,affine,invert & 77.95\% & 78.99\% & 78.24\% \\ 
5 & randomcrop224,colorjitter,invert,gaussianblur & 81.43\% & 77.88\% & 77.46\% \\ 
6 & randomcrop224,morpho\_dilation,colorjitter & 80.56\% & 79.25\% & 79.35\% \\ 
7 & randomcrop224,invert,gaussianblur & 83.29\% & 78.37\% & 77.06\% \\
\end{longtable}

\subsection{Triplet model using ResNet-18 Pre-training on Alpub dataset}%

\begin{longtable}[c]{|p{0.4cm}|p{9.3cm}|p{1.2cm}|p{1.5cm}|p{1.2cm}|}
\hline
\textbf{No.} & \textbf{Augmentations} & \textbf{Train Acc} & \textbf{Validation Acc} & \textbf{Test Acc} \\
\hline
\endfirsthead
\hline
\textbf{No.} & \textbf{Transform Type} & \textbf{Train Acc} & \textbf{Validation Acc} & \textbf{Test Acc} \\
\hline
\endhead
1 & randomcrop224,colorjitter,hflip,invert & 99.27\% & 85.43\% & 83.68\% \\
2 & randomcrop224,hflip,gray & 99.15\% & 85.97\% & 83.68\% \\
\textbf{3} & \textbf{randomcrop224,invert,gaussianblur,gray} & \textbf{99.39\%} & \textbf{86.94\%} & \textbf{85.17\%} \\
4 & randomcrop224,morpho\_dilation,hflip & 99.44\% & 86.26\% & 85.12\% \\
&&&&\\
5 & randomcrop224,morpho\_erosion,affine,colorjitter & 99.53\% & 87.28\% & 85.41\% \\ 
\textbf{6} & \textbf{randomcrop224,morpho\_dilation,affine,colorjitter} & \textbf{99.33\%} & \textbf{87.54\%} & \textbf{87.15\%} \\ 
7 & randomcrop224,affine,colorjitter,gaussianblur & 99.25\% & 85.84\% & 82.89\% \\ 
8 & randomcrop224,morpho\_erosion,morpho\_dilation,affine & 98.74\% & 86.01\% & 82.99\% \\ 

\hline
\caption{Performance of Triplet ResNet-18 Model with tripet loss across various Data Augmentation combinations pertaining on Alpub Dataset}
\label{tab:triplet_18_pretrain_on_alpub}
\end{longtable}

\subsection{Triplet Model Using ResNet-18 fine-tuning on Alpub Dataset Without Backbone}

\begin{longtable}[c]{|p{0.4cm}|p{9.3cm}|p{1.2cm}|p{1.5cm}|p{1.2cm}|}
\hline
\textbf{No.} & \textbf{Augmentations } & \textbf{Train Acc} & \textbf{Validation Acc} & \textbf{Test Acc} \\
\hline
\endfirsthead
\hline
\textbf{No.} & \textbf{Transform Type} & \textbf{Train Acc} & \textbf{Validation Acc} & \textbf{Test Acc} \\
\hline
\endhead
\textbf{1} & \textbf{randomcrop224,morpho\_dilation,hflip} & \textbf{57.41\%} & \textbf{59.66\%} & \textbf{59.88\%} \\
2 & randomcrop224,hflip,gray & 54.89\% & 57.49\% & 59.06\% \\
3 & randomcrop224,colorjitter,hflip,invert & 47.08\% & 42.12\% & 42.58\% \\
4 & randomcrop224,invert,gaussianblur,gray & 43.18\% & 51.03\% & 52.47\% \\
&&&&\\
5 & randomcrop224,morpho\_erosion,affine,colorjitter & 53.62\% & 61.40\% & 61.21\% \\ 
6 & randomcrop224,morpho\_erosion,morpho\_dilation,affine & 54.93\% & 61.28\% & 61.55\% \\ 
7 & randomcrop224,affine,colorjitter,gaussianblur & 50.21\% & 57.31\% & 57.40\% \\ 
\textbf{8} & \textbf{randomcrop224,morpho\_dilation,affine,colorjitter} & \textbf{55.21\%} & \textbf{63.20\%} & \textbf{63.09\%} \\ 
\hline
\caption{Performance of Triplet ResNet-18 Model with tripet loss across various Data Augmentation combinations pertaining on Alpub Dataset \& fine-tuning on ICDAR dataset without backbone}
\label{tab:triplet_18_without_backbone_pretrain_alpub_finetune_icdar}
\end{longtable}

\subsection{Triplet Model Using ResNet-18 fine-tuning on Alpub Dataset With Backbone}

\begin{longtable}[c]{|p{0.4cm}|p{9.3cm}|p{1.2cm}|p{1.5cm}|p{1.2cm}|}
\hline
\textbf{No.} & \textbf{Augmentations} & \textbf{Train Acc} & \textbf{Validation Acc} & \textbf{Test Acc} \\
\hline
\endfirsthead
\hline
\textbf{No.} & \textbf{Transform Type} & \textbf{Train Acc} & \textbf{Validation Acc} & \textbf{Test Acc} \\
\hline
\endhead
\textbf{1} & \textbf{randomcrop224,morpho\_dilation,hflip} & \textbf{82.47\%} & \textbf{78.20\%} & \textbf{77.51\%} \\
2 & randomcrop224,hflip,gray & 79.86\% & 77.73\% & 77.03\% \\
3 & randomcrop224,colorjitter,hflip,invert & 79.63\% & 77.10\% & 76.32\% \\
4 & randomcrop224,invert,gaussianblur,gray & 79.82\% & 77.63\% & 76.79\% \\
&&&&\\
5 & randomcrop224,morpho\_erosion,affine,colorjitter & 83.52\% & 79.31\% & 78.59\% \\ 
\textbf{6} & \textbf{randomcrop224,morpho\_dilation,affine,colorjitter} & \textbf{80.97\%} & \textbf{79.57\%} & \textbf{78.88\%} \\ 
7 & randomcrop224,affine,colorjitter,gaussianblur & 81.35\% & 79.49\% & 79.02\% \\ 
8 & randomcrop224,morpho\_erosion,morpho\_dilation,affine & 81.68\% & 78.96\% & 79.49\% \\
\hline
\caption{Performance of Triplet ResNet-18 Model with tripet loss across various Data Augmentation combinations pertaining on Alpub Dataset \& fine-tuning on ICDAR dataset with backbone}
\label{tab:triplet_18_with_backbone_pretrain_alpub_finetune_icdar}
\end{longtable}

\subsection{Triplet model using ResNet-50 Pre-training on ICDAR dataset}

\begin{longtable}[c]{|p{0.4cm}|p{9.3cm}|p{1.2cm}|p{1.5cm}|p{1.2cm}|}
\hline
\textbf{No.} & \textbf{Augmentations} & \textbf{Train Acc} & \textbf{Validation Acc} & \textbf{Test Acc} \\
\hline
\endfirsthead
\hline
\textbf{No.} & \textbf{Transform Type} & \textbf{Train Acc} & \textbf{Validation Acc} & \textbf{Test Acc} \\
\hline
\endhead
1 & randomcrop224,morpho\_erosion,hflip,gray & 50.62\% & 52.59\% & 53.33\% \\
2 & randomcrop224,morpho\_erosion,morpho\_dilation,gray & 51.84\% & 52.40\% & 52.04\% \\
3 & randomcrop224,morpho\_erosion,morpho\_dilation,invert & 51.63\% & 52.81\% & 51.34\% \\
4 & randomcrop224,morpho\_erosion,colorjitter & 82.06\% & 79.35\% & 79.90\% \\
5 & randomcrop224,gaussianblur & 87.31\% & 80.80\% & 77.51\% \\
6 & randomcrop224,morpho\_dilation,affine,gray & 51.74\% & 53.81\% & 52.04\% \\
7 & randomcrop224,morpho\_erosion,morpho\_dilation,affine & 51.17\% & 52.67\% & 52.49\% \\
8 & randomcrop224,colorjitter,gaussianblur,gray & 51.68\% & 53.49\% & 53.08\% \\
\end{longtable}

\subsection{Triplet Model Using ResNet-50 pretrain on ICDAR dataset fine-tuning on ICDAR Dataset Without Backbone}

\begin{longtable}[c]{|p{0.4cm}|p{9.3cm}|p{1.2cm}|p{1.5cm}|p{1.2cm}|}
\hline
\textbf{No.} & \textbf{Augmentations} & \textbf{Train Acc} & \textbf{Validation Acc} & \textbf{Test Acc} \\
\hline
\endfirsthead
\hline
\textbf{No.} & \textbf{Transform Type} & \textbf{Train Acc} & \textbf{Validation Acc} & \textbf{Test Acc} \\
\hline
\endhead
1 & randomcrop224,morpho\_dilation,gaussianblur & 59.60\% & 62.15\% & 61.37\% \\
2 & randomcrop224,invert,gray & 11.49\% & 12.66\% & 11.66\% \\
3 & randomcrop224,morpho\_erosion,morpho\_dilation, & 11.20\% & 12.55\% & 11.47\% \\
& colorjitter &&&\\
4 & randomcrop224,morpho\_erosion,affine,hflip & 15.05\% & 16.99\% & 16.11\% \\
5 & randomcrop224,affine,colorjitter,gray & 10.78\% & 12.12\% & 11.35\% \\
6 & randomcrop224,colorjitter,invert,gaussianblur & 11.19\% & 12.29\% & 11.43\% \\
7 & randomcrop224,morpho\_erosion,colorjitter,gaussianblur & 17.97\% & 21.02\% & 20.53\% \\
\end{longtable}

\subsection{Triplet Model Using ResNet-50 pretrain on ICDAR dataset \& fine-tuning on ICDAR Dataset With Backbone}

\begin{longtable}[c]{|p{0.4cm}|p{9.3cm}|p{1.2cm}|p{1.5cm}|p{1.2cm}|}
\hline
\textbf{No.} & \textbf{Augmentations} & \textbf{Train Acc} & \textbf{Validation Acc} & \textbf{Test Acc} \\
\hline
\endfirsthead
\hline
\textbf{No.} & \textbf{Transform Type} & \textbf{Train Acc} & \textbf{Validation Acc} & \textbf{Test Acc} \\
\hline
\endhead
1 & randomcrop224,colorjitter,hflip,invert & 76.59\% & 76.36\% & 75.46\% \\
2 & randomcrop224,gaussianblur,gray & 80.43\% & 78.68\% & 77.65\% \\
3 & randomcrop224,morpho\_dilation,affine,colorjitter & 77.38\% & 78.72\% & 77.63\% \\
4 & randomcrop224,hflip,invert & 78.39\% & 77.75\% & 76.59\% \\
5 & randomcrop224,morpho\_erosion,affine,hflip & 73.22\% & 76.61\% & 76.38\% \\
6 & randomcrop224,morpho\_erosion,affine,invert & 76.73\% & 77.65\% & 77.34\% \\
7 & randomcrop224,morpho\_dilation,invert & 80.27\% & 78.51\% & 77.89\% \\
8 & randomcrop224,colorjitter,gray & 80.93\% & 78.45\% & 76.89\% \\
\end{longtable}

\subsection{Triplet model using ResNet-50 Pre-training on Alpub dataset}

\begin{longtable}[c]{|p{0.4cm}|p{9.3cm}|p{1.2cm}|p{1.5cm}|p{1.2cm}|}
\hline
\textbf{No.} & \textbf{Augmentations} & \textbf{Train Acc} & \textbf{Validation Acc} & \textbf{Test Acc} \\
\hline
\endfirsthead
\hline
\textbf{No.} & \textbf{Transform Type} & \textbf{Train Acc} & \textbf{Validation Acc} & \textbf{Test Acc} \\
\hline
\endhead
1 & randomcrop224,morpho\_dilation,hflip & 97.75\% & 88.55\% & 87.31\% \\
2 & randomcrop224,hflip,gray & 97.74\% & 89.25\% & 87.46\% \\
3 & randomcrop224,colorjitter,hflip,invert & 96.57\% & 87.04\% & 86.51\% \\
\textbf{4} & \textbf{randomcrop224,invert,gaussianblur,gray} & \textbf{98.17\%} & \textbf{89.74\%} & \textbf{89.26\%} \\
&&&&\\
5 & randomcrop224,affine,colorjitter,gaussianblur & 97.54\% & 90.47\% & 89.81\% \\
6 & randomcrop224,morpho\_erosion,morpho\_dilation,affine & 97.74\% & 90.00\% & 89.66\% \\
\textbf{7} & \textbf{randomcrop224,morpho\_erosion,affine,colorjitter} & \textbf{97.84\%} & \textbf{90.60\%} & \textbf{90.51\%} \\
8 & randomcrop224,morpho\_dilation,affine,colorjitter & 97.49\% & 89.65\% & 89.21\% \\
\\
\hline
\caption{Performance of Triplet ResNet-50 Model with tripet loss across various Data Augmentation combinations pertaining on Alpub Dataset}
\label{tab:triplet_50_pretrain_on_alpub}
\end{longtable}

\subsection{Triplet Model Using ResNet-50 pretrain on Alpub dataset \& fine-tuning on ICDAR Dataset Without Backbone}

\begin{longtable}[c]{|p{0.4cm}|p{9.3cm}|p{1.2cm}|p{1.5cm}|p{1.2cm}|}
\hline
\textbf{No.} & \textbf{Transform Type} & \textbf{Train Acc} & \textbf{Validation Acc} & \textbf{Test Acc} \\
\hline
\endfirsthead
\hline
\textbf{No.} & \textbf{Augmentations} & \textbf{Train Acc} & \textbf{Validation Acc} & \textbf{Test Acc} \\
\hline
\endhead
1 & randomcrop224,colorjitter,hflip,invert & 55.76\% & 57.37\% & 58.32\% \\
\textbf{2} & \textbf{randomcrop224,hflip,gray} & \textbf{61.83\%} & \textbf{63.38\%} & \textbf{63.80\%} \\
3 & randomcrop224,morpho\_dilation,hflip & 60.65\% & 61.48\% & 62.29\% \\
4 & randomcrop224,invert,gaussianblur,gray & 54.64\% & 63.01\% & 62.78\% \\
&&&&\\
5 & randomcrop224,affine,colorjitter,gaussianblur & 62.26\% & 66.84\% & 67.83\% \\
7 & randomcrop224,morpho\_dilation,affine,colorjitter & 60.05\% & 65.18\% & 66.13\% \\
8 & randomcrop224,morpho\_erosion,morpho\_dilation,affine & 62.79\% & 66.82\% & 67.08\% \\
\textbf{9} & \textbf{randomcrop224,morpho\_erosion,affine,colorjitter} & \textbf{62.96\%} & \textbf{67.10\%} & \textbf{67.48\%} \\ 
\hline
\caption{Performance of Triplet ResNet-50 Model with tripet loss across various Data Augmentation combinations pertaining on Alpub Dataset \& fine-tuning on ICDAR dataset without backbone}
\label{tab:triplet_50_without_backbone_pretrain_alpub_finetune_icdar}
\end{longtable}

\subsection{Triplet Model Using ResNet-50 pretrain on Alpub dataset \& fine-tuning on ICDAR Dataset With Backbone}

\begin{longtable}[c]{|p{0.4cm}|p{9.3cm}|p{1.2cm}|p{1.5cm}|p{1.2cm}|}
\hline
\textbf{No.} & \textbf{Augmentations} & \textbf{Train Acc} & \textbf{Validation Acc} & \textbf{Test Acc} \\
\hline
\endfirsthead
\hline
\textbf{No.} & \textbf{Transform Type} & \textbf{Train Acc} & \textbf{Validation Acc} & \textbf{Test Acc} \\
\hline
\endhead
1 & randomcrop224,hflip,gray & 77.56\% & 77.12\% & 76.20\% \\
2 & randomcrop224,colorjitter,hflip,invert & 75.88\% & 75.89\% & 74.60\% \\
\textbf{3} & \textbf{randomcrop224,invert,gaussianblur,gray} & \textbf{77.33\%} & \textbf{77.90\%} & \textbf{77.03\%} \\
4 & randomcrop224,morpho\_dilation,hflip & 79.29\% & 77.43\% & 76.79\% \\
&&&&\\
5 & randomcrop224,morpho\_erosion,morpho\_dilation,affine & 79.02\% & 78.39\% & 78.24\% \\
6 & randomcrop224,affine,colorjitter,gaussianblur & 79.58\% & 78.76\% & 78.61\% \\
\textbf{7} & \textbf{randomcrop224,morpho\_dilation,affine,colorjitter} & \textbf{78.51\%} & \textbf{79.17\%} & \textbf{78.24\%} \\
8 & randomcrop224,morpho\_erosion,affine,colorjitter & 79.85\% & 79.00\% & 78.22\% \\
\hline
\caption{Performance of Triplet ResNet-50 Model with tripet loss across various Data Augmentation combinations pertaining on Alpub Dataset \& fine-tuning on ICDAR dataset with backbone}
\label{tab:triplet_50_with_backbone_pretrain_alpub_finetune_icdar}
\end{longtable}

\section{SimCLR model Using ResNet-18 and ResNet-50 Networks}

\subsection{SimCLR model using ResNet-18 Pre-training on ICDAR dataset}

\begin{longtable}[c]{|p{0.4cm}|p{9.3cm}|p{1.2cm}|p{1.5cm}|p{1.2cm}|}
\hline
\textbf{No.} & \textbf{Augmentations} & \textbf{Train Acc} & \textbf{Validation Acc} & \textbf{Test Acc} \\
\hline
\endfirsthead
\hline
\textbf{No.} & \textbf{Transform Type} & \textbf{Train Acc} & \textbf{Validation Acc} & \textbf{Test Acc} \\
\hline
\endhead

1 & randomcrop198,morpho\_erosion,morpho\_dilation,invert & 99.97\% & 99.99\% & 99.96\% \\ 
2 & randomcrop198,morpho\_erosion,gray & 99.99\% & 99.98\% & 99.95\% \\ 
3 & randomcrop198,morpho\_dilation,affine,gaussianblur & 99.84\% & 99.91\% & 99.83\% \\ 
4 & randomcrop198,colorjitter,invert,gaussianblur & 99.97\% & 99.98\% & 99.88\% \\ 
5 & randomcrop198,hflip,invert,gaussianblur & 99.93\% & 99.99\% & 99.96\% \\ 
6 & randomcrop198,affine,colorjitter,hflip & 99.85\% & 99.79\% & 99.79\% \\ 
7 & randomcrop198,affine,colorjitter & 99.88\% & 99.90\% & 99.87\% \\ 
8 & randomcrop198,morpho\_erosion,morpho\_dilation,hflip & 99.83\% & 99.95\% & 99.83\% \\
\end{longtable}

\subsection{SimCLR Model Using ResNet-18 pretrain on ICDAR dataset \& fine-tuning on ICDAR Dataset Without Backbone}

\begin{longtable}[c]{|p{0.4cm}|p{9.3cm}|p{1.2cm}|p{1.5cm}|p{1.2cm}|}
\hline
\textbf{No.} & \textbf{Augmentations} & \textbf{Train Acc} & \textbf{Validation Acc} & \textbf{Test Acc} \\
\hline
\endfirsthead
\hline
\textbf{No.} & \textbf{Transform Type} & \textbf{Train Acc} & \textbf{Validation Acc} & \textbf{Test Acc} \\
\hline
\endhead
1 & randomcrop224,gray & 26.00\% & 26.66\% & 25.44\% \\ 
2 & randomcrop224,affine,colorjitter,gaussianblur & 32.69\% & 34.12\% & 33.44\% \\ 
3 & randomcrop224,colorjitter,hflip,invert & 32.53\% & 32.84\% & 32.35\% \\ 
4 & randomcrop224,affine,invert & 27.41\% & 28.13\% & 28.20\% \\ 
5 & randomcrop224,affine,gray & 23.27\% & 23.66\% & 23.07\% \\ 
6 & randomcrop224,affine,colorjitter,hflip & 31.31\% & 32.59\% & 32.82\% \\ 
7 & randomcrop224,morpho\_erosion,gaussianblur,gray & 29.26\% & 30.05\% & 28.65\% \\ 
8 & randomcrop224,invert,gray & 28.71\% & 28.46\% & 27.83\% \\
\end{longtable}

\subsection{SimCLR Model Using ResNet-18 pretrain on ICDAR dataset \& fine-tuning on ICDAR Dataset With Backbone}

\begin{longtable}[c]{|p{0.4cm}|p{9.3cm}|p{1.2cm}|p{1.5cm}|p{1.2cm}|}
\hline
\textbf{No.} & \textbf{Augmentations} & \textbf{Train Acc} & \textbf{Validation Acc} & \textbf{Test Acc} \\
\hline
\endfirsthead
\hline
\textbf{No.} & \textbf{Transform Type} & \textbf{Train Acc} & \textbf{Validation Acc} & \textbf{Test Acc} \\
\hline
\endhead
1 & randomcrop224,hflip,invert,gray & 77.75\% & 77.92\% & 76.95\% \\ 
2 & randomcrop224,hflip,gray & 81.36\% & 78.67\% & 77.42\% \\ 
3 & randomcrop224,invert & 84.67\% & 79.12\% & 78.65\% \\ 
4 & randomcrop224,morpho\_erosion & 82.77\% & 79.17\% & 79.10\% \\ 
5 & randomcrop224,colorjitter,hflip & 81.42\% & 78.63\% & 78.16\% \\ 
6 & randomcrop224,morpho\_erosion,morpho\_dilation, & 82.99\% & 79.94\% & 79.73\% \\ 
& gaussianblur &&&\\
7 & randomcrop224,morpho\_erosion,gaussianblur,gray & 82.60\% & 79.60\% & 79.37\% \\
\end{longtable}

\subsection{SimCLR model using ResNet-18 Pre-training on Alpub dataset}

\begin{longtable}[c]{|p{0.4cm}|p{9.3cm}|p{1.2cm}|p{1.5cm}|p{1.2cm}|}
\hline
\textbf{No.} & \textbf{Augmentations} & \textbf{Train Acc} & \textbf{Validation Acc} & \textbf{Test Acc} \\
\hline
\endfirsthead
\hline
\textbf{No.} & \textbf{Transform Type} & \textbf{Train Acc} & \textbf{Validation Acc} & \textbf{Test Acc} \\
\hline
\endhead
1 & randomcrop198,morpho\_dilation,hflip & 99.99\% & 99.99\% & 99.99\%\\
2 & randomcrop198,colorjitter,hflip,invert & 99.97\% & 99.99\% & 99.98\% \\
3 & randomcrop198,hflip,gray & 99.99\% & 99.99\% & 99.97\% \\
4 & randomcrop198,invert,gaussianblur,gray & 99.98\% & 99.99\% & 99.99\% \\
&&&&\\
5 & randomcrop198,morpho\_erosion,affine,colorjitter & 99.96\% & 99.99\% & 99.97\% \\
6 & randomcrop198,morpho\_erosion,morpho\_dilation,affine & 99.98\% & 99.99\% & 99.98\% \\
7 & randomcrop198,morpho\_dilation,affine,colorjitter & 99.98\% & 99.99\% & 99.97\% \\
8 & randomcrop198,affine,colorjitter,gaussianblur & 99.97\% & 99.98\% & 99.98\% \\
\hline
\caption{Performance of SimCLR model ResNet-18 architecture across various Data Augmentation combinations pertaining on Alpub Dataset}
\label{tab:simclr_18_pretrain_on_alpub}
\end{longtable}

\subsection{SimCLR Model Using ResNet-18 pretrain on Alpub dataset \& fine-tuning on ICDAR Dataset Without Backbone}

\begin{longtable}[c]{|p{0.4cm}|p{9.3cm}|p{1.2cm}|p{1.5cm}|p{1.2cm}|}
\hline
\textbf{No.} & \textbf{Transform Type} & \textbf{Train Acc} & \textbf{Validation Acc} & \textbf{Test Acc} \\
\hline
\endfirsthead
\hline
\textbf{No.} & \textbf{Augmentations} & \textbf{Train Acc} & \textbf{Validation Acc} & \textbf{Test Acc} \\
\hline
\endhead
1 & randomcrop224,morpho\_dilation,hflip & 19.21\% & 20.02\% & 19.16\% \\
2 & randomcrop224,invert,gaussianblur,gray & 21.66\% & 22.35\% & 21.78\% \\
3 & randomcrop224,hflip,gray & 21.38\% & 22.29\% & 21.15\% \\
\textbf{4} & \textbf{randomcrop224,colorjitter,hflip,invert} & \textbf{26.21\%} & \textbf{26.83\%} & \textbf{25.13\%} \\
&&&&\\
5 & randomcrop224,morpho\_dilation,affine,colorjitter & 26.08\% & 26.54\% & 25.36\% \\
\textbf{6} & \textbf{randomcrop224,morpho\_erosion,affine,colorjitter} & \textbf{26.30\%} & \textbf{28.32\%} & \textbf{26.28\%} \\
7 & randomcrop224,affine,colorjitter,gaussianblur & 26.89\% & 28.09\% & 27.42\% \\
8 & randomcrop224,morpho\_erosion,morpho\_dilation,affine & 17.28\% & 17.58\% & 17.44\% \\
\\
\hline
\caption{Performance of SimCLR model ResNet-18 architecture across various Data Augmentation combinations pertaining on Alpub Dataset \& fine-tuning on ICDAR dataset without backbone}
\label{tab:simclr_18__without_backbone_pretrain_alpub_finetune_icdar}
\end{longtable}

\subsection{SimCLR Model Using ResNet-18 pretrain on Alpub dataset \& fine-tuning on ICDAR Dataset With Backbone}

\begin{longtable}[c]{|p{0.4cm}|p{9.3cm}|p{1.2cm}|p{1.5cm}|p{1.2cm}|}
\hline
\textbf{No.} & \textbf{Augmentations} & \textbf{Train Acc} & \textbf{Validation Acc} & \textbf{Test Acc} \\
\hline
\endfirsthead
\hline
\textbf{No.} & \textbf{Transform Type} & \textbf{Train Acc} & \textbf{Validation Acc} & \textbf{Test Acc} \\
\hline
\endhead
1 & randomcrop224,morpho\_dilation,hflip & 78.64\% & 76.98\% & 76.75\% \\
\textbf{2} & \textbf{randomcrop224,colorjitter,hflip,invert} & \textbf{76.95\%} & \textbf{77.55\%} & \textbf{76.14\%} \\
3 & randomcrop224,hflip,gray & 75.70\% & 76.32\% & 75.62\% \\
4 & randomcrop224,invert,gaussianblur, gray & 78.68\% & 77.26\% & 76.81\% \\
&&&&\\
\textbf{5} & \textbf{randomcrop224,morpho\_erosion,affine,colorjitter} & \textbf{80.05\%} & \textbf{79.74\%} & \textbf{79.18\%} \\
6 & randomcrop224,morpho\_erosion,morpho\_dilation,affine & 78.39\% & 79.02\% & 78.98\% \\
7 & randomcrop224,affine,colorjitter,gaussianblur & 78.90\% & 79.29\% & 79.90\% \\
8 & randomcrop224,morpho\_dilation,affine,colorjitter & 78.75\% & 79.08\% & 78.20\% \\
\hline
\caption{Performance of SimCLR model ResNet-18 architecture across various Data Augmentation combinations pertaining on Alpub Dataset \& fine-tuning on ICDAR dataset with backbone}
\label{tab:simclr_18_with_backbone_pretrain_alpub_finetune_icdar}
\end{longtable}

\subsection{SimCLR model using ResNet-50 Pre-training on ICDAR dataset}

\begin{longtable}[c]{|p{0.4cm}|p{9.3cm}|p{1.2cm}|p{1.5cm}|p{1.2cm}|}
\hline
\textbf{No.} & \textbf{Augmentations} & \textbf{Train Acc} & \textbf{Validation Acc} & \textbf{Test Acc} \\
\hline
\endfirsthead
\hline
\textbf{No.} & \textbf{Transform Type} & \textbf{Train Acc} & \textbf{Validation Acc} & \textbf{Test Acc} \\
\hline
\endhead
1 & randomcrop198,hflip & 99.97\% & 100.00\% & 99.95\% \\
2 & randomcrop198,morpho\_dilation,gaussianblur & 99.96\% & 99.97\% & 99.92\% \\
3 & randomcrop198,morpho\_dilation,colorjitter,hflip & 99.91\% & 99.95\% & 99.82\% \\
4 & randomcrop198,affine,colorjitter,gray & 99.76\% & 99.80\% & 99.73\% \\
5 & randomcrop198,morpho\_erosion,morpho\_dilation, & 99.97\% & 99.98\% & 99.95\% \\
& gaussianblur &&&\\
6 & randomcrop198,morpho\_erosion,morpho\_dilation,affine & 99.87\% & 99.87\% & 99.80\% \\
7 & randomcrop198,morpho\_dilation,invert,gray & 99.93\% & 99.92\% & 99.86\% \\
\end{longtable}

\subsection{SimCLR Model Using ResNet-50 pretrain on ICDAR dataset \& fine-tuning on ICDAR Dataset Without Backbone}

\begin{longtable}[c]{|p{0.4cm}|p{9.3cm}|p{1.2cm}|p{1.5cm}|p{1.2cm}|}
\hline
\textbf{No.} & \textbf{Augmentations} & \textbf{Train Acc} & \textbf{Validation Acc} & \textbf{Test Acc} \\
\hline
\endfirsthead
\hline
\textbf{No.} & \textbf{Transform Type} & \textbf{Train Acc} & \textbf{Validation Acc} & \textbf{Test Acc} \\
\hline
\endhead
1 & randomcrop224,affine,invert,gray & 29.21\% & 29.83\% & 28.86\% \\
2 & randomcrop224,morpho\_erosion,morpho\_dilation, & 35.76\% & 37.37\% & 36.16\% \\
& colorjitter &&&\\
3 & randomcrop224,morpho\_erosion,hflip,gray & 26.03\% & 25.93\% & 26.01\% \\
4 & randomcrop224,affine,colorjitter,hflip & 31.68\% & 32.86\% & 32.33\% \\
5 & randomcrop224,affine,colorjitter,gaussianblur & 32.29\% & 35.08\% & 35.03\% \\
6 & randomcrop224,hflip,invert,gray & 25.63\% & 25.74\% & 25.54\% \\
7 & randomcrop224,morpho\_erosion,affine,gray & 26.96\% & 27.03\% & 27.57\% \\
\end{longtable}

\subsection{SimCLR Model Using ResNet-50 pretrain on ICDAR dataset \& fine-tuning on ICDAR Dataset With Backbone}

\begin{longtable}[c]{|p{0.4cm}|p{9.3cm}|p{1.2cm}|p{1.5cm}|p{1.2cm}|}
\hline
\textbf{No.} & \textbf{Augmentations} & \textbf{Train Acc} & \textbf{Validation Acc} & \textbf{Test Acc} \\
\hline
\endfirsthead
\hline
\textbf{No.} & \textbf{Transform Type} & \textbf{Train Acc} & \textbf{Validation Acc} & \textbf{Test Acc} \\
\hline
\endhead
1 & randomcrop224,morpho\_erosion,morpho\_dilation,affine & 77.50\% & 78.20\% & 77.79\% \\
2 & randomcrop224,morpho\_erosion,morpho\_dilation,invert & 81.64\% & 79.47\% & 79.49\% \\
3 & randomcrop224,morpho\_erosion,colorjitter,invert & 81.63\% & 79.59\% & 78.57\% \\
4 & randomcrop224,gaussianblur & 82.47\% & 79.27\% & 79.26\% \\
5 & randomcrop224,morpho\_dilation,affine,gaussianblur & 77.49\% & 79.53\% & 78.94\% \\
6 & randomcrop224,morpho\_erosion,colorjitter,hflip & 78.46\% & 78.67\% & 77.69\% \\
7 & randomcrop224,morpho\_dilation,colorjitter,gray & 81.92\% & 79.62\% & 80.08\% \\
8 & randomcrop224,morpho\_erosion,affine,colorjitter & 77.99\% & 79.15\% & 79.26\% \\
\end{longtable}

\subsection{SimCLR model using ResNet-50 Pre-training on Alpub dataset}

\begin{longtable}[c]{|p{0.4cm}|p{9.3cm}|p{1.2cm}|p{1.5cm}|p{1.2cm}|}
\hline
\textbf{No.} & \textbf{Augmentations} & \textbf{Train Acc} & \textbf{Validation Acc} & \textbf{Test Acc} \\
\hline
\endfirsthead
\hline
\textbf{No.} & \textbf{Transform Type} & \textbf{Train Acc} & \textbf{Validation Acc} & \textbf{Test Acc} \\
\hline
\endhead
1 & randomcrop198,morpho\_dilation,hflip & 99.98\% & 99.99\% & 99.99\% \\
2 & randomcrop198,invert,gaussianblur,gray & 99.98\% & 99.99\% & 99.98\% \\
3 & randomcrop198,hflip,gray & 99.98\% & 99.99\% & 99.97\% \\
4 & randomcrop198,colorjitter,hflip,invert & 99.99\% & 99.99\% & 99.97\% \\
&&&&\\
5 & randomcrop198,morpho\_dilation,affine,colorjitter & 99.97\% & 99.99\% & 99.97\% \\
6 & randomcrop198,morpho\_erosion,morpho\_dilation,affine & 99.97\% & 99.98\% & 99.97\% \\
7 & randomcrop198,morpho\_erosion,affine,colorjitter & 99.96\% & 99.99\% & 99.98\% \\
8 & randomcrop198,affine,colorjitter,gaussianblur & 99.97\% & 99.98\% & 99.96\% \\

\hline
\caption {Performance of SimCLR model ResNet-150 architecture across various Data Augmentation combinations pertaining on Alpub Dataset}
\label{tab:simclr_50_pretrain_on_alpub}
\end{longtable}

\subsection{SimCLR Model Using ResNet-50 pretrain on Alpub dataset \& fine-tuning on ICDAR Dataset Without Backbone}

\begin{longtable}[c]{|p{0.4cm}|p{9.3cm}|p{1.2cm}|p{1.5cm}|p{1.2cm}|}
\hline
\textbf{No.} & \textbf{Augmentations} & \textbf{Train Acc} & \textbf{Validation Acc} & \textbf{Test Acc} \\
\hline
\endfirsthead
\hline
\textbf{No.} & \textbf{Transform Type} & \textbf{Train Acc} & \textbf{Validation Acc} & \textbf{Test Acc} \\
\hline
\endhead
1 & randomcrop224,morpho\_dilation,hflip & 17.11\% & 17.81\% & 16.87\% \\
2 & randomcrop224, hflip, gray & 21.26\% & 21.90\% & 21.39\% \\
3 & randomcrop224,invert,gaussianblur,gray & 20.97\% & 21.61\% & 20.37\% \\
\textbf{4} & \textbf{randomcrop224,colorjitter,hflip,invert} & \textbf{24.94\%} & \textbf{25.07\%} & \textbf{23.95\%} \\
&&&&\\
5 & randomcrop224,affine,colorjitter,gaussianblur & 24.93\% & 26.25\% & 24.76\% \\
\textbf{6} & \textbf{randomcrop224,morpho\_dilation,affine,colorjitter} & \textbf{27.04\%} & \textbf{28.50\%} & \textbf{27.20\%} \\
7 & randomcrop224,morpho\_erosion,affine,colorjitter & 26.09\% & 27.03\% & 25.87\% \\
8 & randomcrop224,morpho\_erosion,morpho\_dilation,affine & 16.49\% & 16.34\% & 15.77\% \\
\hline
\caption{Performance of SimCLR model ResNet-50 architecture across various Data Augmentation combinations pertaining on Alpub Dataset \& fine-tuning on ICDAR dataset without backbone}
\label{tab:simclr_50_pretrain_alpub_finetune_icdar_without_backbone}
\end{longtable}

\subsection{SimCLR Model Using ResNet-50 pretrain on Alpub dataset \& fine-tuning on ICDAR Dataset With Backbone}

\begin{longtable}[c]{|p{0.4cm}|p{9.3cm}|p{1.2cm}|p{1.5cm}|p{1.2cm}|}
\hline
\textbf{No.} & \textbf{Augmentations} & \textbf{Train Acc} & \textbf{Validation Acc} & \textbf{Test Acc} \\
\hline
\endfirsthead
\hline
\textbf{No.} & \textbf{Transform Type} & \textbf{Train Acc} & \textbf{Validation Acc} & \textbf{Test Acc} \\
\hline
\endhead
1 & randomcrop224,morpho\_dilation,hflip & 74.14\% & 74.87\% & 74.46\% \\
\textbf{2} & \textbf{randomcrop224,invert,gaussianblur,gray} & \textbf{74.36\%} & \textbf{76.90\%} & \textbf{76.59\%} \\
3 & randomcrop224,hflip,gray & 74.33\% & 74.10\% & 73.97\% \\
4 & randomcrop224,colorjitter,hflip,invert & 75.67\% & 75.71\% & 74.03\% \\
&&&&\\
5 & randomcrop224,morpho\_erosion,affine,colorjitter & 76.31\% & 78.23\% & 78.16\% \\
6 & randomcrop224,morpho\_erosion,morpho\_dilation,affine & 73.56\% & 76.12\% & 75.52\% \\
7 & randomcrop224,morpho\_dilation,affine,colorjitter & 75.69\% & 78.53\% & 77.83\% \\
\textbf{8} & \textbf{randomcrop224,affine,colorjitter,gaussianblur} & \textbf{78.03\%} & \textbf{78.68\%} & \textbf{78.85\%} \\

\hline
\caption{Performance of SimCLR model ResNet-50 architecture across various Data Augmentation combinations pertaining on Alpub Dataset \& fine-tuning on ICDAR dataset with backbone}
\label{tab:simclr_50_pretrain_alpub_finetune_icdar_with_backbone}
\end{longtable}

\end{document}